\pdfoutput=1
\documentclass[11pt]{article}

\usepackage{acl}
\usepackage{times}
\usepackage{latexsym}

\usepackage[T1]{fontenc}
\usepackage[utf8]{inputenc}

\usepackage{microtype}

\usepackage{inconsolata}
\usepackage{graphicx}

\usepackage{array}
\usepackage{longtable}
\usepackage{hyperref}
\usepackage{comment}

\usepackage[table]{xcolor}
\usepackage{colortbl}
\usepackage{booktabs}
\usepackage{todonotes}

\usepackage{amssymb}  % Provides the checkmark and times symbols

\newcommand{\ie}{{\em i.e., }}
\newcommand{\eg}{{\em e.g., }}
\newcommand{\witem}{\vspace*{-1.2 ex}\item}
\usepackage[dvipsnames]{xcolor}
\usepackage{caption}
\usepackage{subcaption}

\title{
Is Machine Translation Ready for the Next Crisis or Pandemic?}

\author{Vipasha Bansal ~ William D. Lewis ~ Nicole Mihaylova ~ Elizabeth Brown ~ Benjamin Pong\\
Computational Linguistics Program \\ 
University of Washington\\
\{\texttt{vipashab},~\texttt{wlewis2},~\texttt{nicolemi},~\texttt{egollhof},~\texttt{benpong}\}\texttt{@uw.edu}
\\}

\begin{document}
\maketitle
\begin{abstract}

Communication in times of crisis is essential. However, there is often a mismatch between the language of governments, aid providers, doctors, and those to whom they are providing aid. %Commercial MT systems are reasonable tools to turn to in these scenarios. 
It is reasonable to turn to commercial MT tools in these scenarios. But how effective are these tools 
for translating to and from low resource languages, particularly in the crisis or medical domain? In this study, we evaluate 5 commercial systems used for MT %\todo{are you saying that GPT, Gemini and Claude are MT systems? Good point. Changed the text slightly.} 
(Google Translate, Microsoft Translator, GPT, Gemini and Claude) against the TICO-19 dataset, which contains pandemic-related sentences from a large set of high priority languages spoken by communities most likely to be adversely affected in the next pandemic. We assess the current degree of 
``readiness'' for this next pandemic based on the usability of the output translations. %We found that there is a large degree of variability in readiness across the languages covered by TICO, especially among the lesser resourced languages in the dataset.

%Communication in times of crisis is essential. However, there is often a mismatch between the language of aid providers and those to whom aid is provided. Commercial MT systems are reasonable tools to turn to in these scenarios. But how good are these commercial tools on the languages most likely to be affected by crisis? In this study, we examine four commercial MT systems to determine their degree of ``readiness'' for the next pandemic, using as our proxy the TICO-19 test which is composed of data for a large set of priority, low-resource languages most likely to be affected adversely in the next pandemic.
\end{abstract}

\section{Introduction}
\label{sec:intro}

COVID-19 was the worst pandemic in over a century. In order to slow the spread of the virus and to reduce hospitalizations and death, it was vital to communicate information to vulnerable populations. However, official communications from health agencies such the World Health Organization were only available in high resource languages.
%\cite{anastasopoulos-etal-2020-tico}.

The TICO-19 dataset~\cite{anastasopoulos-etal-TICO19-2020} was created during the 
%COVID-19 
pandemic to encourage MT research in under-resourced languages, specifically languages whose communities might be impacted by the pandemic but had not yet been touched by it.~\footnote{TICO-19 was created in mid-2020, early in the pandemic.}\textsuperscript{,}\footnote{A secondary goal of TICO-19 was to provide translation memories and glossaries for translators. These, along with the MT benchmark data used in this paper, are available for download from the TICO site: https://tico-19.github.io/.} Translators without Borders (TwB) was the driving force behind the choice of languages in TICO-19; TwB was, and still is, engaged in translating varied content for numerous languages spoken across Africa, Southeast Asia and other parts of the world.

The dataset contains n-way parallel test and dev sets across 36 languages. At the time of TICO-19's development, more than half of its languages had no commercial  translation systems available (\eg Dari, Kurdish, Zulu). This situation has changed markedly in the ensuing years: 36 of the 37 languages in the TICO-19 test set 
are now supported by at least one commercial system. This change in MT availability raises the following questions: How good are commercial MT systems at translating into and out of these languages? In particular, how good are they for translating pandemic-related content? Would they be adequate for aid providers, medical professionals, or governments to communicate with speakers of these languages on the ground during a future %pandemic or 
public health crisis?

Our study measures the quality of five existing commercial systems in the context of pandemic response. We use the TICO-19 dataset to systematically evaluate two major MT-specific providers, Google Translate and Microsoft Translator. Given the global popularity of LLMs, including in MT research~\cite{Kocmi-etal-WMTLLMS-2024}, and their potential competitiveness with existing translation services~\cite{Hendy-et-al-2023,robinson-etal-2023-chatgpt}, we also evaluate three %public and 
popular LLMs: %namely, 
GPT, Gemini and Claude. For the translation providers, we captured three snapshots over time, in 2023, 2025, and 2026, giving us a sense not only of their current performance, but also of how this performance has changed over the past three years.

%These give us a sense not only of how well the engines perform on pandemic-related content currently, but also how their performance has changed, in particular, on domain-specific content.  
%Further, given the global popularity of LLMs, including in MT research~\cite{Kocmi-etal-WMTLLMS-2024}, and their potential utility for translation (with strong evidence of them being competitive with existing commercial systems (\citealp{Hendy-et-al-2023,robinson-etal-2023-chatgpt})), we also look at how well OpenAI’s GPT, Google’s Gemini and Anthropic's Claude do on the same content in each language. Given the growth in the use of LLMs, it seems likely that current or future users will use LLMs for translation tasks, potentially in crisis scenarios such as pandemics.

We recognize the value of the output generated in our study to various fields, including Machine Translation, crisis response, medicine, and public health. We see this paper as the beginning and want to facilitate further research in this area. To this end, we are releasing all data collected in our study here: https://mt-crisis-ready.github.io.

\section{Background and Related Work}
\label{sec:background}
\subsection{The Use of MT and Other Language Technologies in Crisis Scenarios}
\label{sec:MT_in_crisis}

There has been a %significant 
large
amount of research on the use of language technologies in crisis response over the past several years. \citealp{Lewis-etal-2025}, for instance, reviews 355 papers published over a five-year timeframe ending in 2025. These papers focus on various language tasks, albeit specific to social media use in crises. Most concentrate on classification, with a smattering of work on other technologies including NER, geo-location, summarization, and topic modeling. Oddly, given the prevalence of large-scale crises that occur in areas where English is not spoken (see \citealp{tin-et-al-2024-global-disasters}, and discussion in~\citealp{Lewis-etal-2025}), almost none of the papers in the five-year study show work on translation or  MT as it relates to crises and social media.

That is not to say that there has been no work on crises and translation. The INTERACT project\footnote{https://cordis.europa.eu/project/id/734211}, an EU-funded research consortium, produced a number of publications on translation in crises, focused on various aspects of \textit{crisis communication} including translation workflows and training translators ~\cite{federici-cadwell-training-citizen-translators}, translation needs assessments in times of crisis ~\cite{obrien-federici-2019-crisis-translation-needs}, and ethics and crisis translation ~\cite{hunt-etal-2019-crisis-translation-ethics}. Similarly, \citealp{bird-2022-local-languages} looks at crisis communication holistically, considering cross-cultural communication and the use of "trade" languages (\ie regional \textit{linguae francae}). Further, \citealp{p-lankford-2024} examine the use of LLMs for low-resource MT in a pandemic, and \citealp{p-roussis-2022} walks through the data collection and training of an end-to-end neural model dedicated to pandemic translations. And of course, \citealp{lewis-2010-haitian} established the crucial need for commercial MT for low-resource languages in a crisis scenario.

Both commercial MT tools %---such as Google Translate, Microsoft Translator, and even LLMs such as GPT, Gemini and Claude---
and LLMs have seen broad general-purpose usage and acceptance, and commercial MT systems are often used in medical settings, including for doctor-patient communication ~\cite{khoong-etal-2019-assessing, khoong-rodriguez-2022-research}. It is likely that MT is used similarly in 
%all manner of 
crises, with some work showing MT being used by crisis managers in the United States \cite{wilson-etal-2026,Foubert-et-al-2026-roleofAIinEM}.

\subsection{Quality, Usability and Readiness}
\label{sec:quality-usability-readiness}

The end goal of much MT research is to design systems that produce high-quality translations across domains, \eg in bake-offs and competitions~\cite{Haddow-et-al-2025-wmt,iwslt-2026}.  
Quality is often measured for competing systems using automated metrics, such as CHRF++~\cite{Popvic-CHRFPP-2017} and BLEU~\cite{papineni2002bleu}, or human evaluation~\cite{graham-etal-2013-continuous}.

Our focus in this paper however, is to look at the readiness of competing systems in the context of pandemic response.  Our goal is therefore to assess whether a translation system is good enough to be used in the field during an active crisis. The key to this is usability: it is essential that the translated information be \textit{understood}, whether the translations are \textit{to} a population from, say, aid workers or government bodies~\cite{obrien-federici-2019-crisis-translation-needs}, or \textit{from} a population, say, as posted to social media ~\cite{Lewis-etal-2025,Imran-2015-survey}.

So how does one %go about measuring 
measure usability, especially in the context of  MT usage in a crisis scenario? One could ideally identify tasks that the MT would be used for in the field, \eg communicating directly with those affected, or sending emergency alerts to multilingual populations \cite{wilson-etal-2026,Foubert-et-al-2026-roleofAIinEM}. Although usability is not the same as quality,
%---and there are additional factors beyond translation that impact the outcome for any of these tasks---
there is a minimum threshold for quality that translations must meet, or any task dependent on those translations will fail. But how do we determine if output translations meet such a threshold? Further, how do we set this threshold?
%if a minimum quality bar is not achieved, a translation will not be usable, and all dependent tasks will fail. \todo{These last two sentences are saying essentially the same thing.}Thus, 
%there is a minimum threshold for quality that translations must meet below which no task will succeed. 

We use four evaluation metrics against our translated output: BLEU~\cite{papineni2002bleu}, CHRF++~\cite{Popvic-CHRFPP-2017}, BERTScore~\cite{Zhang-2020BERTScore}, and COMET~\cite{rei-etal-2020-comet}. Of these, BLEU and CHRF++ are n-gram based; despite their dependence on matching surface output with reference(s),\footnote{One of the main weaknesses of n-gram metrics is that an output could preserve meaning when compared with a reference but be assigned a low score because there is low n-gram overlap between the two~\cite{google_cloud_eval}.} they have been shown to correlate well with human evaluation, the gold standard %of any evaluation
~\cite{papineni2002bleu,Popvic-CHRFPP-2017}. 
%Further, they  have broad language coverage, supporting all languages in our study.

BERTScore and COMET are ``semantic'' based, and instead focus on the holistic ``meaning'' of the output~\cite{Zhang-2020BERTScore}. Unfortunately, these metrics do not adequately support many of the low resource languages in the TICO-19 dataset. This lack of coverage, along with issues with reliable performance even on supported low-resource languages and on specialized (or unseen) domains, has been noted in various studies \cite{nllb-2022, tang-etal-BERTScore-low-resource-2024, falcao-etal-2024-comet, zouhar-etal-2024-pitfalls, zouhar-etal-2024-domain}. In particular, \citet{zouhar-etal-2024-domain} notes the poor performance of COMET in the biomedical domain---%coincidentally 
a domain close to that of TICO-19.

Given the limited language coverage of BERTScore and COMET across the TICO-19 dataset, we turned to BLEU and CHRF++ to build our threshold. We follow a traditional rule-of-thumb to interpret BLEU scores, captured by Google's MT API documentation~\footnote{\url{https://cloud.google.com/translate/docs/advanced/automl-evaluate}} \cite{googleCloud} and shown in Table~\ref{bleuscoreinterpretation}. We see 30 as our usability threshold, as this indicates ``understandable to good'' quality translations (per the table). With CHRF++, we follow \citet{nllb-2022}, who identified a CHRF++ score of 45 as a minimum threshold. %for the removal of most toxic content
Although their task was focused on toxicity detection, we see this as a reasonable starting point. %for CHRF++.
%to effectively act as a toxicity detector. 
Section \ref{sec:eval_metrics} provides further details on how we use these metrics for thresholding, and Section \ref{sec:thresholds} outlines the results of a preliminary human evaluation to verify our proposal. \footnote{We realize that there is significant debate about using n-gram based metrics, particularly BLEU, for thresholding. Please refer to Appendix~\ref{app:eval_metric_choice} for additional discussion on this topic and related issues.}

\begin{table}[hbt!]
\centering
\small
  \begin{tabular}{@{} l *{2}{c} @{}}
\toprule
\textbf{BLEU Score} & \textbf{Interpretation} \\ 
\midrule
< 10      & Almost useless \\ 
10 - 19   & Hard to get the gist \\ 
20 - 29   & Clear gist, major grammatical errors \\
30 - 40   & Understandable to good translations \\ 
40 - 50   & High quality translations \\
50 - 60   & Very high quality, fluent translations \\ 
> 60      & Quality often better than human \\
\bottomrule
\end{tabular}
\caption{BLEU Score Interpretation}
\label{bleuscoreinterpretation}
\end{table}

\section{The TICO-19 Dataset}
\label{sec:data}
%\subsection{The TICO-19 Dataset}
%\label{sec:TICO-data}

The dataset consists of 3,071 sentences (971 dev, 2,100 test) drawn from %sources such as 
Wikipedia, PubMed, and existing corpora such as the CMU English-Haitian Creole dataset \cite{anastasopoulos-etal-TICO19-2020}. The sentences were translated into 37 languages, %thus 
making TICO uniquely n-ways parallel.
%\footnote{Such parallelism allows for MT development and testing of any pairing of languages in the dataset, \eg Oromo to Amharic, Marathi to Hindi, etc. It can also facilitate work on closely related dialects, \eg Eritrean and Ethiopian Tigrinya.} 
There are 9 \textit{pivot} languages, representing global or regional \textit{linguae francae}; 20 \textit{priority} languages identified by TwB based on translation requests during the pandemic or their importance to partner aid organizations such as the Red Cross; and 8 additional languages labeled \textit{important}, included because they have large populations of speakers across Asia. %in South and Southeast Asia. 
%The pivot languages were not explicitly requested by TwB, but were added to allow for extending TICO-19 in the future. 
%\footnote{Pivots were, in fact, used during the development of TICO-19 dataset, notably, for Dari (from Farsi) and Congolese Swahili (from French).}
Although these labels were not explicitly designed to map to language resource level, they align remarkably well: %with it: 
Pivot languages tend to be high resource, Priority languages low resource, and Important languages somewhere in between.

The dataset's structure allows for bidirectional evaluation that enables the measurement of MT quality in scenarios where English could be either the source, \eg of public health content, or the destination, \eg for translated reports or social media. %Additionally, each sentence is tagged with information about its source. 
Domain-source correspondence is shown in Table~\ref{tab:domain-list}. We utilize these subdomains to examine potential training data contamination (see Appendix \ref{app:contamination}).%\footnote{
%Sentences from the Travel domain appeared only in the dev set (and not test), and were not included in our analysis.}

%Two languages were removed from our study due to technical issues: the data for Congolese Swahili was not available for download, and the data file for Nuer was corrupted.

Two languages were excluded from our study: the data for Congolese Swahili was unavailable for download, and the data file for Nuer was corrupted.

\begin{table}[h]
\centering
\small
\begin{tabular}{lcc}
\toprule
\textbf{Data Source} & \textbf{Domain} & \textbf{\# Sentences}  \\
\midrule
CMU         & Conversational, medical     & 104  \\
PubMed      & Scientific, medical   	   & 676	  \\
Wikipedia   & General                      & 1,153 \\
Wikivoyage  & Travel         	               & dev only \\
Wikinews    & News         	 	           & 66 \\
Wikisource  & Announcements 	 	           & 99 \\
\bottomrule
\end{tabular}
\caption{TICO-19 Test Data Statistics by Domain}
\label{tab:domain-list}
\end{table}

\section{Methods}
\label{sec:methods}

%This section outlines the process used to produce and evaluate translations for each language for both the neural systems and the LLMs.

\subsection{Dedicated Machine Translation Systems}
We evaluated Google and Microsoft Translation services for three time frames: April-May 2023, April-July 2025, and May-June 2026. We used Microsoft Azure Cognitive Services Translator in 2023 and Azure Translator in 2025 and 2026\footnote{Although there were changes to the Azure API between 2023 and 2026, likely indicating reorganizations within Microsoft, 
the underlying translation service remained the same.} and Google Cloud's API throughout to access each model. The test sentences were identical across all three snapshots.\footnote{
The TICO-19 benchmark dataset has remained stable since its release in 2020, with no changes made to its  test data.
}

%These services were used to translate all languages in the dataset both to and from English.

%We are so short on space, I'm removing the locale info for now.
%Where locales listed in API documentation coincided with the locales in the TICO-19 dataset, \eg Brazilian Portuguese, Latin American Spanish, this information was passed to the API, \eg pt-BR, es-LA. \textcolor{red}{Two notable cases where locales were not specified by either MT system were Tigrinya and Fulfulde; TICO-19 has data specifically for Nigerian Fulfulde, and for two distinct dialects of Tigrinya (Eritrean and Ethiopian)}.  .

%For Tigrinya and Fulfulde, neither Microsoft nor Google indicated the %locales 
%locale supported by their translation engines. TICO-19 has data for two dialects of Tigrinya (Eritrean and Ethiopian) as well as Nigerian Fulfulde. \textcolor{red}{We did not specify locales to the API when processing these data.} 

%for these dialects were processed against the non-locale specific engines available. %(without specifying the locale in the argument string to the API). 
%did not specify the dialects supported by their translation engines for Tigrinya and Fulfulde, so the translations from English to standard Tigrinya and Fulfulde were used for Ethiopian Tigrinya, Eritrean Tigrinya and Nigerian Fulfulde. 

\subsection{LLM systems}
\label{sec:llm_systems}

We evaluated three GPT models (GPT-4o, GPT-5.4, and GPT-5.4-mini), 
Gemini-3.1-Flash-Lite, (for which translation is listed as a recommended use case),\footnote{https://ai.google.dev/gemini-api/docs/models/gemini-3.1-flash-lite} and Claude-Haiku-4.5. The models were accessed through Azure's OpenAI Service, Google’s Cloud API, and Anthropic's Claude Platform. \footnote{See Appendix \ref{app:modelconfigurations} for details on model configurations.}

Unlike the dedicated MT systems, which provide specific and documented language translation support, LLMs can generate text in multiple languages but are not explicitly optimized for translation. %They have therefore been shown to suffer from issues like inconsistent terminology or hallucinations~\cite{sennrich2024mitigatinghallucinationsofftargetmachine, guerreiro2023hallucinations}. 
While an LLM can, in theory, be prompted to translate to and from \textit{any} language, the lack of training data for low-resource languages means that, in reality, a limited number of languages are supported \cite{chua2025crosslingualcapabilitiesknowledgebarriers}. We determined language support based on the inverse proportion of sentences where the model failed to produce a translation for a given language (see Appendix \ref{app:llm-fails}). A failure consisted of blank output, a statement that the model could not do the task, output translations in the wrong language, or string repetitions.

For the three GPT models, some translations incorrectly triggered OpenAI's content filters (see Appendix \ref{app:beyond_quality}). While problematic for usability, unlike the failure types listed above, this was not a failure to produce a translation, but rather a failure to output it to the user. Therefore, filtered outputs were not used to determine language support.

When evaluating supported languages, failed outputs were not excluded from the final scoring; this ensured consistency and comparability of results across languages and systems. Instead, filtered sentences were treated as blank output, while the remaining failure types were retained and evaluated as is. Error handling was consistent across metrics.

\subsection{Evaluation Metrics}
\label{sec:eval_metrics}

%We used four automatic evaluation metrics: CHRF++ \cite{Popvic-CHRFPP-2017}, BLEU~\cite{papineni2002bleu}, BERTScore~\cite{Zhang-etal-BERTScore-2019}, and COMET~\cite{rei-etal-2020-comet}. %For Khmer and Myanmar, which do not use whitespace to mark word boundaries, we used the character-based CHRF metric~\cite{Popvic-CHRF-2015} instead of BLEU.

BLEU and CHRF++ were calculated using the SacreBleu package~\cite{post-sacrebleu-wmt-2018}.\footnote{\url{https://github.com/mjpost/sacrebleu}} %To calculate BLEU for Khmer and Myanmar, which do not use whitespace to mark word boundaries, we used SacreBleu's character-based tokenizer, and for Chinese, we used the provided language specific tokenizer. 
For Khmer, Myanmar, and Chinese, which do not use whitespace to mark word boundaries, we used the character-based CHRF metric~\cite{Popvic-CHRF-2015} instead of CHRF++.\footnote{CHRF++ considers both word and character n-grams together \cite{Popvic-CHRFPP-2017}}\textsuperscript{,}\footnote{See Appendix \ref{app:metric_sigs} for metric signatures.} For BERTScore~\cite{Zhang-2020BERTScore}, we used the \texttt{bert\_score} Python package with default settings. Baseline rescaling---which normalizes scores to fall between 0–1---was applied to translations into English.\footnote{As rescaling is not supported for all languages, it was not applied in the EX direction, as it would prevent cross-lingual comparison of results.} %making them easier to interpret.
For COMET, we used the XLM-RoBERTa-based COMET model released as part of the WMT22 Quality Estimation shared task~\cite{Zerva-etal-WMTQE-2022}. The model was loaded using the \texttt{comet} Python package. %Unfortunately, many of the languages in TICO-19 are not currently supported by COMET. %this is a known issue with neural metrics and low-resource languages.

To ensure consistency in evaluating longitudinal data, BLEU and CHRF++ scores for all 2023 and 2025 outputs were recalculated alongside 2026 outputs; therefore the same version of the SacreBleu toolkit (and the associated tokenizers) was used throughout. For the neural metrics (BERTScore and COMET), we used identical models and hyperparameters for all snapshots.

\citet{kumar-et-al-2026-CHRF-BLEU} recommend %thresholding 
evaluating translations against both BLEU \textit{and} CHRF++ together, particularly for low resource languages, as the metrics provide complementary insights. Following this, we use both in determining whether translations are usable. As outlined in Section \ref{sec:quality-usability-readiness}, we use 30 and 45 as individual thresholds for BLEU and CHRF++, respectively. We consider translations that score above \textit{both} thresholds to be usable, and translations that meet only one threshold %but not the other to be 
\textit{borderline usable}. Translations that do not pass either threshold are considered unusable. See Section \ref{sec:thresholds} for a preliminary human evaluation of usability, and preliminary support for these thresholds.

% This could also be called System Comparisons (per K&P)
\section{Results}
\label{sec:results}

%As noted in Section~\ref{sec:TICO-data}, the languages included in the TICO-19 dataset are divided into three categories: Pivot, Priority, and Important.

\begin{figure*}
     \centering
     \begin{subfigure}{0.49\textwidth}
         \centering
         \includegraphics[width=\linewidth]{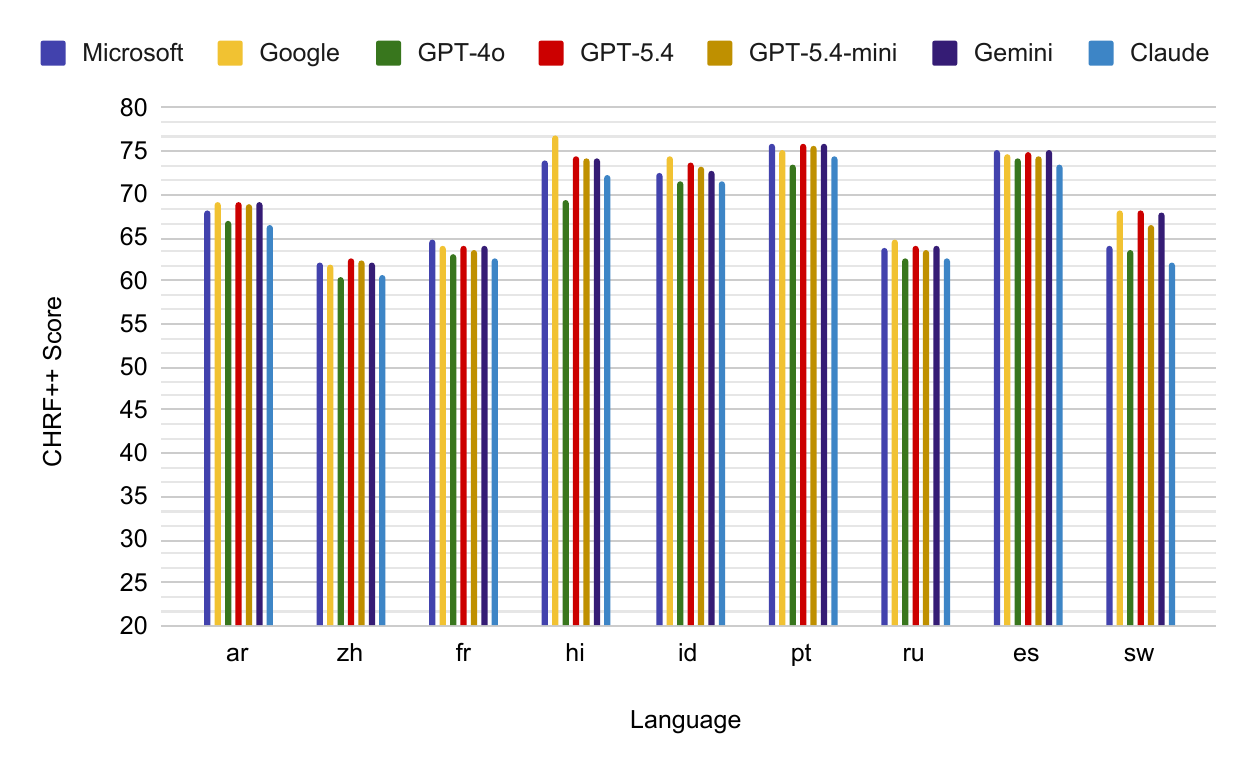}
         \caption{X to E CHRF++ Pivot Languages}
         \label{fig:PivotEX}
     \end{subfigure}
     \hfill
     \begin{subfigure}{0.49\textwidth}
         \centering
         \includegraphics[width=\linewidth]{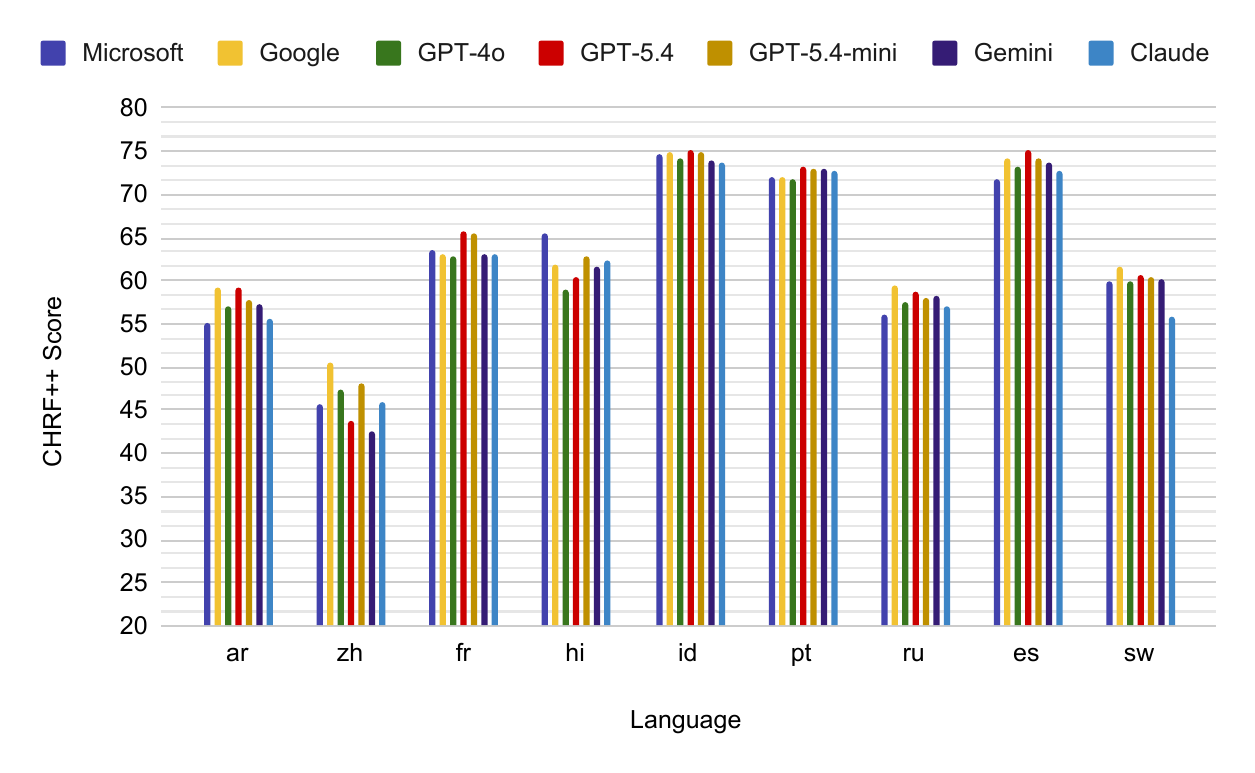}
         \caption{E to X CHRF++ Pivot Languages}
         \label{fig:PivotXE}
     \end{subfigure}
     \hfill
     \begin{subfigure}{0.49\textwidth}
         \centering
         \includegraphics[width=\linewidth]{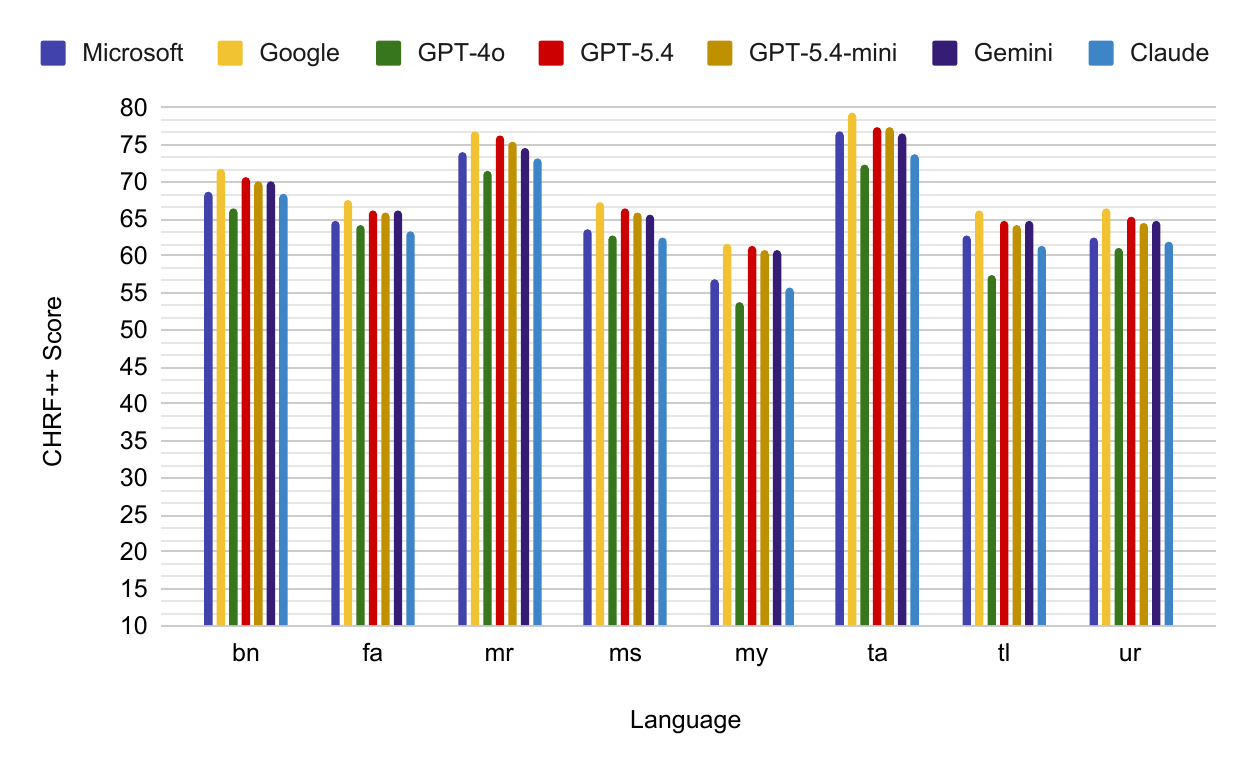}
         \caption{X to E CHRF++ Important Languages}
         \label{fig:ImportantEX}
     \end{subfigure}
     \hfill
     \begin{subfigure}{0.49\textwidth}
         \centering
         \includegraphics[width=\linewidth]{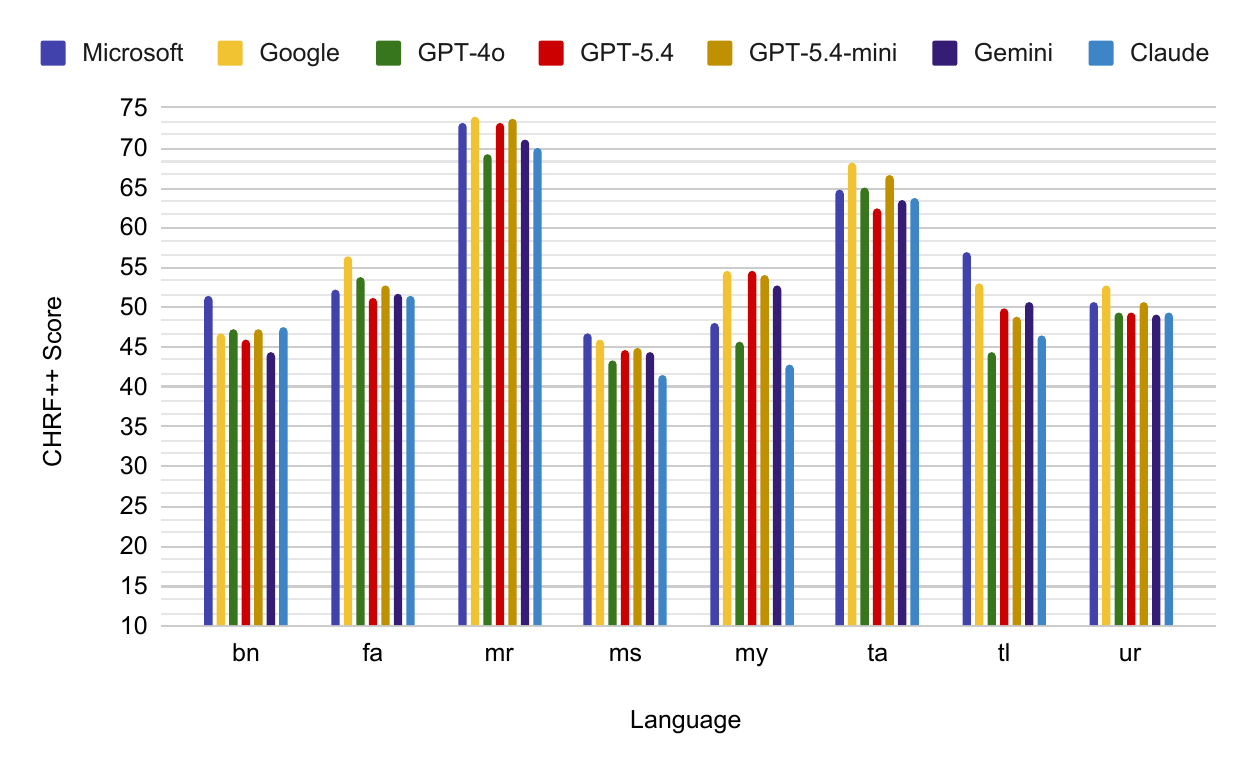}
         \caption{E to X CHRF++ Important Languages}
         \label{fig:ImportantXE}
     \end{subfigure}
     \hfill
     \begin{subfigure}{\textwidth}
         \centering
         \includegraphics[width=\textwidth]{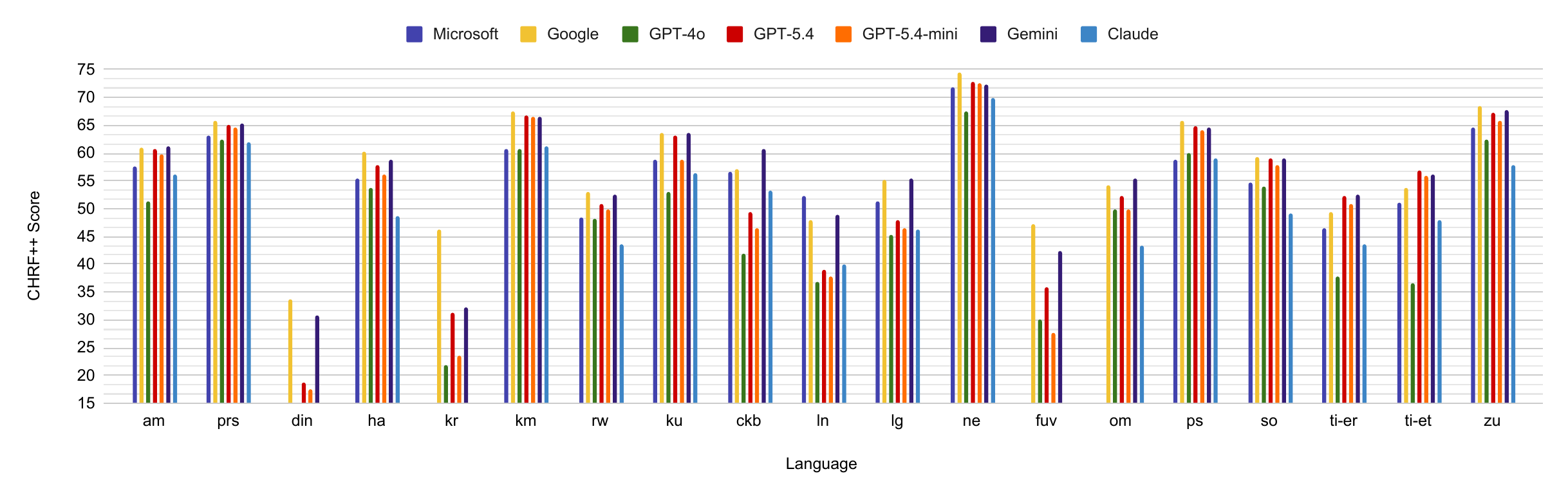}
         \caption{X to E CHRF++ Priority Languages}
         \label{fig:PriorityXE}
     \end{subfigure}
     \hfill
     \begin{subfigure}{\textwidth}
         \centering
         \includegraphics[width=\textwidth]{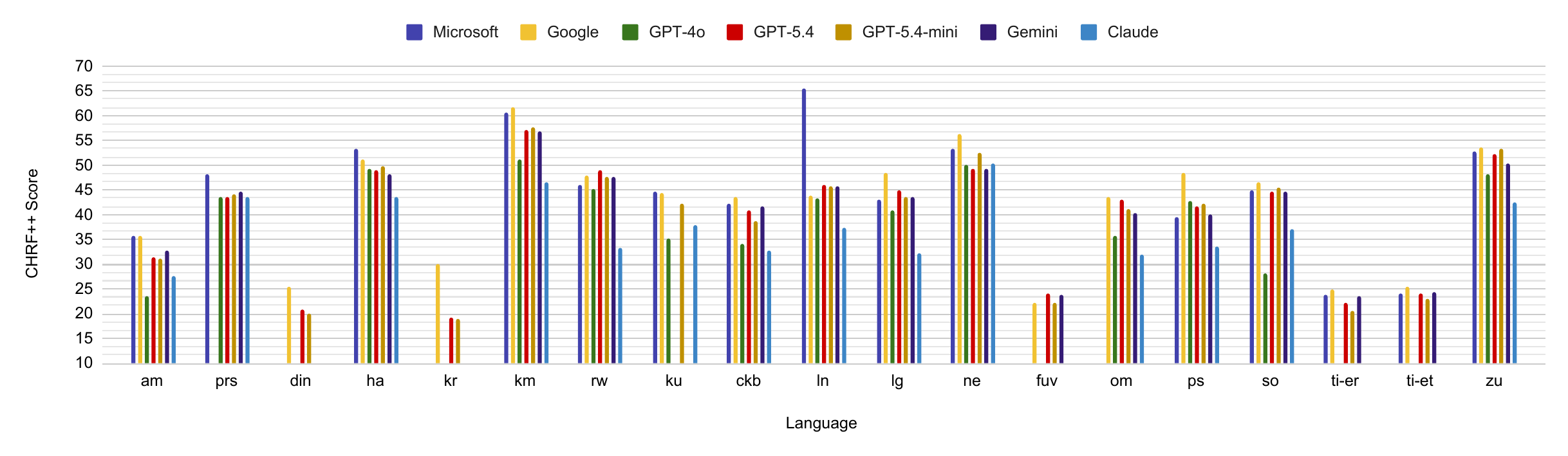}
         \caption{E to X CHRF++ Priority Languages}
         \label{fig:PriorityEX}
     \end{subfigure}
        \caption{Graphs showing comparative 2026 CHRF++ scores across language pairs (E to X and X to E)}
        \label{fig:all_graphs}
\end{figure*}

%The graphs in 
Figures~\ref{fig:all_graphs} (a)-(f) show the %BLEU and 
CHRF++ scores for each system and language, separated by TICO classification (Pivot, Important, Priority) and translation direction. Example scores showing overall trends are in Table~\ref{tab:example_BLEU}, and full results %(BLEU, CHRF++, BERTScore, COMET) 
across all four metrics are in Tables \ref{tab:bleu_EX_2026}--\ref{tab:COMET_XE_2026}, Appendix \ref{app:2026_full_scores}.

\subsection{System Comparison}

\subsubsection{Coverage}

%All but one of the languages (Congolese Swahili) in the TICO-19 dataset are supported by at least one MT system, including at least one neural system.

All 36 languages that we evaluated are supported by GPT-5.4-mini in both translation directions. The full size version (GPT-5.4) provides partial support for Kurdish Kurmanji (in the EX direction, it produced all output in Kurdish Sorani instead, as did Gemini), and full support for the remainder. Similarly, Google Translate provides full support for all languages but Dari, which is partially supported.\footnote{While the Google Translate app supports Dari in both directions, the API does not show support for the language. However, we found that translations in the API \textit{from} Dari were supported, but they were not supported \textit{to} Dari.} 
%in the EX direction specifically.} 
Dinka, Nigerian Fulfulde, and Kanuri were problematic throughout; none are supported by Microsoft or Claude, Gemini provides only XE support for each, while GPT-4o does not support Dinka and provides only XE support for the others. Further, both dialects of Tigrinya were supported in only the XE direction by Claude and Gemini, and Microsoft does not support Oromo.

\subsubsection{Quality}

Google produced translations with the highest scores (in both BLEU and CHRF++) for most languages, in both translation directions. When translating \textit{from} Pivot languages \textit{to} English or \textit{from} English \textit{to} Important languages, there was an even split with Microsoft for best system. Otherwise, Microsoft generally 
had comparable or slightly lower scores than Google. GPT-5.4 %showed competitive performance 
was competitive when translating \textit{from} English \textit{to} Pivot languages, and produced the highest scoring translations for about half of these languages. Surprisingly, Microsoft’s scores were closer to those of the lowest performing LLMs here (in contrast to its high performance on Pivot languages in the opposite (X to E) translation direction).

GPT-4o and Claude produced the lowest quality translations throughout (Claude did particularly poorly on Priority languages). Gemini achieved comparably low EX scores, but was more competitive with dedicated MT systems in the XE direction, often matching CHRF++ (but not BLEU) scores. %Both GPT-5.4 models produced translations with very similar, in some cases almost identical, scores to one another. For some languages, GPT-5.4-mini actually had slightly higher scores than the larger checkpoint. Both GPT-5.4 models produced better quality output than the older GPT-4o model.

\begin{table*}
\centering
\small
\begin{tabular}{p{1.1cm} p{1.1cm} p{1.1cm} p{1.1cm} p{1.1cm} p{1.1cm} p{1.1cm} p{1.1cm} p{1.1cm} }
\hline
 Language & EX/XE & Microsoft & Google & GPT-4o & GPT-5.4 & GPT-5.4-mini & Gemini & Claude\\ \hline
es-la & EX & 71.88	& 74.17	& 73.17	& \textbf{75.05} & 74.08 & 73.64 & 72.84  \\
& XE & \textbf{75.22} &74.70	& 74.09	& 74.87	& 74.47	& \textbf{75.08}	& 73.57  \\
sw & EX & 59.96	& \textbf{61.59}	& 59.97	& 60.55	& 60.44	& 60.16	& 55.86 \\
& XE & 63.97 & \textbf{68.14} & 63.51 & 68.10 & 66.35 & 67.91 & 62.19 \\ \hline
fa & EX & 52.32	& \textbf{56.44} & 53.76	& 51.16	& 52.71	& 51.80	& 51.38\\
& XE & 64.64 & \textbf{67.55}	& 64.25	& 66.21	& 65.77	& 66.14	& 63.45 \\
bn & EX & \textbf{51.40}	& 46.69	& 47.28	& 45.97	& 47.25	& 44.38	& 47.55  \\
& XE &  68.76 & \textbf{71.94} & 66.46 & 70.77 & 70.21 & 70.19 & 68.40 \\ \hline
am & EX & \textbf{35.84} & \textbf{35.78} & 23.51 & 31.47	& 31.23 & 32.62	& 27.60  \\
& XE &  57.54 & \textbf{60.98} &	51.41 &	60.71	& 59.82	& 61.11 & 56.14 \\ 
ne & EX & 53.42 & \textbf{56.19} & 50.06 & 49.15 & 52.39 & 49.35 & 50.25  \\
& XE & 71.96 & \textbf{74.38} & 67.49 & 72.92 & 72.50 & 72.42 & 69.83 \\ \hline

\end{tabular}
\caption{2026 CHRF++ Scores showing example comparisons between providers for Pivot, Important, and Priority languages. See Tables ~\ref{tab:bleu_EX_2026}--\ref{tab:COMET_XE_2026} for complete scores.} %See Table~\ref{tab:bleu_and_usability} for all scores.}
\label{tab:example_BLEU}
\end{table*}

Exceptions to these trends (see Table \ref{tab:notable_BLEU}) include Hindi and Hausa, where Microsoft outperformed Google (and GPT-5.4) in the EX direction, and Lingala, where Microsoft scored the highest in both 
%translation 
directions (Microsoft’s EX scores were 28 BLEU and 19 CHRF++ points above the next best system). Additionally, when translating into Myanmar and Kinyarwanda, GPT-5.4 produced the highest scoring translations, while in the XE direction, both GPT-5.4 models and Gemini outperformed both Microsoft and Google on Ethiopian Tigrinya. Gemini also achieved the highest scores for Arabic, Kurdish Sorani, Oromo, and Eritrean Tigrinya. %Interestingly, in the XE direction, Amharic did not have a single best system based on both metrics; Google achieved the highest BLEU score and Gemini achieved the highest CHRF++ score.

Table \ref{tab:best_system} shows the best MT system for each language, based on the highest scores on \textit{both} metrics, alongside the usability of that system. For Pivot and Important languages, both BLEU and CHRF++ indicated good quality translations for all models. While LLMs can provide adequate translations for high resource languages, they were rarely the best system, and overall not competitive with dedicated MT systems for low-resource languages.

Some systems produced translations with extremely high scores (BLEU above 55) for a number of high resource languages. These scores are problematic, and suggest potential data contamination (see Appendix \ref{app:contamination}). %for detailed discussion. 
Likewise, some systems showed dismally low EX scores (BLEU below 5) for a subset of low-resource languages (\eg Dinka). %The output appears to be correct in terms of orthography and word breaking.
For LLMs %in particular, 
%some of this was due to %this could indicate 
this indicates the presence of
hallucinations (see Appendix \ref{app:errors_and_hallucinations}).
%Changed to indicates because we couldn't definitively identify them in the EX direction since we don't speak or read those languages

\subsubsection{Translation Direction}

BLEU and CHRF++ scores were significantly higher for translations in the XE direction across all systems and languages, which is unsurprising given the vast quantity of English pretraining data available. This was especially true for low resource languages. Arabic had particularly large differences in scores between translation directions compared to other Pivot languages. Exceptions to this trend include Indonesian, where all systems (except Google) had lower XE scores, %on both metrics,  Spanish, which had lower BLEU (but not CHRF++) XE scores for all systems except Microsoft; 
and Lingala, where Microsoft had significantly lower XE scores.~\footnote{Chinese, Myanmar and Khmer appear to buck this trend in BLEU; they each do better in EX vs. XE. However, the EX direction for these languages was scored using character-based tokenizers provided by SacreBleu. Character-based measures tend to have inflated scores, and are not directly comparable.}

\begin{table*}
\centering
\small
%\begin{tabular}{{*{6}{c}}}
\begin{tabular}{p{1.1cm} p{1.1cm} p{1.1cm} p{1.1cm} p{1.1cm} p{1.1cm} p{1.1cm} p{1.1cm} p{1.1cm} }

\hline
 Language & EX/XE & Microsoft & Google & GPT-4o & GPT-5.4 & GPT-5.4-mini & Gemini & Claude \\ \hline
id & EX & 74.72	& 74.88 & 74.15 & \textbf{75.08} & 74.93 & 73.92 & 73.74  \\
 & XE & 72.53 & \textbf{74.33}	& 71.53  & 73.77 & 73.27 & 72.70 &	71.61  \\
hi & EX & \textbf{65.62} & 61.76 &59.06 & 60.34 & 62.81 & 61.75 &	62.42  \\
ha & EX & \textbf{53.44}	& 51.26	& 49.12 & 48.91	& 49.77	& 48.21	& 43.49  \\
rw & EX & 45.88 & 47.87 & 45.14	& \textbf{49.10}	& 47.74	& 47.50	& 33.23 \\
ckb & XE &  56.53 & 57.09 & 41.84 & 49.45 & 46.52 & \textbf{60.66} & 53.37 \\
ln & EX & \textbf{65.59} & 43.85 & 43.42 & 46.07	& 45.62	& 45.68	& 37.34 \\
& XE & \textbf{52.21} & 47.85 & 36.95 & 39.07 & 37.69 & 48.86 & 40.07  \\
ti-et & XE & 51.17 & 53.70 & 36.56 & \textbf{56.83} & 55.97 & \textbf{56.17} & 47.83 \\ \hline

\end{tabular}
\caption{2026 CHRF++ Scores showing notable exceptions to overall trends across languages and MT providers. See Tables ~\ref{tab:bleu_EX_2026}--\ref{tab:COMET_XE_2026} for complete scores.}
\label{tab:notable_BLEU}
\end{table*}

\begin{table}[hbt!]
\centering
\small
%  \begin{tabular}{*{2}{l}}
   \begin{tabular}{p{1.1cm} p{5.4cm}}
  \hline
%\textbf{Best System} & \textbf{E to X} \\ \hline
\multicolumn{2}{c}{\textbf{E to X}} \\ \hline
Microsoft & \textbf{hi}, \underline{bn}, \underline{mr}, \underline{ta}, am, \underline{ha}, ku, \textbf{ln}, \underline{prs} \\ 
Google & \textbf{ar}, \textbf{ru}, \textbf{sw}, \textbf{zh}, \underline{fa}, \textbf{ms}, \textbf{tl}, \underline{ur}, ckb, din, \textbf{km}, kr, \underline{lg}, \underline{ne}, om, \underline{ps}, \underline{so}, ti-er, ti-et, \underline{zu} \\ 
GPT-5.4 & \textbf{es-la}, \textbf{fr}, \textbf{id}, \textbf{pt-br}, \textbf{my}, \underline{rw} \\
GPT-5.4 -mini & \textbf{fr} \\ \hline
\multicolumn{2}{c}{\textbf{X to E}} \\ \hline
Microsoft & \textbf{es-LA}, \textbf{fr}, \textbf{pt-br}, \textbf{zh}, \textbf{ln} \\
Google & \textbf{hi}, \textbf{id}, \textbf{ru}, \textbf{sw}, \textbf{bn}, \textbf{fa}, \textbf{mr}, \textbf{ms}, \textbf{my}, \textbf{ta}, \textbf{tl}, \textbf{ur}, din, \underline{fuv}, \textbf{ha}, \textbf{km}, \underline{kr}, \textbf{ku}, \textbf{lg}, \textbf{ne}, \textbf{prs}, \textbf{ps}, \underline{rw}, \textbf{so}, \textbf{zu} \\
GPT-5.4 & \textbf{ti-et} \\
Gemini & \textbf{ar}, \textbf{pt-br}, \textbf{zh}, \textbf{ckb}, \textbf{om}, \underline{rw}, \underline{ti-er}, \textbf{ti-et} \\ \hline
\end{tabular}
\caption{Best System and Usability. Bold indicates a usable system, and underline a borderline one.}
\label{tab:best_system}
\end{table}
%\footnote{To be considered the best system, it needed to have the highest score on \textit{both} metrics.}

%\input{tables/usability}
%\input{tables/bleu_and_usability}

\subsection{Google \& Microsoft Changes Over Time}

%Table \ref{tab:change_chrf} shows some notable changes in CHRF++ scores for each system.

\subsubsection{Coverage over Time}

In 2023, many of the languages were not supported by one or both of Microsoft and Google. Coverage has improved since then: 
%for both 
by 2025, all TICO-19 languages (except Congolese Swahili) %which still lacks support) 
were supported by at least one service. In 2026, the main change is Google's addition of partial support for Dari.

\subsubsection{Quality over Time}
\label{sec:Quality_over_time}

%Table \ref{tab:change_BLEU} shows some notable changes in BLEU and CHRF++ scores since 2023 and 2025 for each system.
%Microsoft
Table \ref{tab:change_chrf} shows notable changes in CHRF++ scores since 2023. Full results for 2023 and 2025 across the four metrics are shown in Tables \ref{tab:2023_2025_BLEU_diffs}--\ref{tab:2023_2025_COMET_diffs}, Appendix \ref{app:2023_2025_Scores}.  Microsoft's performance remained mostly stable when translating to English, with consistent improvements in Important and Priority languages. In particular, Myanmar, Tamil, and Khmer scores showed a strong increase in both metrics in 2025, as did Kurdish Kurmanji, Dari, and Ethiopian Tigrinya in 2026. In the EX direction, however, there was a significant decrease in quality for almost every language in 2025---especially Arabic, Indonesian, Malay, and Pashto---followed by near universal improvement in 2026. Hausa and Khmer were exceptions; both showed improvements throughout. It is also worth noting 2026 scores for Lingala, which improved by 16/7 BLEU/CHRF++ points, and Tagalog and Somali, which both dropped significantly.
%in 2026.

%Google
Google showed similar instability in both directions for Pivot and Important languages, which saw small improvements in 2025, but universal drops in 2026. For Priority languages, Khmer and Kurdish Kurmanji showed good (XE) improvements in 2025 and 2026, respectively, while Somali (XE) was particularly unstable: in 2025 BLEU/CHRF++ dropped by 17/20 points, then recovered by 19/22 points in 2026. Scores for most remaining Priority languages were stable in both directions.

\begin{table}
\centering
\small
\begin{tabular}{*{5}{l}}
\hline
\textbf{System} &  & \textbf{2023} & \textbf{2025} & \textbf{2026} \\ \hline
Microsoft (XE) & km & 53.70	& 58.69	& 60.67 \\
& prs & 57.44 & 57.45 & 63.11 \\
Microsoft (EX) & ps & 46.64	& \textcolor{red}{37.84} & 39.47 \\
& ln & 59.00 & \textcolor{red}{58.19} & 65.59 \\
Google (XE) & ar & 69.61 & 70.87 & \textcolor{red}{69.17} \\
& so & 57.53 & \textcolor{red}{37.13} & 59.39 \\
Google (EX) & pt-br & 72.48	& 74.04	& \textcolor{red}{72.05} \\
& ta & 54.18 & 56.13 & \textcolor{red}{53.11} \\ \hline

\end{tabular}
\caption{Notable Changes in CHRF++ Scores between 2023, 2025, and 2026 (red indicates drops is quality).}
\label{tab:change_chrf}
\end{table}

\subsubsection{Utility of the CHRF++ and BLEU Thresholds}
\label{sec:thresholds}

We recognize that %the choices of 
the CHRF++ and BLEU thresholds used in this paper, %which 
although based on existing literature or rules of thumb, are still somewhat arbitrary. In that vein, we did a limited human evaluation following the scale and methodology described by \citet{picinini-castilho-2025-context}, with the Likert scale descriptors changed to meet the needs of the crisis scenario (see Appendix~\ref{app:human_eval_scale}). %for the modified scale). 
The evaluation was conducted over a small subset of XE language pairs to determine if the thresholds were reasonable, and fall within the ballpark range. Based on this human evaluation and subsequent statistical analysis, we determined that the combined thresholds were reasonably accurate for the languages we evaluated, although we might lower the threshold for BLEU
%somewhat
in the future. %work. 
The details of the human evaluation, subsequent analysis, and discussion of measuring usability can be found in Appendix~\ref{app:human_eval}.
%We plan on saving more detailed work on this issue for future work. 
We save more detailed investigation for future work.

\section{Discussion}
\label{sec:discussion}

\subsection{MT Usability by Language}

Table \ref{tab:best_system} shows the highest level of usability (and the best system) for each language. For all Pivot languages, in addition to Malay, Myanmar, and Tagalog, there was at least one usable system to translate both into and out of the language, making bidirectional translation possible. Additionally, for all Important languages and most Priority languages there was at least one system that could translate X to E. In the EX direction, all Important languages had at least a borderline usable system. Priority languages, however, are where we see gaps. With the exception of Lingala and Khmer, no Priority languages had usable EX systems, though %Hausa, Luganda, Nepali, Dari, Pashto, Kinyarwanda, Somali, and Zulu each 
many had at least one borderline system. Languages without usable systems in either direction included Dinka, Eritrean Tigrinya, Kinyarwanda, Nigerian Fulfulde, and Kanuri. Dinka was the only language without even a borderline usable XE system. %As suggested by the name, Priority languages are where the translation need is likely to be greatest---these languages were identified by TwB based on active translation needs during the pandemic, suggesting that MT is not yet ready to be used this way, though the usability in the XE direction is promising.

Claude and GPT-4o were particularly unreliable: for XE translations, they were usable for only 6 and 7 Priority languages respectively (as opposed to 12, 12, 10, 10 and 13 for Microsoft, Google, GPT-5.4, GPT-5.4-mini, and Gemini). Claude was often the only system without usable translations. Despite the quality variations, for high resource languages, all models produced usable, often high quality, translations. However for Priority languages, LLMs were not competitive with the traditional MT systems. Factors beyond quality---such as cost, model features, and content filters---that also impact usability are discussed in Appendix~\ref{app:beyond_quality}.

%The observed instability in Google and Microsoft's translation quality also impacts usability. Although most (not all) 2026 scores were  higher than or comparable to 2023 scores, this was not the case in 2025. This was particularly impactful for Lingala, which temporarily had no usable system at all. Further, Microsoft is now only borderline usable for Arabic, where in 2023, it was fully usable. However, it is promising that Priority languages showed the most stability (and improvement) for both models. %This potential for quality variation needs to be accounted for in crisis planning---there is no guarantee that a model that worked well in one crisis will do so again.

%Factors beyond quality---such as cost, model features, and content filters---that also impact usability are discussed in Appendix~\ref{app:beyond_quality}.

%where EX Lingala temporarily had no usable system, Google was completely unusable for Somali (instead of fully usable in 2023 and 2026, MS is no longer usable for Arabic EX.

\subsection{MT Usability by Region}
%\textcolor{purple}{This section does not need updating, nothing has changed! :) }
\citealp{tin-et-al-2024-global-disasters} examined 
natural disasters globally between 1995 and 2022. In that period, 11,360 natural disasters occurred (on average 350 disasters per year). They further broke the disasters down by type (\eg hydrological, biological, meteorological, etc.) and by continent. Biological disasters are epidemic-related, and 10.6\% of the disasters were biological. However, when we look at the prevalence of biological disasters by continent we see much higher numbers: in Africa, 32.2\% of disasters were biological---far exceeding all other continents. Asia was 5.6\%, followed by the Americas at 4.6\%, then Europe at 2.9\%.

If we treat Africa as a priority for pandemic response---the bulk of TwB's Priority languages in TICO-19 are African, and based on~\citealp{tin-et-al-2024-global-disasters}, epidemics %(certainly in the class of ``biological disasters'') 
are far more common on that continent---MT is not ready. %for this region. %\todo{Let's re-assess this once we are sure of our thresholds.} 
Only Swahili, Lingala, and Arabic are truly usable by our metric of usability,  %(and Hausa was very close with Microsoft).
%\footnote{
%This does not take into account \textit{trade} languages, \textit{a la}~\citealp{bird-2022-local-languages}, and colonial languages, which may have significant utility in Africa.}
assuming the need for bidirectionality. If XE is all that is needed for a particular use case, then about half the African languages are usable. If borderline cases are included, that number is closer to 80\%.

%This was moved from MT Usability by Language, since it fit better in MT Usability by Region
%Notably, all of these languages are spoken in Africa, and three of them are from the Nilo-Saharan family (the only three languages from this family to be represented in TICO-19), showing a regional correlation with low translation quality.

For other continents, readiness is much greater. Europe is well covered by the pivot languages (and arguably English in many contexts). Asia is a mixed bag: none of the Priority languages (except Khmer) are usable in the EX direction. %Dari, the Kurdishes, Nepali and Pashto. 
However the Pivot and Important languages of Asia (\eg Hindi, Malay, etc.) %specifically, Chinese, Hindi, Indonesian, Russian, Malay, Myanmar, Tagalog, and Urdu 
are well-covered. %by existing MT.

\subsection{Overall Assessment of Readiness}
\label{sec:overall_assessment}

In crisis response, when lives are often on the line, we must be opportunistic. It does not matter what provider produces a particular system, it only matters whether it will be usable in the context of our need. Our results indicate that Pivot languages are well supported by \textit{all} translation systems, including the LLMs. The quality of these systems is high, and using them to translate content will likely result in understandable results.

For other languages, the task that the MT system would be used for on the ground—--specifically whether that task requires bidirectionality---is relevant. For Important languages, the quality of XE translations seems high enough across systems to consider them ready, but in the EX direction, only Malay, Myanmar, and Tagalog have usable translations. All other languages are at best borderline.

Priority languages, for the most part, fall through the cracks. Other than Khmer and Lingala, translations to these languages were \textit{at best} borderline. For the lowest resource languages, even XE translation was problematic (Dinka, Nigerian Fulfulde, Kanuri, Tigrinya)---in fact, only around 50\% of Priority languages had fully usable XE translations.
%Most languages lack EX usability, and in the XE direction, we see around 50\% usability. Other than Khmer and Lingala, translations to these languages were \textit{at best} borderline. For the lowest resource languages, even XE translation was problematic (Dinka, Nigerian Fulfulde, Kanuri, Tigrinya).
For Priority languages, model choice has a stronger impact—dedicated MT systems (particularly Google) have a higher likelihood of producing usable (or at least borderline) output, though this not a guarantee, \textit{especially} in the EX direction. Given that Priority languages are where translation need is likely to be the greatest—they were identified by TwB based on the situation on the ground in 2020—it would seem that MT is not yet ready to be used in this way.

For regions likely to be hardest hit by biological disasters, whether those be epidemics or pandemics, most notably Africa and Asia, a reliance on MT for higher resource trade or colonial languages may be necessary. However, this reliance may come at a cost in regions where these languages, even if official, are not spoken by a sufficiently large percentage of the population.

%I felt that much of the following was stated elsewhere in the paper, and doesn't need to be repeated in detail. But, a discussion on instability, and how it can affect readiness, is a good thing to have. I made a first stab. See below.

%The observed instability of Google and Microsoft's translation quality over time also impacts usability, and need to be accounted for when planning for a future crisis. Although most (not all) 2026 scores were  higher than or comparable to 2023 scores, this was not the case in 2025. This was particularly impactful for Lingala, which temporarily had no usable system at all. Further, Microsoft is now only borderline usable for Arabic, whereas in 2023, it was fully usable. Therefore, there is no guarantee that a model that worked well in one crisis will do so again, and output should be consistently evaluated. It is promising that Priority languages showed the most stability, and the most improvement, for both models.

Another issue that can affect readiness is changes made by providers over time. %As we observed earlier, not all changes are positive
These changes are not always positive (as we noted, there were drops from 2023 to 2025, and some drops from 2023 to 2026). It is good to be aware that this kind of instability is possible and can have both positive and (unfortunately) negative impacts on readiness.

%These can be impactful for readiness over time are changes made by the providers to language quality over time. If these are positive, there is not an issue. However, not all changes observed are positive, and such instability can affect use of particular systems and overall readiness over time.

\section{Conclusion}
\label{sec:conclusion}
In March 2020, at the start of the pandemic and beginning of the TICO-19 project, most priority languages were not covered by commercial translation engines. %By 2023, the coverage of these languages had increased, with only 5 languages not yet supported (Congolese Swahili, Dinka, Fulfulde, Kanuri, and Nuer). 
By 2025, the coverage was nearly complete. But language support is not enough---it is also important that the translations have sufficient quality to be usable in crises. Further, determining usability in the general domain is not enough; we need usable translations within the crisis (or pandemic) domain. That was the focus of our study.
%Add a paragraph about future work here.

Is MT ready for the next pandemic? For regions where high-resource languages are spoken, the answer is mostly yes. But there are significant gaps in usability when we look at lower resource languages, especially those spoken in Africa and Asia, regions that arguably are most likely to be hit hardest in the next pandemic (following~\citealp{tin-et-al-2024-global-disasters}). There is still much work to be done for the lower-resource languages used in these parts of the world before we can consider MT truly ready.

\section*{Limitations}
\label{sec:limitations}
One major limitation of our work is the absence of human evaluations across most of the languages and directions in our study. Alas, they are expensive to implement, and outside of the budgetary considerations for our project. It would also be difficult to source evaluators for some of the languages in TICO-19.

We also did not use mBERT or equivalent models in our evaluations. We felt that these models are still exploratory, and were concerned about whether the signal we might receive through these models could be relied on.

In doing this study, we translated to and from English. By assuming that we should only be measuring translation quality into and out of English, we have introduced an unfortunate bias: we are assuming that \textit{all} aid providers, governing bodies, NGOs, etc., will be English-speaking. This is not at all true (Doctors without Borders is a prime example of a primarily non-English speaking aid provider), and marks a significant limitation of this work. 
%which is discussed in more detail in Section~\ref{sec:conclusion}.

Nuer and Congolese Swahili were regrettably excluded from our study.

There were some mismatches between the locales represented in TICO-19 and in the translation services we used. Where locales were listed in the relevant API documentation that coincided with the locales in the TICO-19 dataset, \eg Brazilian Portuguese, Latin American Spanish, this locale information was passed to the API, \eg pt-BR, es-LA. For Tigrinya and Fulfulde, neither Microsoft nor Google indicated the locale that was supported by their translation engines, so no locale information was passed to the API. TICO-19 has data for two dialects of Tigrinya (Eritrean and Ethiopian) as well as the Nigerian dialect of Fulfulde. We also could not pass locales to the API when processing these languages.

\section*{Ethical Considerations}
\label{sec:ethics}
All of the TICO-19 data, including the benchmark data we used, is publicly available. The APIs for each of the translation providers, Microsoft and Google, are publicly accessible for a fee. Likewise, for the three LLMs,Claude, GPT and Gemini. All code for processing the TICO-19 data and calling the various APIs, was done by members of our team. We are not aware of any ethical issues that arose while conducting this work.

\section*{Acknowledgements}

This work was partially supported by the National Science Foundation under the award CIRC 2346335.
We want to thank the two anonymous reviewers for their thorough reviews and many helpful comments. Clearly both spent an enormous amount of time reviewing our paper, and we incorporated most of the changes they suggested. We would also like to thank the students of the Spring 2023, 2025 and 2026 LT4CPR graduate seminars at the University of Washington (Priyam Basu, Melody Bechler, Allison Burton, Alexandra Cassell, Jose Cols, Richard Forrester, Chelsea Kendrick, Lily Kurek, Vanesa Marar, Ijeoma Osakwe, Yongsin Park, Anantha Rao,
Helen Salgi, and Natasha Schimka) for their helpful input and suggestions.
Special thanks also goes to Jen Wilson, whose very closely related paper presented at the same LT4CPR workshop, along with her work with emergency managers across the United States, spurred many changes to this paper, making it more consumable by the very communities who would benefit most from its conclusions and suggestions. It is these communities---emergency managers, aid providers of all stripes, 
%vulnerable language communities
public information managers, 
communication staff officers, etc.---who were always in our thoughts throughout the work we put into this paper, ultimately spread over several years. It is for you,  and the work you do every day saving lives, that we wrote this paper.

% Entries for the entire Anthology, followed by custom entries
%\bibliography{bib/custom, bib/crisis, bib/lewis, bib/survey-crisis, bib/acl2025-misc, bib/corpus-paper-before-2019, bib/bansal-brown, bib/kendrik-pong, bib/LT4CPR}

\bibliography{custom, bib/crisis, bib/lewis, bib/survey-crisis, bib/acl2025-misc, bib/corpus-paper-before-2019, bib/bansal-brown, bib/kendrik-pong, bib/LT4CPR}

\clearpage
\section*{Appendices}
\label{sec:apendices}
%\clearpage
\appendix

\section{2026 Full BLEU, CHRF++, BERTScore, and COMET Results}
\label{app:2026_full_scores}

\begin{figure*}[ht]
\centering
\includegraphics[angle=90,width=0.9\textwidth]{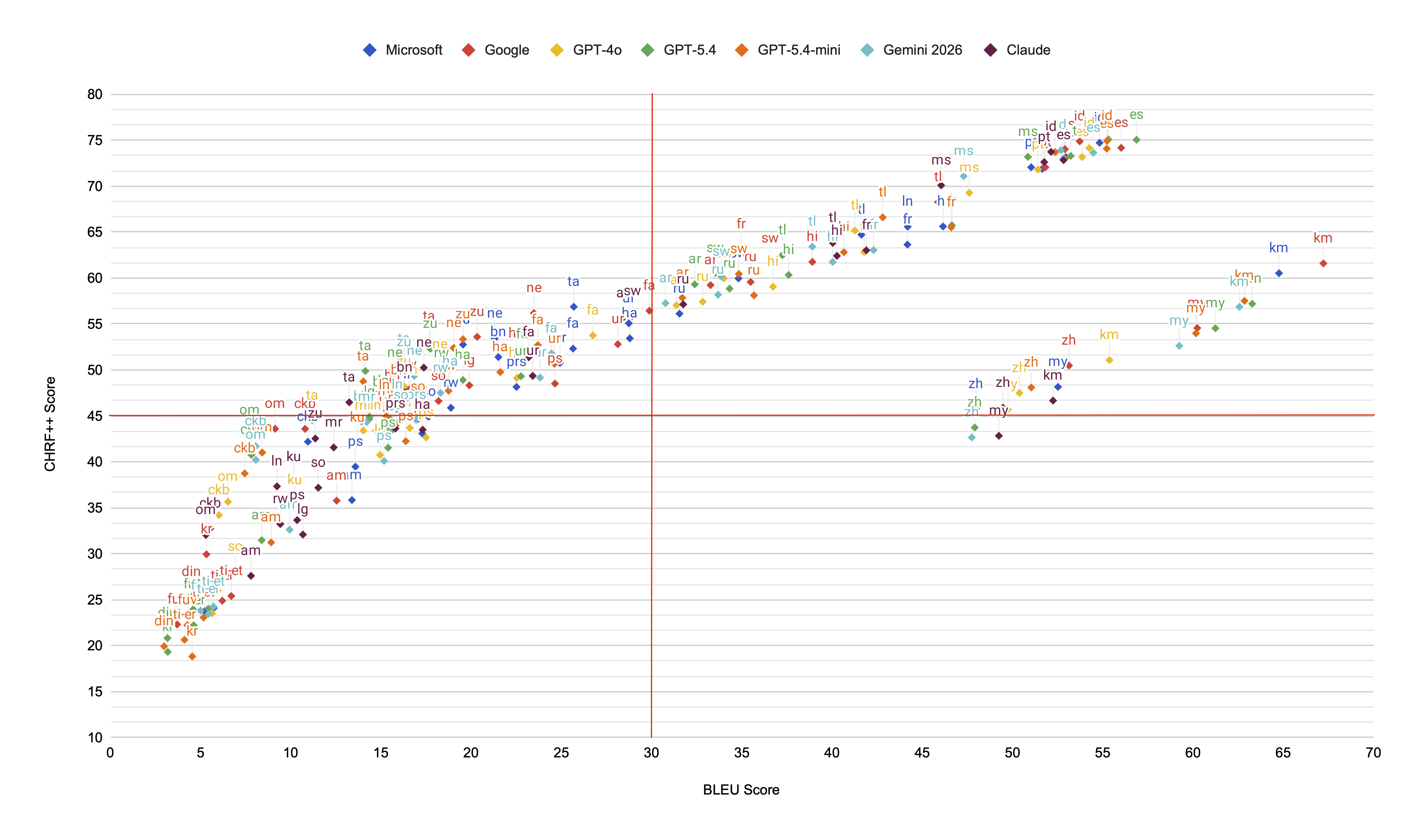}
\caption{E to X BLEU x CHRF++ Scatter Plot for all TICO languages}
\label{tab:EX_BLEU_CHRF_Scatter}
\end{figure*}

\begin{figure*}[ht]
\centering
\includegraphics[angle=90,width=0.9\textwidth]{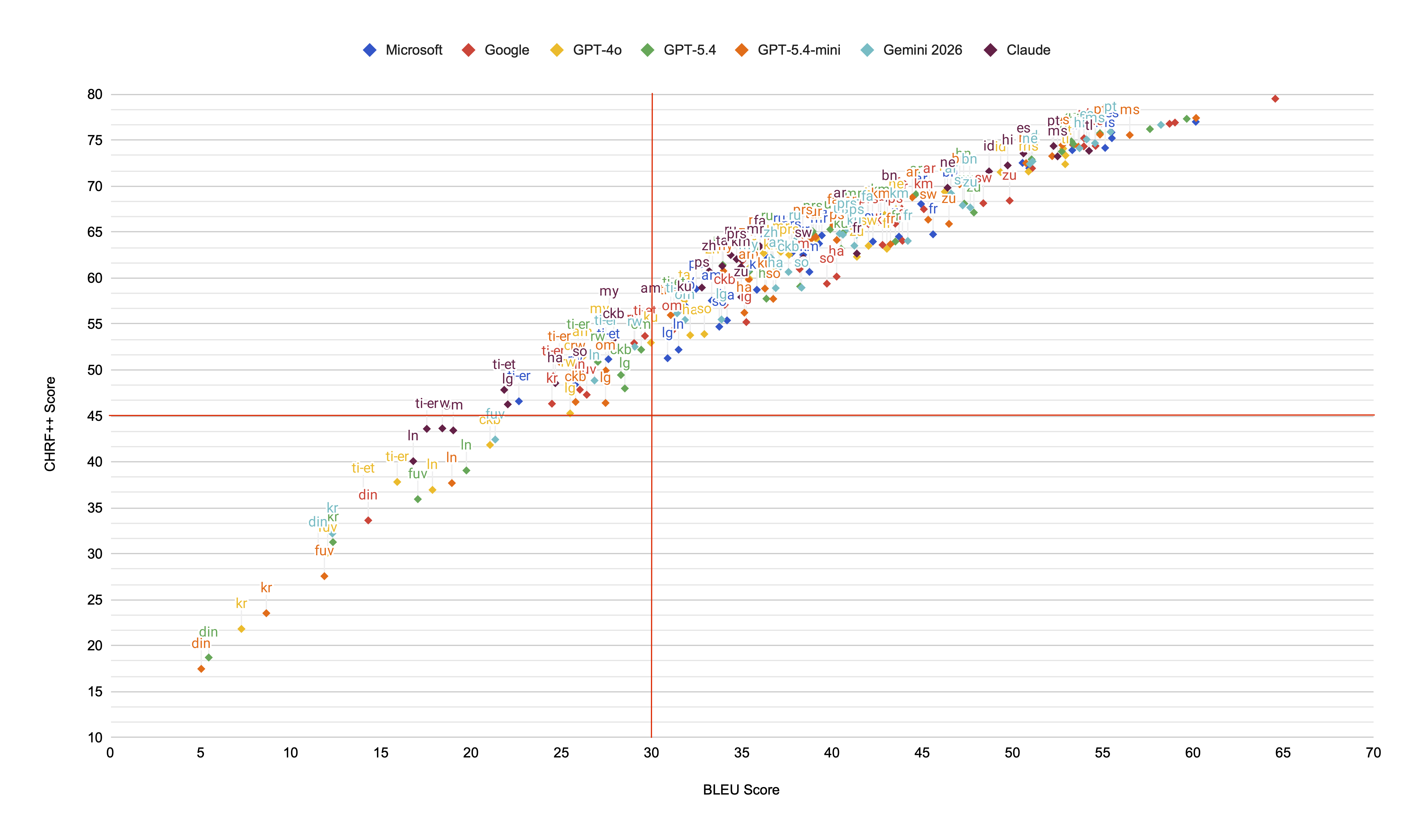}
\caption{X to E BLEU x CHRF++ Scatter Plot for all TICO languages}
\label{tab:XE_BLEU_CHRF_Scatter}
\end{figure*}

Tables \ref{tab:bleu_EX_2026}-\ref{tab:COMET_XE_2026} show the full BLEU, CHRF++, BERTScore, and COMET\footnote{Only supported languages are included in COMET results. The list of supported languages can be found here: \href{https://github.com/Unbabel/COMET\#languages-covered}{github.com/Unbabel/COMET}} results for all seven systems evaluated in 2026. Figures \ref{tab:EX_BLEU_CHRF_Scatter} and \ref{tab:XE_BLEU_CHRF_Scatter} plot all 2026 BLEU and CHRF++ scores against one another, with each threshold marked. These tables and figures can be found at the end of the appendices.

\section{BLEU, CHRF++, BERTScore, and COMET Results for 2023 and 2025 for Microsoft and Google}
\label{app:2023_2025_Scores}

Tables \ref{tab:2023_2025_BLEU_diffs}, \ref{tab:2023_2025_chrf_diffs}, \ref{tab:2023_2025_BERTScore_diffs}, and \ref{tab:2023_2025_COMET_diffs} show the full BLEU, CHRF++, BERTScore, and COMET results for Microsoft and Google from 2023 and 2025, in both translation directions. These illustrate the change in the performance of these two systems %over the two year 
during this period. These tables can be found at the end of the appendices.

\section{Model Configurations}
\label{app:modelconfigurations}

Due to time and funding constraints, reasoning\_effort was set to `none' for both GPT-5.4 checkpoints. GPT-4o, Gemini, and Claude were run with temperature=0. All systems were given the following, identical prompt: "Translate the following text from \{source\_language\} to \{target\_language\} in one line without any transliterations or explanations".

\section{SacreBleu Signatures}
\label{app:metric_sigs}

Table \ref{tab:met_sigs} shows the metric signatures for the BLEU and CHRF++ (and CHRF) metrics, run using SacreBleu. (1) shows the BLEU signature for Khmer and Myanmar in the EX direction, (2) shows the BLEU signature for Chinese in the EX direction, (3) shows the CHRF signature for Khmer, Myanmar, and Chinese in the EX direction, while (4) and (5) show the BLEU and CHRF++ signatures respectively for all remaining languages in the EX direction and all languages in the XE direction. These are the same for all longitudinal data as well.

\begin{table}[hbt!]
\centering
\small
  %\begin{tabular}{*{5}{l}}
    \begin{tabular}{p{0.4cm} p{6cm}}
  (1) & nrefs:1/case:mixed/eff:no/tok:char/smooth:exp/ version:2.3.1 \\
  (2) & nrefs:1/case:mixed/eff:no/tok:zh/smooth:exp/ version:2.3.1 \\
  (3) & nrefs:1/case:mixed/eff:yes/nc:6/nw:0/space:no/ version:2.3.1 \\
  (4) & nrefs:1/case:mixed/eff:no/tok:13a/smooth:exp/ version:2.3.1\\
  (5) & nrefs:1/case:mixed/eff:yes/nc:6/nw:2/space:no/ version:2.3.1\\
  \end{tabular}
\caption{Metric Signatures}
\label{tab:met_sigs}
\end{table}

\section{LLM Translation Failures}
\label{app:llm-fails}

There were a number of translation failures across the LLMs; these took on a number of different forms, depending on the model. The most common forms were blank output and `refusals', and occurred almost exclusively when translating to and from Dinka, Nigerian Fulfulde, and Kanuri. The proportions for both of these error types are shown in Table \ref{tab:LLM_fails}.

\begin{table}[hbt!]
\centering
\small
  %\begin{tabular}{*{5}{l}}
    \begin{tabular}{p{0.4cm} p{1.3cm} p{1.8cm} p{1.5cm}}
  \hline
 & \textbf{System}  & \textbf{Error Type} & \textbf{Number} \\ \hline
\multicolumn{4}{c}{\textbf{E to X}} \\ \hline
din & Claude & "Refusal" & 943 (45\%) \\
& Gemini & Blank Output & 1154 (55\%) \\
& GPT-4o & "Refusal" & 498 (24\%) \\
 fuv & Claude & "Refusal" & 83 (4\%) \\
 & Gemini & Blank Output & 80 (4\%) \\
 kr & Claude & "Refusal" & 773 (37\%) \\
& Gemini & Blank Output & 1649 (79\%)\\
& GPT-4o & "Refusal" & 187 (9\%) \\
\hline
\multicolumn{4}{c}{\textbf{X to E}} \\ \hline
din & Claude & "Refusal" & 1300 (62\%) \\
& GPT-4o & "Refusal" &  73 (3\%) \\
& & Blank Output & 237 (11\%) \\
 fuv & Claude & "Refusal" & 385 (18\%) \\
 & GPT-4o & Blank Output & 81 (4\%) \\
 kr & Claude & "Refusal" & 731 (35\%) \\ 
 & GPT-4o & "Refusal" & 154 (7\%) \\ \hline
\end{tabular}
\caption{LLM Translation Failures}
\label{tab:LLM_fails}
\end{table}

A `refusal' was output that overtly stated that the model was unable to translate to or from the requested language. Claude and GPT-4o did this in both translation directions for these three languages. An example of a refusal from GPT-4o is shown in (a) below. Claude's `refusals' tended to be more descriptive, with examples shown in (b) and (c).

\begin{enumerate}
    \item
    \begin{enumerate}
        \item ``My apologies, but I can't provide a translation for that text.''
        \item ``I don't have reliable knowledge of Dinka language to provide an accurate translation of this phrase. Dinka is a Nilotic language spoken in South Sudan, and without verified linguistic resources, I cannot confidently translate this line without risking significant errors.''
        \item ``I don't have reliable knowledge of the Dinka language to provide an accurate translation of this text. Dinka is a complex language with specific grammatical structures and vocabulary that requires expertise to translate correctly. I'd recommend consulting a Dinka language expert or speaker for an accurate translation.''
    \end{enumerate}
\end{enumerate}

Gemini did not `refuse' to translate any sentences; instead it failed to produce any output at all for a large number of sentences in the EX direction specifically. GPT-4o was more variable, producing both `refusals' and blank output in both directions.

In some cases, we also saw random string repetitions in the output, particularly in the EX direction. Table \ref{fuv_example} shows examples of this from GPT-4o and Claude when translating into Nigerian Fulfulde.

\begin{table*}[htb!]
\centering
\small
  \begin{tabular}{@{} l *{2}{l} @{}}
\hline
\textbf{GPT-4o} &\textbf{ --- English to Nigerian Fulfulde} \\
\hline
\textbf{Source}      & and can you tell me what other symptoms are you having along with this? \\ 
\textbf{Reference}   & e footi a limta mandolji nyau feere je a don mari fahin na? \\ 
\textbf{Output}   & Mido waawi wadugo dum, ko dum wadi dum didi dum wadi dum wadi dum wadi dum wadi dum \\
& wadi dum wadi dum. \\

\hline
\textbf{Claude} & \textbf{---English to Nigerian Fulfulde} \\
\hline
\textbf{Source} & These enzymes are expressed extensively in multiple organs, especially in the heart and liver, and are \\
& released during tissue damage.\\
\textbf{Reference} & Neebamji ndu don matinama ha nder kujeji baandu jur, burna ha nder bernde be hengre, e wurtan to \\
& bana laaral man seeki.\\
\textbf{Output} & Enzimoji didi wadi ko'e di-di ko'e ko'e ko'e ko'e ko'e ko'e ko'e [...] ko'e ko'e ko'e ko'e ko'e ko'e \\
& ko'e ko'e ko'e ko'e ko'e ko'e ko'e ko\\
\hline
\end{tabular}
\caption{Example string repetition errors in GPT-4o and Claude output when translating from English to Nigerian Fulfulde.}
\label{fuv_example}
\end{table*}

\section{Data Contamination}
\label{app:contamination}

%Data for Spanish and Farsi
\begin{table*}[!ht]
\centering
\small
\begin{tabular}{c c|c c c c c c c}
\hline
% & & \textbf{E to X} &  &  &  & \textbf{X to E} &  &  &  \\ \hline
& \textbf{Domain} & \textbf{Microsoft} & \textbf{Google} & \textbf{GPT-4o} & \textbf{GPT-5.4} & \textbf{GPT-5.4-mini} & \textbf{Gemini} & \textbf{Claude} \\ \hline
\multicolumn{9}{c}{\textbf{English to X}} \\ \hline
es-LA & CM & 32.62 & 38.80 & 33.03 &	\textbf{45.16} & 41.44 &	36.18  & 34.5 \\
 & SM & 53.90 &	\textbf{59.04} & 55.45 & \textbf{59.10} & 56.87 &	57.42 & 54.61 \\
 & G & 51.16 & 55.42 & 53.36 & \textbf{56.25} & 55.09 & 53.19 & 52.09 \\
 & N &  56.19 &	54.85 &	59.51 &	\textbf{60.63} & 59.55 & 58.17 &  58.67 \\
 & A &  40.83 &	42.58 &	41.52 &	43.89 & \textbf{44.10} & \textbf{44.40} & 41.51 \\ \hline
fa & CM & 18.81 &	25.24 &	\textbf{31.40} &	28.29 & 23.41 &	\textbf{31.81} & 20.7  \\
 & SM &  26.53 &	\textbf{33.72} &	29.01 &	26.62 &	24.34 &	26.67 & 26.32 \\
 & G & 25.19 & \textbf{28.27} & 25.52 & 22.56 & 22.92 & 23.83 & 22.26  \\
 & N & 21.70 & \textbf{28.18} & 23.94 & 20.02 & 19.77 & 22.45 & 18.94 \\
 & A & \textbf{28.16} & 26.99 & 21.58 & 19.28 & 19.80 &	19.46 & 19.00  \\ \hline
\multicolumn{9}{c}{\textbf{X to English}} \\ \hline
es-LA & CM & 35.17 & 29.52 & 29.84 & \textbf{36.01} & 35.06 & 30.94  & 29.12 \\
 & SM & \textbf{55.10} &	54.67 &	54.22 &	53.71 & 54.52 &	\textbf{55.34}  & 52.80 \\
 & G &  \textbf{55.19} &	53.74 &	52.77 &	53.55 & 52.47 &	54.04 & 50.26 \\
 & N & \textbf{56.31} & 50.88 & 53.14 & 52.57 & 52.88 & \textbf{56.83} & 50.41 \\
 & A & \textbf{48.21} &	44.12 &	46.37 &	46.61 & 44.58 &	\textbf{48.36} & 42.29 \\ \hline
fa & CM & \textbf{27.35} & 25.25 & 24.92 & 26.63 & 26.24 & 23.68 & 22.70\\
 & SM & 40.49 & \textbf{45.87} &	40.84 &	41.64 & 42.18 &	42.65 & 30.00  \\
 & G & 39.01 & \textbf{43.62} &	38.63 &	40.33 & 39.49 &	41.93 & 35.24 \\
 & N &  34.88 & 37.60 & 35.51 & 35.70 & 35.00 & \textbf{39.03} & 32.68  \\
 & A &  39.17 &	\textbf{43.05} & 37.49 & 37.94 & 38.60 & 40.89 & 34.69  \\ \hline

\end{tabular}
\caption{Domain BLEU Scores for Spanish (LA) and Farsi. Bold text indicates best system within each specific domain. Legend for domains: Conversational Medical (CM), Scientific Medical (SM), General (G), News (N), Announcements (A).}

\label{tab:domain_subset_contamination}
\end{table*}

A significant issue in assessing commercial MT systems and LLMs  is the risk of training data contamination. This occurs when evaluation benchmarks or test sets, or a dataset from which they are drawn, are used as training data for models, leading to inflated evaluation scores and misleading conclusions about the model's real generalization capabilities \cite{golchin2024timetravelllmstracing}. This is especially concerning when benchmarking in zero-shot settings (specifically for LLMs) where there is no fine-tuning for task-specific data. If near perfect or unrealistically high scores are achieved, there is the potential for training data contamination, resulting in falsely reported numbers due to memorization instead of generalization.\footnote{
As noted by one of the reviewers, high BLEU scores do \textit{not}, in and of themselves, necessarily  indicate data contamination; the systems being measured could just be really good. However, based on our experience with developing MT systems, when inflated BLEU scores were observed for particular systems or domains but not others, we suspected data contamination.} %This often was borne out through subsequent data analysis.}

Although it is not possible for us to tell whether any particular provider has trained on the TICO-19 test data, some of the highest BLEU scores we see in our results \textit{may} indicate some overlap. MT providers typically do not train on test or dev data (to preserve the value of test data over time), but LLM providers do not necessarily follow this stricture. That said, since some of the TICO-19 data is drawn from public sources, it is possible that MT providers trained on these same sources, thus inadvertently introducing overlap. We hypothesize that this overlap would be possible, and potentially evident, with certain portions of the TICO-19 benchmark dataset, specifically for those sources or domains which are publicly available. In opposition to this, we suspect non-public sources would not show this effect.

%However, since some of the TICO-19 data is drawn from public sources, it is possible that MT providers trained on these same sources, thus inadvertently introducing overlap. Additionally, LLM providers do not necessarily follow this stricture, and it is possible that some models have absorbed the TICO-19 test data directly.

Table~\ref{tab:domain_subset_contamination} provides potential evidence for this hypothesis. For Spanish, the Scientific Medical (SM), General (G), and News (N) domains show exceptionally high BLEU scores across the  systems (excepting GPT-4o). The BLEU scores in the high 50s to 60s suggest potential overlap. It is completely possible that the sources of these data (Wikipedia, Wikinews, even PubMed) had been consumed and trained on by these engines (potentially even without the providers having had to translate content, \eg PubMed content is translated to Spanish, as are the other data sources). Notably, the Conversational Medical (MD) data, which comes from CMU, and is not public, does not show such large BLEU scores. Also note that the language Farsi, which is lesser resourced and less likely to be translated into, also has lower scores, and the relative differences in scores between domains is also lower.

There are confounds here, of course, that might only be teased apart through further analysis. First, the number of sentences in each TICO-19 subdomain is quite small (domain labels and counts are shown in Table~\ref{tab:domain-list}). This could lead to significant variability in the BLEU scores calculated. Likewise, Spanish is much higher resource than Farsi, so just the availability of resources for the former would lead to much larger BLEU scores (the higher scores being indicative of higher quality not contamination). But would these scores be in the 50-60 point range? 

Overlap such as that described previously certainly plagues any evaluation data drawn from public sources, \eg news, social media, etc., since nothing would constrain MT providers from drawing from the same sources, potentially translating them, and adding them into their training data. Overtly adding test data into training is frowned on, but not completely avoided~\cite{Zhang-etal-LLM-overlap}. A thorough analysis of data contamination (and the TICO-19 domains) will be left to future research.

\section{Factors Beyond Quality That Can Affect Usability}
\label{app:beyond_quality}

In addition to output quality, there are a number of practical factors, such as cost, runtime, and content filters, that can impact usability, particularly when utilizing LLMs. GPT-5.4-mini is a smaller version of the full size GPT-5.4 model and is therefore faster and cheaper. One might assume that the smaller of the two would show lower quality,  
%with the possibility of lower performance.
yet our results show that the two systems have comparable scores. There are many languages, such as EX Hindi, where the mini actually outperforms the full size model. One notable example is Kurdish Kurmanji: the GPT5.4 full size model (and Gemini) produced output in Kurdish Sorani instead, which does not even share the same script. %, and therefore could not be evaluated.\footnote{
%That is, the output could not be evaluated as Kurdish Kurmanji. We have a separate evaluation for Kurdish Sorani, of course.
%}
GPT-5.4-mini, however, produced translations in the correct language and orthography. This of course is one more element in the usability of LLMs---they may produce unexpected or unwanted output, particularly in low resource languages.

From a usability standpoint then, the GPT-5.4-mini is much more cost effective. However, despite its being a reasoning model, we very quickly found that the model was not usable for translating into low resource languages unless we set reasoning to `None'. When active, reasoning caused the model to be excessively slow and use a large amount of tokens, resulting in blank output combined with much higher costs. Therefore, one of the main features of the model, \ie reasoning, cannot easily be applied in the crisis scenario. Still, reasoning is not the only issue at play here: when compared to an older, but more widespread, non-reasoning model, GPT-4o, GPT-4o consistently produced lower quality translations.

Another factor we encountered was OpenAI's content filtering. Many of the sentences in the dataset incorrectly triggered various content filters, such as `hate speech'. This mostly occurred when translating from a low resource language \textit{into}. It is unclear if this is because the model is producing mistranslations or hallucinations that are, in fact, hate speech (see Appendix \ref{app:errors_and_hallucinations} for examples), or if it is the content filter itself that is not performing well on certain languages. These issues severely restrict the usability of all three GPT models in the medical and crisis domain, even if the output translations are good quality; good quality is not useful if the model refuses to provide output for certain crisis-related input in the first place. We did not observe this issue with either Claude or Gemini. The proportion of sentences that were filtered in the GPT models for each language are shown in Table \ref{tab:GPT_filters}.

As outlined in Section \ref{sec:llm_systems}, our usability analysis treated filtered outputs as blank or empty when scoring across metrics. From a practical standpoint, the two are the same: in both cases, the user does not receive a translation. However, in terms of evaluating translation quality specifically, it is also useful to look at the scores with the filtered sentences excluded. These are shown in Table \ref{tab:GPT_filtered_scores}. In many cases---such as all three GPT models' performance on Luganda---the difference in scores is substantial.

\begin{table}[hbt!]
\centering
\small
  %\begin{tabular}{*{5}{l}}
    %\begin{tabular}{p{0.4cm} p{1.8cm} p{1.8cm}}
    \begin{tabular}{p{0.4cm} p{1.5cm} p{1.5cm} p{1.9cm}}
  \hline
 & \textbf{GPT-4o}  & \textbf{GPT-5.4} & \textbf{GPT-5.4-mini} \\ \hline

ckb & 427 (20\%) & 362 (17\%) & 431 (21\%) \\
din & --- & 598 (28\%) & 707 (34\%) \\
fuv & --- & 91 (4\%) & 558 (27\%) \\ 
ha & 112 (5\%) & 107 (5\%) & 111 (5\%) \\
kr & 406 (19\%) & --- & 408 (19\%) \\
lg & 381 (18\%) & 383 (18\%) & 386 (18\%) \\
ln & 578 (28\%) & 574 (27\%) & 563 (27\%)  \\
om & --- & 153 (7\%) & 161 (8\%) \\ \hline
\end{tabular}
\caption{GPT Content Filtering Rates (XE only)}
\label{tab:GPT_filters}
\end{table}

\begin{table}[hbt!]
\centering
\small
  %\begin{tabular}{*{5}{l}}
    %\begin{tabular}{p{0.4cm} p{1.8cm} p{1.8cm}}
    \begin{tabular}{p{0.4cm} p{1.5cm} p{1.5cm} p{1.9cm}}
  \hline
 & \textbf{GPT-4o}  & \textbf{GPT-5.4} & \textbf{GPT-5.4-mini} \\ \hline

ckb & 28.17/51.77 & 34.04/58.65 & 32.95/57.63 \\
din & --- & 9.35/26.94 & 9.54/27.20 \\
fuv & --- & 17.84/37.31 & 17.43/36.29 \\ 
ha & 34.13/56.36 & 38.49/60.43 & 37.00/58.86  \\
kr & 11.95/29.64  & --- & 11.64/29.28 \\
lg & 31.97/53.59 & 34.72/56.82 & 32.58/54.97 \\
ln & 26.18/48.16 & 28.28/50.73 & 26.46/48.67  \\
om & --- & 32.14/55.99 & 29.94/53.75 \\ \hline
\end{tabular}
\caption{BLEU/CHRF++ Scores for GPT Models with filtered sentences excluded (XE only).}
\label{tab:GPT_filtered_scores}
\end{table}

Finally, a known issue with LLMs is hallucinations---we observed this across the models we evaluated, and, unsurprisingly, hallucinations were much more frequent and more severe in low resource languages. Unfortunately, many of these hallucinations produced output that was racist, violent, or potentially contained misinformation, and would have consequences far beyond poor quality translation, 
%(\eg with poor grammar etc.) 
thus further affecting the model's usability. Fortunately, frequent hallucinations often result in low BLEU/CHRF++ scores (which means that thresholding against these scores can potentially identify output containing a high number of hallucinations, at least for LLMs).
 
%See Appendix for example outputs. 

\section{Human Evaluation and CHRF++ and BLEU Thresholds}
\label{app:human_eval}

We performed a limited, blind human evaluation on seven samples of output produced by two MT systems (Microsoft and Gemini), when translating into English from four languages ((Dinka, Eritrean Tigrinya, Lingala, Hindi). 
%comprising four language pairs (Dinka, Eritrean Tigrinya, Lingala, Hindi) produced by two MT systems (Microsoft and Gemini). 
The evaluation was done only in the XE direction, with the primary focus on establishing the utility of the CHRF++ and BLEU thresholds discussed in Sections \ref{sec:quality-usability-readiness} and \ref{sec:eval_metrics}.

The four languages were selected based on output quality---we wanted to ensure that a range of BLEU/CHRF++ scores were represented in our analysis. Output translations from Dinka (Gemini only) were unusable, while those from Eritrean Tigrinya were borderline for both systems. Lingala output was borderline for Gemini and just about passed the usability thresholds for Microsoft, and both systems had high performance for Hindi.

%The human evaluation was conducted by four evaluators, and the design is as follows: 
%Sets of forty-two sentences were extracted from both Microsoft and Gemini outputs. For Microsoft, the language pairs were Eritrean Tigrinya-English, Hindi-English, and Lingala-English; for Gemini, they were Dinka-English, Eritrean Tigrinya-English, Hindi-English, and Lingala-English. The same set of 42 sentences for both Dinka-English and Hindi-English were used and added to every evaluator’s sample, to assess and establish the level of inter-annotator agreement. 

The human evaluation was conducted by four evaluators; each evaluator received 42 reference-translation pairs from each of the seven sets of output (one for each system and language.)\footnote{Dinka was unsupported by Microsoft.}

In order to establish the level of inter-annotator agreement, each evaluator received the same 42 sentences from the Dinka-English and Hindi-English outputs. For the four remaining outputs (Eritrean Tigrinya-English and Lingala-English from both Gemini and Microsoft),  each evaluator was assigned a distinct set of 42 sentences. They then scored reference-translation pairs (of the same 42 sentences) across each of the four sets of output. Therefore, a total of 168 distinct sentence pairs were evaluated for each of these four systems-language pairs. Given the high inter-annotator agreement established by Dinka-English and Hindi-English evaluations (to be presented shortly), this decreased the margin of error from 14.97\% (if all evaluators scored the same 42 sentence pairs) to 7.33\% (168 distinct sentence pairs scored).

%However, sampling for the four remaining outputs (Eritrean Tigrinya-English and Lingala-English from both Gemini and Microsoft) was done differently as to decrease the margin of error of the scoring results. We randomly determined four distinct sets of 42 sentences. Each evaluator was assigned one set, scoring pairs with the same 42 reference sentences across each of the four outputs. For each output, this comprised 168 distinct sentence pairs across all evaluators. By using different samples for these pairs, the margin of error decreased from 14.97\% (if all evaluators scored the same 42 sentence pairs) to 7.33\% (168 distinct sentence pairs scored) given the high inter-annotator agreement established by Dinka-English and Hindi-English evaluations (to be presented shortly).

After all four language pairs were sampled (with a total of 294 sentence pairs from across the seven system-language sets assigned to each evaluator), the sentences were placed in a random order, and the evaluators reviewed the English reference and English MT output (we assumed that none of the evaluators could speak the source languages, so the input sentences were not presented). The evaluators were asked to evaluate the translated output using the modified~\citet{picinini-castilho-2025-context} Likert scale shown in  Appendix~\ref{app:human_eval_scale}, %. Then, 
%for all 294 sentences (of which 42 randomly selected sentences were shared across Hindi and Dinka extracts, for all evaluators, and 42 different randomly selected sentences were shared across Lingala and Eritrean Tigrinya extracts, for each distinct evaluator, for a total of 168 evaluated sentences for each Lingala and Eritrean Tigrinya output), 
%evaluators compared the reference with the translation for each sentence in order to 
and assign an integer score ranging from 1-5. Based on this scale, we considered a score of 3 as the threshold for usable translation in the crisis scenario.

As different numbers of distinct sentences were evaluated for various outputs, a statistical analysis assuming maximum variability (p=0.5) indicates varying margins of error at a 95\% confidence level. We established inter-annotator agreement using evaluator scores for Hindi-English (Gemini and Microsoft) and Dinka-English (Gemini), (since, as noted, all evaluators scored the same sentences across each of these outputs). Krippendorff’s Alpha over all evaluations of these three extracts is 0.7925, suggesting good inter-annotator agreement \cite{artstein-annotator-agreement}. Tables \ref{tab:human_eval_2026} and \ref{tab:human_eval_props} show the results of the human evaluation, including margins of error.

%See Table \ref{tab:human_eval_2026} for the results of the human evaluation, including margins of error.

\begin{table*}[!ht]
\centering
%\small
\begin{tabular}{|c|c c c c |}
\hline
\multicolumn{1}{|c|}{\textbf{Language}} & 
\multicolumn{4}{|c|}{\textbf{X to English}}  \\ \hline
& Microsoft & Gemini & Margin of Error & \% Margin of Error  \\ \hline

Hindi	 & 	4.5298	   & 	4.7679	   & 	0.5988	   & 	14.97\%	  \\
Dinka	 & 	---	   & 	1.8453	   & 	0.5988	   & 	14.97\%	  \\
Lingala	 & 	3.5536	   & 	3.5774	   & 	0.2932	   & 	7.33\%	  \\
Tigrinya (Eritrean)	 & 	3.0774	   & 	3.7262	   & 	0.2932	   & 	7.33\%	  \\
\hline

\end{tabular}
\caption{2026 XE average human evaluation scores and margin of error for all seven evaluated sets. See Appendix~\ref{app:human_eval_scale} for scale used in evaluation. %(See Section~\ref{sec:TICO-data} for TwB definitions of these terms, and Section~\ref{sec:quality-usability-readiness} for our definition of usability.)
}
\label{tab:human_eval_2026}
\end{table*}

\begin{table*}[!ht]
\centering
%\small
\begin{tabular}{l| c c |c c }
\hline
\multicolumn{1}{l}{\textbf{Language}} & 
\multicolumn{2}{|c}{\textbf{\% Critical Errors (score of 1)}} &
\multicolumn{2}{|c}{\textbf{\% Usable Sentences (score of 3-5)}}\\
& \textbf{Microsoft} & \textbf{Gemini} & \textbf{Microsoft} & \textbf{Gemini}  \\
%& (score of 1) & (score of 3-5) & (score of 1) & (score of 3-5) \\ 
\hline

\textbf{Hindi}	 & 	0.0/0.0/0.0/2.4	& 	0.0/0.0/0.0/2.4   & 	97.6/95.2/   & 100.0/100.0/\\
& & & 100.0/95.2 & 100.0/97.6\\
\textbf{Dinka}	 & 	---	   & 59.5/54.8/   & 	---   &  19.1/19.1/\\
& & 52.4/38.1 & & 19.1/33.3 \\
\textbf{Lingala}	 & 	7.1   & 	7.1   & 75.6   & 	81.0  \\
\textbf{Tigrinya (Eritrean)}	 & 	11.3   & 	6.6   & 	65.5   &	85.1  \\
\hline

\end{tabular}
\caption{ Proportion of sentences that were rated as having a critical error and the proportion that were considered usable by human evaluators for each language and system. For Dinka and Hindi, where each evaluator had the same set of sentences, the proportions for each evaluator's responses are shown separately.}
\label{tab:human_eval_props}
\end{table*}

At a 95\% confidence level, Hindi-English (Microsoft and Gemini), Lingala-English (Microsoft and Gemini), and Eritrean Tigrinya-English (Gemini only) met the human evaluation threshold of 3, while Dinka-English (Gemini) did not. At a 95\% confidence level, it is inconclusive whether Eritrean Tigrinya-English (Microsoft) met the threshold (given the corresponding margin of error).

%Based on our human evaluation scale (see Appendix~\ref{app:human_eval_scale}), we considered a score of 3 as usable for crisis scenarios. 
We ran an Ordinary Least Squares regression to establish a linear relationship between BLEU and our human evaluations (Pearson r = 0.9411 with p = 0.0016) and between CHRF++ and our human evaluations (Pearson r = 0.9703 with p = 0.0003), subsequently identifying the BLEU and CHRF++ values corresponding to a human evaluation of 3.0 from the linear regression. The results suggest thresholds of 23.49 for BLEU and 45.11 for CHRF++. Thus, the results of the linear regression suggest lowering the BLEU score threshold, while keeping a similar CHRF++ threshold.

Changing the BLEU threshold to 23.49 in the XE direction would suggest that a number of systems and languages currently classified as borderline would actually be considered usable. No languages classified as fully unusable would change in their usability categorization. However, with only seven data points, this is a preliminary analysis; it suggests the thresholds we used in our analysis are within a reasonable range of accuracy. A full threshold sensitivity analysis in both translation directions along with additional human evaluation (particularly in the EX direction)  %would be 
is warranted and would be a valuable area for future research.

Overall, the strong linear correlation of BLEU and CHRF++ with our human evaluation suggests that these automatic metrics may be good predictors of minimum usability as we defined it.
%as defined by human evaluation. 
Exceeding these thresholds suggests higher degrees of usability.

%Thus, exceeding a defined threshold in BLEU and CHRF++ based on a correlation with human evaluation scores may indicate an MT system is adequate.

\section{Human Evaluation Scale}
\label{app:human_eval_scale}

The Likert scale described by \citet{picinini-castilho-2025-context} used in our human evaluation, with our modifications for the crisis scenario italicized, can be found below.

\begin{enumerate}
\item Critical/Incomprehensible: \\When the translation: 
\begin{itemize}
\item Is incomprehensible (hallucinations, meanings that make no sense in the text),
\item Is completely not in the target language,
\item Contains critical errors, 
\item Can be a misleading mistranslation, 
\item May carry health, safety, legal, reputation, religious or financial implications, 
\item 
Contains profanities/insulting words not appropriate in the context. 
\end{itemize}
\witem Significant errors/effort:
\\ Translation contains significant errors and may be incomplete (such as significant mistranslations, significant portions untranslated, significant omissions) and requires significant effort to understand it. It would require a significant amount of change to be usable.

\item Some errors/some effort: 
\\ There are errors in translation, such as untranslated words and mistranslations (wrong but not misleading translations). The translation requires some effort to understand it and change it. \textit{The translation would be partially or borderline usable in the crisis context (\eg the necessary information would be conveyed and understood), but might require subsequent verification.}

\item Do not affect understanding: \\There are minor grammar errors: wrong agreement, word or der, missing prepositions, minor omissions or other errors that do not affect understanding. It would require only minor changes. \textit{The translation would still be usable in the crisis context (\eg the necessary information would be conveyed and understood).}

\item  Complete/fluent: \\Translation is (or seems to be) correct, complete, fluent, easy to read, and contains no errors. It looks like (almost) human translation and requires no changes. \textit{The translation would be usable in the crisis context and would be fully understood.}
\end{enumerate}

It is important to note here, that the use of the word `borderline' 
%as used in 
for entry 3 above 
%in describing a score of 3 
differs from our use of the word in our overall analysis. As described previously, we defined systems that were `borderline usable' as producing output translations that scored higher than the threshold for one metric but not both, indicating that they were almost, but not quite, of high enough quality to be usable. In this human evaluation scale however, we use `borderline' to mean just barely usable---the translation is low quality, and if it were reviewed by a human post-editor, they would likely make changes. This distinction was unintentional and was a source of confusion for at least one of our reviewers. We regret this error and would fix this in future human evaluation work.

\section{Qualitative Error Analysis and Hallucinations}
\label{app:errors_and_hallucinations}

% We performed a qualitative error analysis on the data, which will be described in the following section.
We also conducted a preliminary error analysis on LLM output translations for a subset of languages in the XE translation direction. We looked first at some of the lowest scoring systems, (which had BLEU scores of around 10). We then looked at Gemini's output when translating from Nigerian Fulfulde to English---while scores here were still too low to be usable, they were much closer to reasonable (BLEU/CHRF++ were 21.21/42.42). Finally, we examined some of the errors produced by systems that were considered borderline usable.

%We approached error analysis by categorizing output translations~\footnote{We performed this qualitative analysis in the X to English direction.} according to their performance in BLEU and CHRF++ metrics, then identifying patterns of errors that emerged in low-quality translations. Performance for outputs we analyzed fell into three categories: 
%make a table w/ relevant languages in each category?
%ADD TABLE W/ LANGUAGES/OUTPUTS IN EACH CATEGORY HERE (maybe not suuuper necessary)
%very low performing (BLEU scores around 10); better quality, but still unusable (BLEU scores around 20); and borderline usable (BLEU less than 30 with CHRF++ greater than 45).
%; and just over threshold (BLEU and CHRF++ slightly over 30 and 45, respectively). 

In the very low performing languages---Dinka and Kanuri consistently fell into this category across models---hallucinations were prevalent. The conversational medical domain was particularly prone to this; in many cases, the output for the entire domain is fully hallucinated. We identified several types of hallucinations: 
%this included 
output that was completely irrelevant or random, output that was unrelated to the source sentence but on topic, and output where the source sentence was translated, but with extra, hallucinated content included. Many of the hallucinations included religious themes, and in a few instances, centered around one key word in the source. Unfortunately, there were also many hallucinations (or severe mistranslations) that were problematic, containing content that was racist, sexist, actively spreading misinformation, or were otherwise aggressive or violent. Examples of various hallucinations are shown in 1-7 below:

\begin{enumerate}
\item \textbf{Irrelevant Output, GPT-5.4-full, Kanuri-English }

\textbf{Reference:} extensions of monitoring and surveillance would have to have sunset clauses;

\smallskip

\textbf{Output:} The people have ten fingers and ten toes, and all of them are counted together.

\item \textbf{Extra Hallucinated Content, Gemini, Dinka-English} 

\textbf{Reference:} As of 19 March drive in tests were offered in several large cities.

\smallskip

\textbf{Output:} On the 19th of the month of Kol, many people were tested for the disease, and \textit{they were all remembered in prayer.}

\item \textbf{Religious Hallucinations, GPT-4o, Nigerian Fulfulde-English} 

\textbf{Reference:} since you have a fever

\smallskip

\textbf{Output:} Prepare your mat, we are about to pray.

\item  \textbf{Key Word Based Hallucination, Gemini, Kanuri-English} 

Reference: \textit{Panama} has made it obligatory to wear a face mask whenever going outside, while also recommending the manufacture of a homemade face mask for those who cannot purchase face masks.

\smallskip

\textbf{Output:} \textit{The Panama Papers} revealed how the wealthy hide their assets within the community, and also how various types of shell companies are used to consult with those who wish to evade taxes.

\item \textbf{Problematic Hallucination: Aggressive, GPT-5.4-mini, Dinka-English} 

\textbf{Reference:} Patients with fever and/or respiratory symptoms and acute fever, even without pulmonary imaging abnormalities, should be screened for the virus for early diagnosis.

\smallskip

\textbf{Output:} People are lazy, foolish, and full of anger and hatred, so if they are not careful, they will end up making mistakes and being deceived.

\item  \textbf{Problematic Hallucination/Mistranslation: Violent, Gemini, Dinka-English }

\textbf{Reference:} "after patient care.The addition of antiseptic chemicals to soap (`medicated' or `antimicrobial' soaps) confers killing action to a hand-washing agent."

\smallskip

\textbf{Output:} If there is no one sick in Akoor to be helped with soap, then they should be killed immediately.
% it said Akɔɔr in the output but it was producing a unicode character error

\item  \textbf{Problematic Hallucination/Mistranslaion: Spreads Misinformation, Gemini, Dinka-English }

\textbf{Reference:} Other types of respirators can provide greater protection and improve worker comfort.The WHO does not recommend coveralls, as COVID-19 is a respiratory disease rather than being transmitted through bodily fluids.

\smallskip

\textbf{Output:} Some of the best ways to help save lives have not been implemented, and the World Health Organization has not yet approved the use of masks to prevent the spread of COVID-19. 

\end{enumerate}

Other common errors among the lowest performing outputs include: difficulty with medical terminology, archaic word choice, meaning of translation is the opposite of the intended/original meaning, and unusual phrasing. Some examples are shown in 1-4 below.

%ADD TABLE/EXAMPLES HERE
\begin{enumerate}
\item \textbf{Medical Terminology, Gemini, Dinka-English}

\textbf{Reference:} and with your history of \textit{diabetes}

\smallskip

\textbf{Output:} What is the name you have been given in the \textit{sugar}?

\item \textbf{Medical Terminology, Gemini, Dinka-English}

\textbf{Reference:} Instead, they may cause severe side effects, especially avascular \textit{osteonecrosis}, dramatically affecting the prognosis.

\smallskip

\textbf{Output:} There is something that is done in the body, just as death is the \textit{bone of death}, and it is in that which is found to be bad

\item \textbf{Archaic/Unusual Phrasing, Gemini, Dinka-English} 

\textbf{Reference:} Many different hand dryer manufacturers exist, and hand driers have been compared against drying with paper towels.

\smallskip

\textbf{Output:} The people of the village of the hand did not do it, but they were found in the village of the hand wearing clothes that covered them well.

\item \textbf{Opposite Meaning, Gemini, Kanuri-English} 

\textbf{Reference:} In addition, Turkey seized hundreds of ventilators destined for Spain.

\smallskip

\textbf{Output:} "Because of that, Turkey is sending the medical equipment called a 'ventilator to' Spain, which has been hit hard by the world pandemic."

%\item \textbf{Unusual Phrasing and Medical Terminology, Gemini, Dinka-English} 

%\textbf{Reference:} but i have difficulty breathing

%\smallskip

%\textbf{Output:} The wind is strong inside me.

\end{enumerate}

We then turned to Gemini's output when translating from Nigerian Fulfulde to English. %output to categorize mistranslations among the better quality, but still unusable outputs. 
As noted earlier, the BLEU/CHRF++ scores were still too low to be considered usable, but more reasonable than those for Dinka or Kanuri. In general, we observed the same error types as with the very low performing languages, but with fewer numbers of problematic outputs and errors that did not seem as severe.
%in lesser quantity and often in lesser intensity. 
The conversational medical domain was not fully hallucinated here; instead, the medical terms specifically were hallucinated or mistranslated. The mistranslations tended to be consistent (\eg ``frog'' was present in several translations where the reference sentence referred to a ``fever''). Some of the mistranslations from this output could be inferred from context (\eg ``shortness of breath'' in the reference sentence aligned with ``breath runs out'' in the translation). However, there are still some extremely problematic mistranslations and hallucinations present in the output. See 1-5 in the list below for examples.
%ADD TABLE/EXAMPLES HERE

\begin{enumerate}
\item \textbf{Poor Quality, Mistranslation, Gemini, Nigerian Fulfulde-English }

\textbf{Reference:} are you having any shortness of breath?

\smallskip

\textbf{Output:} Is it okay if your breath runs out?

\item \textbf{Poor Quality, Gemini, Nigerian Fulfulde-English} 

\textbf{Reference:} are there other people sick as you at home with your same symptoms?

\smallskip

\textbf{Output:} Are there other people in your town who are sick with the same kind of illness as you?

\item \textbf{Medical Terminology, Gemini, Nigerian Fulfulde-English}

\textbf{Reference:} I've very concerned of this chest pain

\smallskip

\textbf{Output:} My heart is aching and I am very troubled.

\item \textbf{Opposite/Untrue Meaning, Gemini, Nigerian Fulfulde-English} 

\textbf{Reference:} The White House Coronavirus Task Force was established on 29 January.

\smallskip

\textbf{Output:} The White House will release the coronavirus report on January 29th.

\item \textbf{Problematic Mistranslation: Spreads Misinformation, Gemini, Nigerian Fulfulde-English }

\textbf{Reference:} Several measures are commonly used to quantify mortality.

\smallskip

\textbf{Output:} Many tricks are being used to inflate the death toll.
\end{enumerate}

Finally, we turn to systems that produced borderline usable translations, focusing on various Lingala-English and Oromo-English outputs across systems. Translations were significantly better than the lower-performing, unusable, systems, especially in the conversational medical domain. In many cases, medical or technical terms were simplified, but not necessarily incorrect. We also see mismatches in tone between the reference and output (or some added dramatization), and translations that were almost or partially correct (\ie some parts were conveyed correctly, but some information was mistranslated or omitted). We do still note some archaic phrasing or word choices, and some problematic mistranslations. Some examples are shown in 1-10 in the list below.

%Most mistranslations fell into one of the following patterns: simplifying technical terms, tone difference or dramatization, archaic word usage, close translations (partially correct; \ie some parts conveyed correctly, but mistranslating or omitting other information). The list below displays examples of these mistranslation patterns.
%ADD TABLE/EXAMPLES HERE

\begin{enumerate}
\item \textbf{Simplifying Technical Terms, Gemini, Lingala-English} 

\textbf{Reference:} digital surveillance would have to avoid exacerbating discrimination and marginalisation;

\smallskip

\textbf{Output:} Computer-based monitoring should not be done with malice toward others.

\item \textbf{Simplifying Technical Terms, GPT-5.4-mini, Lingala-English} 

\textbf{Reference:} Patients often develop an atypical pneumonia, acute lung injury, and acute respiratory distress syndrome (ARDS).

\smallskip

\textbf{Output:} Often, sick people had lung disease, severe lung inflammation, and severe breathing difficulty (ARDS).

\item \textbf{Tone Difference or Dramatization, Gemini, Lingala-English }

\textbf{Reference:} But it's almost impossible to make any sensible projection right now about when that will be.

\smallskip 

\textbf{Output:} But there is no way to know how things will be these days.

\item \textbf{Tone Difference or Dramatization, GPT-5.4-full, Oromo-English} 

\textbf{Reference:} \textit{The measure was taken} to enforce quarantine and protect those who may come into contact with infected citizens.

\smallskip

\textbf{Output:} \textit{The punishment is imposed} to influence the duration of isolation and also to prevent people from touching citizens infected with the disease.

\item \textbf{Tone Difference or Dramatization, Gemini, Lingala-English}

\textbf{Reference:} Many consider an economic recession to be likely.

\smallskip

\textbf{Output:} Many people think that the economy will truly collapse.

\item \textbf{Archaic Phrasing, GPT-5.4-mini, Lingala-English} 

\textbf{Reference:} \textit{As of March 19}, these three had already recovered.

\smallskip

\textbf{Output:} \textit{On the 19th day of the third month}, three people survived.

\item \textbf{Archaic Word Choice, GPT-5.4-mini, Lingala-English} 

\textbf{Reference:} \textit{Doctors} have said they are using the drug when "there's no other option."

\smallskip

\textbf{Output:} The \textit{healers} say they use that medicine only when there is no other medicine.

\item \textbf{Partially Correct, GPT-5.4-full, Oromo-English} 

\textbf{Reference:} The study failed to find evidence of viral replication in \textit{pigs, ducks,} and chickens.

\smallskip

\textbf{Output:} The study could not find evidence that the virus reproduces in \textit{sheep, donkeys,} and chickens.

\item \textbf{Problematic Mistranslation, Gemini, Lingala-English}

\textbf{Reference:} Women who are pregnant or breastfeeding or who do not have an effective contraceptive method \textit{are not eligible}.

\smallskip

\textbf{Output:} Women who are pregnant, breastfeeding, or not using effective contraception \textit{are not permitted to enter}.

\item \textbf{Mistranslation, Gemini, Lingala}

\textbf{Reference:} Pakistan has said that it will \textit{not be evacuating any citizens} from China.

\smallskip

\textbf{Output:} Pakistan said that it would \textit{launch its own person into space} with China.

\end{enumerate}

% Although we focused on X-to-English error analysis, we noticed an important error type in the English-to-X direction: many medical terms (\eg "chest pain," "fever," and "diabetes") stay in English. In some cases, the entire translated sentence remained in English. This was an issue among very low performing outputs and requires future analysis.

%\newpage
\section{The Choice of Evaluation Metric(s)}
\label{app:eval_metric_choice}

The choice of evaluation metric(s) is always an issue in a study such as ours. While human evaluation is the gold standard, it is both time consuming and expensive to implement~\cite{bugliarello-etal-2020-easier}. In fact, expense aside, locating reviewers for many of the lower resource languages in TICO-19 would be difficult at best. As an example, translators and reviewers were a significant problem for these languages during the development of TICO-19 a few years ago.

Still, the choice of evaluation metric is important and should not be taken lightly. %In no way have we de-prioritized the selection of metrics in this paper. 
Yet, the options available for a study such as ours are limited, given the languages that we focused on.

On the surface, given the underlying technology used for translation today, neural metrics such as BERTScore or COMET would make the most sense. We knew this going in and made sure that we calculated scores for these two metrics for almost every output across our three-year study. But we could not use these metrics for all data points. Alas, these metrics do not work well for lower-resource languages and specialized domains
%given the reliability issues even on under-resourced supported languages that are under-resourced, and there is evidence that they may not work well in specialized domains 
(as noted in Section~\ref{sec:quality-usability-readiness}, per \cite{nllb-2022, tang-etal-BERTScore-low-resource-2024, falcao-etal-2024-comet, zouhar-etal-2024-pitfalls, zouhar-etal-2024-domain}).  Further, there is no clear consensus on how to interpret these metrics directly.

BLEU has had a long and, dare we say, sordid past in the field of Machine Translation. It is a metric that everyone loves to hate, and most would freely admit that it has outlived its shelf-life. That said, it still manages to cling to life, and is still used widely in the field, even in industrial settings (\eg for hill climbing during development). Why won’t BLEU just die already? The reasons vary greatly, but we feel it is mostly due to the fact that it is cheap and easy to implement, that it is universal (\ie it can be applied to \textit{any} language pair), that we understand the scores and what they mean, and that it provides some sort of continuity across the lengthy history of our field.

In an earlier version of this paper, which was submitted to and rejected by a major conference, we used BLEU for most of our analysis (we still had BERTScore and COMET, but could not apply these broadly, as noted). Concerns about BLEU dominated in the reviews and in the final acceptance assessment. Some of the reviewers suggested we should use CHRF++ instead, which we subsequently adopted wholeheartedly.

CHRF++ ~\cite{Popvic-CHRFPP-2017} is favored in low-resource MT workshops, such as LoResMT~\cite{loresmt-2026-1}, now in its ninth year. CHRF++ is still an n-gram based metric, so it will not catch semantic similarities across texts (something that the neural methods are more adept at). But, because it uses a composite word and character-based n-gram score, it is a more effective metric for evaluating the output of typologically diverse languages, and much more effective than BLEU for morphologically rich languages in particular~\cite{Popvic-CHRFPP-2017, Popvic-CHRF-2015}.

Because of BLEU’s lengthy history in our field, and because of CHRF++’s frequent use in low-resource contexts, we felt a combination of the two metrics was warranted. This follows other work in the field for low-resource languages that show that a composite score across the two metrics provides complementary insights into the text being measured (\eg ~\citet{kumar-et-al-2026-CHRF-BLEU}), and is in line with the recommendation made by \citet{marie-etal-2021-scientific} for the use of additional metrics alongside BLEU.

There is still the question of comparability between languages. BLEU has been shown to not be fully comparable between languages (\eg \citet{marie-etal-2021-scientific}), although we could not find a reference saying the same for CHRF++. %(That being said, since they are both basically n-gram based, issues with one may still apply to the other.) 
This issue has been noted by a couple of reviewers. Yet, we feel it is important to recognize that these measures are not just producing white noise; there \textit{is} signal that is returned, and that signal can be interpreted, as long as we ensure that we are using standard methods and practices, \eg SacreBLEU~\cite{post-sacrebleu-wmt-2018}. Incorporating both measures into one threshold, and testing that this threshold is at least in the ballpark (\eg through human eval), gives us enough signal to be able to say that certain language pairs may be usable in the context of crisis preparedness and response and others not. %It is noisy signal but it is still an interpretable signal.

It is crucial to note that the ultimate goal of this paper was to offer insights to crisis managers, aid providers and the like about what tools %and what languages 
\textit{might} be usable in the field, and for which languages. If a particular language pair is well below our threshold, it %may in fact 
will likely be necessary to pursue other options. %for using available translation tools for it.
Outside of other available information, these communities and practitioners are left completely in the dark, in that they have no information about what tools they should be using, what the quality of these tools are, and whether the tools are usable for certain languages versus others.~\footnote{
In fact, some emergency managers, outside of additional information, have determined overall and cross-lingual usability of particular tools, such as Google Translate, based on the evaluation of \textit{one} language, for which they had a native speaker in house~\cite{wilson-etal-2026}.
}
We hope to give these communities at least some glimmer of light in this dark about the tools that so many of them are seeking to use or have already become dependent on. As a field, burying our heads in the sand and saying that we have no way to measure quality for these language pairs is not an adequate response. The development of high-performing and broadly applicable evaluation metrics that fully include low-resource languages is a vital area for further research (though this is a distinct goal from those of this study), but we cannot wait for this work to be completed before evaluating translation tools on %most 
low resource languages, particularly when such tools are likely already being used. We owe these communities more than that. 
%And that’s what we sought to do here.

\clearpage

\clearpage

\clearpage
\begin{table*}
%\begin{tabular}{|*{9}{c|}}
%\small
\begin{tabular}{|c|c c c c c c c |}
\hline
\multicolumn{1}{|c|}{\textbf{Language}} & 
\multicolumn{7}{|c|}{\textbf{English to X}}  \\ \hline
& MS & Google & GPT-4o & GPT-5.4 & GPT-5.4-mini & Gemini & Claude  \\ \hline

%% Pivot Languages first
\multicolumn{1}{|c|}{\textbf{------Pivot------}} & \multicolumn{7}{|c|}{} \\ \hline
Arabic	 & 	28.71	   & \textbf{	33.24	 }  & 	31.35	   & 	32.37	   & 	31.68	   & 	30.75	   & 	28.36	   \\
Chinese	 & 	47.95	   & \textbf{	53.10	 }  & 	50.36	   & 	47.88	   & 	51.01	   & 	47.72	   & 	49.45	   \\
French	 & 	44.15	   & 	34.94	   & 	41.75	   & \textbf{	46.61	 }  & 	46.58	   & 	42.28	   & 	41.87	   \\
Hindi	 & \textbf{	46.12	 }  & 	38.88	   & 	36.70	   & 	37.57	   & 	40.63	   & 	40.01	   & 	40.24	   \\
Indonesian	 & 	54.80	   & 	53.70	   & 	54.23	   & \textbf{	55.29	 }  & 	55.21	   & 	52.66	   & 	52.11	   \\
Portuguese (BR)	 & 	51.01	   & 	51.79	   & 	51.39	   & \textbf{	53.20	 }  & 	52.88	   & 	52.77	   & 	51.73	   \\
Russian	 & 	31.52	   & 	35.46	   & 	32.81	   & 	34.29	   & \textbf{	35.65	 }  & 	33.66	   & 	31.73	   \\
Spanish (LA)	 & 	51.62	   & 	55.99	   & 	53.83	   & \textbf{	56.84	 }  & 	55.19	   & 	54.45	   & 	52.80	   \\
Swahili	 & 	34.79	   & \textbf{	36.54	 }  & 	33.98	   & 	33.50	   & 	34.81	   & 	33.83	   & 	28.91	   \\
 \hline

%% Important languages
%\\ \hline
\multicolumn{1}{|c|}{\textbf{------Important------}} & \multicolumn{7}{|c|}{} \\ \hline    
Bengali	 & \textbf{	21.49	 }  & 	15.61	   & 	16.58	   & 	14.96	   & 	15.97	   & 	13.90	   & 	16.30	   \\
Farsi	 & 	25.62	   & \textbf{	29.85	 }  & 	26.72	   & 	22.81	   & 	23.67	   & 	24.43	   & 	23.17	   \\
Malay	 & \textbf{	52.93	 }  & 	52.88	   & 	47.58	   & 	50.83	   & 	52.34	   & 	47.27	   & 	46.02	   \\
Marathi	 & \textbf{	16.20	 }  & 	15.18	   & 	14.00	   & 	14.34	   & 	15.27	   & 	14.19	   & 	12.36	   \\
Myanmar	 & 	52.49	   & 	60.21	   & 	49.72	   & \textbf{	61.21	 }  & 	60.14	   & 	59.21	   & 	49.22	   \\
Tagalog	 & 	41.60	   & \textbf{	45.85	 }  & 	41.24	   & 	37.22	   & 	42.78	   & 	38.88	   & 	40.01	   \\
Tamil	 & \textbf{	25.66	 }  & 	17.65	   & 	11.17	   & 	14.12	   & 	14.00	   & 	16.26	   & 	13.23	   \\
Urdu	 & 	24.90	   & \textbf{	28.11	 }  & 	23.39	   & 	22.74	   & 	24.59	   & 	23.80	   & 	23.39	   \\
 \hline 

%% Priority languages
%\\ \hline
\multicolumn{1}{|c|}{\textbf{------Priority------}} & \multicolumn{7}{|c|}{} \\ \hline
Amharic	 & \textbf{	13.37	 }  & 	12.53	   & 	5.62	   & 	8.37	   & 	8.89	   & 	9.92	   & 	7.78	   \\
Dari	 & \textbf{	22.49	 }  & 	---	   & 	16.58	   & 	15.60	   & 	15.91	   & 	16.93	   & 	15.78	   \\
Dinka	 & 	---	   & \textbf{	4.48	 }  & 	---	   & 	3.15	   & 	2.96	   & 	---	   & 	---	   \\
Hausa	 & \textbf{	28.77	 }  & 	22.49	   & 	22.50	   & 	19.52	   & 	21.59	   & 	18.82	   & 	17.29	   \\
Kanuri	 & 	---	   & \textbf{	5.31	 }  & 	---	   & 	3.17	   & 	4.52	   & 	---	   & 	---	   \\
Khmer	 & 	64.74	   & \textbf{	67.20	 }  & 	55.35	   & 	63.25	   & 	62.83	   & 	62.55	   & 	52.22	   \\
Kinyarwanda	 & \textbf{	18.86	 }  & 	16.54	   & 	17.14	   & 	18.33	   & 	18.72	   & 	18.28	   & 	9.40	   \\
Kurdish (Kurmanji)	 & \textbf{	17.06	 }  & 	15.40	   & 	10.21	   & 	---	   & 	13.68	   & 	---	   & 	10.16	   \\
Kurdish (Sorani)	 & \textbf{	10.93	 }  & 	10.78	   & 	6.00	   & 	7.80	   & 	7.43	   & 	8.04	   & 	5.53	   \\
Lingala	 & \textbf{	44.17	 }  & 	15.18	   & 	14.70	   & 	15.12	   & 	15.17	   & 	15.89	   & 	9.22	   \\
Luganda	 & 	17.27	   & \textbf{	19.88	 }  & 	14.93	   & 	14.34	   & 	15.48	   & 	15.48	   & 	10.66	   \\
Nepali	 & 	21.28	   & \textbf{	23.45	 }  & 	18.24	   & 	15.75	   & 	19.01	   & 	16.83	   & 	17.35	   \\
Nigerian Fulfulde	 & 	---	   & 	3.69	   & 	---	   & 	4.58	   & 	4.26	   & \textbf{	4.98	 }  & 	---	   \\
Oromo	 & 	---	   & \textbf{	9.11	 }  & 	6.50	   & 	7.71	   & 	8.40	   & 	8.04	   & 	5.28	   \\
Pashto	 & 	13.56	   & \textbf{	24.62	 }  & 	17.47	   & 	15.37	   & 	16.36	   & 	15.15	   & 	10.34	   \\
Somali	 & 	17.62	   & \textbf{	18.17	 }  & 	6.92	   & 	15.49	   & 	17.03	   & 	16.09	   & 	11.51	   \\
Tigrinya (Eritrean)	 & 	5.22	   & \textbf{	6.19	 }  & 	---	   & 	4.61	   & 	4.09	   & 	5.42	   & 	---	   \\
Tigrinya (Ethiopian)	 & 	5.71	   & \textbf{	6.69	 }  & 	---	   & 	5.42	   & 	5.15	   & 	5.70	   & 	---	   \\
Zulu	 & 	19.53	   & \textbf{	20.31	 }  & 	16.26	   & 	17.72	   & 	19.53	   & 	16.27	   & 	11.34	   \\
 \hline
 
%Congolese Swahili & \cellcolor{gray}{---} & \cellcolor{gray}{---} & \cellcolor{gray}{---} & \cellcolor{gray}{---} & \cellcolor{gray}{---} & \cellcolor{gray}{---} & \cellcolor{gray}{---} & \cellcolor{gray}{---} \\ \hline 

\end{tabular}
\caption{2026 EX BLEU scores. The best system for each language is in bold. %(See Section~\ref{sec:TICO-data} for TwB definitions of these terms, and Section~\ref{sec:quality-usability-readiness} for our definition of usability.)
}
\label{tab:bleu_EX_2026}
\end{table*}
\clearpage

\begin{table*}
%\begin{tabular}{|*{9}{c|}}
%\small
\begin{tabular}{|c|c c c c c c c |}
\hline
\multicolumn{1}{|c|}{\textbf{Language}} & 
\multicolumn{7}{|c|}{\textbf{English to X}}  \\ \hline
& MS & Google & GPT-4o & GPT-5.4 & GPT-5.4-mini & Gemini & Claude  \\ \hline

%% Pivot Languages first
\multicolumn{1}{|c|}{\textbf{------Pivot------}} & \multicolumn{7}{|c|}{} \\ \hline
Arabic	 & 	55.08	   & 	59.23	   & 	57.05	   & \textbf{	59.32	 }  & 	57.86	   & 	57.27	   & 	55.64	   \\
Chinese	 & 	45.74	   & \textbf{	50.48	 }  & 	47.50	   & 	43.74	   & 	48.10	   & 	42.66	   & 	45.89	   \\
French	 & 	63.64	   & 	63.13	   & 	62.83	   & \textbf{	65.74	 }  & 	65.52	   & 	63.05	   & 	63.02	   \\
Hindi	 & \textbf{	65.62	 }  & 	61.76	   & 	59.06	   & 	60.34	   & 	62.81	   & 	61.75	   & 	62.42	   \\
Indonesian	 & 	74.72	   & 	74.88	   & 	74.15	   & \textbf{	75.08	 }  & 	74.93	   & 	73.92	   & 	73.74	   \\
Portuguese (BR)	 & 	72.06	   & 	72.05	   & 	71.81	   & \textbf{	73.29	 }  & 	72.99	   & 	72.99	   & 	72.62	   \\
Russian	 & 	56.13	   & \textbf{	59.57	 }  & 	57.42	   & 	58.85	   & 	58.11	   & 	58.19	   & 	57.14	   \\
Spanish (LA)	 & 	71.88	   & 	74.17	   & 	73.17	   & \textbf{	75.05	 }  & 	74.08	   & 	73.64	   & 	72.84	   \\
Swahili	 & 	59.96	   & \textbf{	61.59	 }  & 	59.97	   & 	60.55	   & 	60.44	   & 	60.16	   & 	55.86	   \\
 \hline

%% Important languages
%\\ \hline
\multicolumn{1}{|c|}{\textbf{------Important------}} & \multicolumn{7}{|c|}{} \\ \hline    
Bengali	 & \textbf{	51.40	 }  & 	46.69	   & 	47.28	   & 	45.97	   & 	47.25	   & 	44.38	   & 	47.55	   \\
Farsi	 & 	52.32	   & \textbf{	56.44	 }  & 	53.76	   & 	51.16	   & 	52.71	   & 	51.80	   & 	51.38	   \\
Malay	 & 	73.16	   & \textbf{	74.03	 }  & 	69.28	   & 	73.19	   & 	73.70	   & 	71.09	   & 	70.09	   \\
Marathi	 & \textbf{	46.80	 }  & 	46.05	   & 	43.40	   & 	44.67	   & 	44.97	   & 	44.33	   & 	41.57	   \\
Myanmar	 & 	48.16	   & \textbf{	54.56	 }  & 	45.70	   & 	54.54	   & 	53.99	   & 	52.62	   & 	42.84	   \\
Tagalog	 & 	64.72	   & \textbf{	68.27	 }  & 	65.17	   & 	62.50	   & 	66.61	   & 	63.42	   & 	63.83	   \\
Tamil	 & \textbf{	56.87	 }  & 	53.11	   & 	44.49	   & 	49.88	   & 	48.75	   & 	50.68	   & 	46.48	   \\
Urdu	 & 	50.77	   & \textbf{	52.80	 }  & 	49.32	   & 	49.32	   & 	50.67	   & 	49.17	   & 	49.37	   \\
 \hline 

%% Priority languages
%\\ \hline
\multicolumn{1}{|c|}{\textbf{------Priority------}} & \multicolumn{7}{|c|}{} \\ \hline
Amharic	 & \textbf{	35.84	 }  & 	35.78	   & 	23.51	   & 	31.47	   & 	31.23	   & 	32.62	   & 	27.60	   \\
Dari	 & \textbf{	48.15	 }  & 	---	   & 	43.69	   & 	43.47	   & 	44.23	   & 	44.56	   & 	43.63	   \\
Dinka	 & 	---	   & \textbf{	25.31	 }  & 	---	   & 	20.82	   & 	19.93	   & 	---	   & 	---	   \\
Hausa	 & \textbf{	53.44	 }  & 	51.26	   & 	49.12	   & 	48.91	   & 	49.77	   & 	48.21	   & 	43.49	   \\
Kanuri	 & 	---	   & \textbf{	29.94	 }  & 	---	   & 	19.32	   & 	18.83	   & 	---	   & 	---	   \\
Khmer	 & 	60.53	   & \textbf{	61.60	 }  & 	51.07	   & 	57.20	   & 	57.52	   & 	56.86	   & 	46.66	   \\
Kinyarwanda	 & 	45.88	   & 	47.87	   & 	45.14	   & \textbf{	49.10	 }  & 	47.74	   & 	47.50	   & 	33.23	   \\
Kurdish (Kurmanji)	 & \textbf{	44.70	 }  & 	44.38	   & 	35.29	   & 	---	   & 	42.09	   & 	---	   & 	37.80	   \\
Kurdish (Sorani)	 & 	42.19	   & \textbf{	43.59	 }  & 	34.20	   & 	40.77	   & 	38.74	   & 	41.70	   & 	32.75	   \\
Lingala	 & \textbf{	65.59	 }  & 	43.85	   & 	43.42	   & 	46.07	   & 	45.62	   & 	45.68	   & 	37.34	   \\
Luganda	 & 	43.09	   & \textbf{	48.32	 }  & 	40.75	   & 	44.92	   & 	43.53	   & 	43.64	   & 	32.09	   \\
Nepali	 & 	53.42	   & \textbf{	56.19	 }  & 	50.06	   & 	49.15	   & 	52.39	   & 	49.35	   & 	50.25	   \\
Nigerian Fulfulde	 & 	---	   & 	22.31	   & 	---	   & \textbf{	23.96	 }  & 	22.23	   & 	23.80	   & 	---	   \\
Oromo	 & 	---	   & \textbf{	43.60	 }  & 	35.66	   & 	42.89	   & 	41.01	   & 	40.23	   & 	32.05	   \\
Pashto	 & 	39.47	   & \textbf{	48.52	 }  & 	42.66	   & 	41.54	   & 	42.25	   & 	40.10	   & 	33.66	   \\
Somali	 & 	44.92	   & \textbf{	46.62	 }  & 	28.05	   & 	44.68	   & 	45.44	   & 	44.53	   & 	37.18	   \\
Tigrinya (Eritrean)	 & 	23.75	   & \textbf{	24.89	 }  & 	---	   & 	22.22	   & 	20.64	   & 	23.45	   & 	---	   \\
Tigrinya (Ethiopian)	 & 	24.09	   & \textbf{	25.41	 }  & 	---	   & 	24.03	   & 	23.07	   & 	24.25	   & 	---	   \\
Zulu	 & 	52.76	   & \textbf{	53.61	 }  & 	48.18	   & 	52.30	   & 	53.34	   & 	50.31	   & 	42.54	   \\
 \hline
 
%Congolese Swahili & \cellcolor{gray}{---} & \cellcolor{gray}{---} & \cellcolor{gray}{---} & \cellcolor{gray}{---} & \cellcolor{gray}{---} & \cellcolor{gray}{---} & \cellcolor{gray}{---} & \cellcolor{gray}{---} \\ \hline 

\end{tabular}
\caption{2026 EX CHRF++ scores. The best system for each language is in bold. %(See Section~\ref{sec:TICO-data} for TwB definitions of these terms, and Section~\ref{sec:quality-usability-readiness} for our definition of usability.)
}
\label{tab:chrf_EX_2026}
\end{table*}
\clearpage

\begin{table*}
%\begin{tabular}{|*{9}{c|}}
%\small
\begin{tabular}{|c|c c c c c c c |}
\hline
\multicolumn{1}{|c|}{\textbf{Language}} & 
\multicolumn{7}{|c|}{\textbf{X to English}}  \\ \hline
& MS & Google & GPT-4o & GPT-5.4 & GPT-5.4-mini & Gemini & Claude  \\ \hline

%% Pivot Languages first
\multicolumn{1}{|c|}{\textbf{------Pivot------}} & \multicolumn{7}{|c|}{} \\ \hline
Arabic	 & 	44.91	   & 	45.36	   & 	42.88	   & 	44.62	   & 	44.42	   & \textbf{	46.57	 }  & 	40.40	   \\
Chinese	 & 	36.40	   & 	35.00	   & 	33.35	   & 	35.38	   & 	35.21	   & \textbf{	36.59	 }  & 	33.16	   \\
French	 & \textbf{	45.57	 }  & 	43.87	   & 	43.01	   & 	43.48	   & 	43.21	   & 	44.17	   & 	41.35	   \\
Hindi	 & 	53.28	   & \textbf{	58.67	 }  & 	46.22	   & 	53.37	   & 	53.62	   & 	53.69	   & 	49.70	   \\
Indonesian	 & 	50.53	   & \textbf{	53.93	 }  & 	49.31	   & 	52.70	   & 	52.17	   & 	51.08	   & 	48.67	   \\
Portuguese (BR)	 & \textbf{	55.46	 }  & 	53.94	   & 	52.91	   & 	54.82	   & 	54.82	   & 	55.40	   & 	52.25	   \\
Russian	 & 	37.06	   & \textbf{	38.87	 }  & 	36.18	   & 	36.39	   & 	35.69	   & 	37.93	   & 	34.36	   \\
Spanish (LA)	 & \textbf{	55.48	 }  & 	53.22	   & 	52.78	   & 	53.23	   & 	52.73	   & 	54.08	   & 	50.58	   \\
Swahili	 & 	42.24	   & \textbf{	48.35	 }  & 	42.00	   & 	47.30	   & 	45.30	   & 	47.21	   & 	38.38	   \\

\hline

%% Important languages
%\\ \hline
\multicolumn{1}{|c|}{\textbf{------Important------}} & \multicolumn{7}{|c|}{} \\ \hline    
Bengali	 & 	46.53	   & \textbf{	51.06	 }  & 	42.22	   & 	47.26	   & 	47.05	   & 	47.60	   & 	43.17	   \\
Farsi	 & 	39.41	   & \textbf{	43.84	 }  & 	39.10	   & 	40.41	   & 	40.04	   & 	41.94	   & 	35.99	   \\
Malay	 & 	55.11	   & \textbf{	58.98	 }  & 	50.86	   & 	57.59	   & 	56.47	   & 	54.54	   & 	52.47	   \\
Marathi	 & 	39.25	   & \textbf{	43.69	 }  & 	37.13	   & 	41.14	   & 	40.85	   & 	40.97	   & 	35.72	   \\
Myanmar	 & 	31.82	   & \textbf{	36.66	 }  & 	27.10	   & 	33.92	   & 	33.95	   & 	35.34	   & 	27.63	   \\
Tagalog	 & 	60.13	   & \textbf{	64.53	 }  & 	52.89	   & 	59.63	   & 	60.14	   & 	58.20	   & 	54.21	   \\
Tamil	 & 	37.75	   & \textbf{	42.55	 }  & 	31.81	   & 	38.96	   & 	38.82	   & 	40.58	   & 	33.89	   \\
Urdu	 & 	38.37	   & \textbf{	43.04	 }  & 	35.42	   & 	39.87	   & 	39.07	   & 	40.39	   & 	34.98	   \\

\hline 

%% Priority languages
%\\ \hline
\multicolumn{1}{|c|}{\textbf{------Priority------}} & \multicolumn{7}{|c|}{} \\ \hline
Amharic	 & 	33.31	   & \textbf{	38.18	 }  & 	26.15	   & 	35.38	   & 	35.36	   & 	37.04	   & 	29.94	   \\
Dari	 & 	37.72	   & \textbf{	41.99	 }  & 	37.60	   & 	38.90	   & 	38.35	   & 	40.79	   & 	34.68	   \\
Dinka	 & 	---	   & \textbf{	14.27	 }  & 	---	   & 	5.44	   & 	5.02	   & 	11.50	   & 	---	   \\
Hausa	 & 	34.16	   & \textbf{	40.23	 }  & 	32.11	   & 	36.34	   & 	35.12	   & 	36.85	   & 	24.64	   \\
Kanuri	 & 	---	   & \textbf{	24.45	 }  & 	7.25	   & 	12.32	   & 	8.63	   & 	12.29	   & 	---	   \\
Khmer	 & 	38.72	   & \textbf{	45.06	 }  & 	36.71	   & 	42.66	   & 	42.68	   & 	43.71	   & 	34.93	   \\
Kinyarwanda	 & 	25.74	   & 	29.01	   & 	25.36	   & 	26.99	   & 	25.89	   & \textbf{	29.04	 }  & 	18.38	   \\
Kurdish (Kurmanji)	 & 	35.81	   & \textbf{	42.77	 }  & 	29.94	   & 	40.48	   & 	36.26	   & 	41.21	   & 	31.80	   \\
Kurdish (Sorani)	 & 	31.78	   & 	34.03	   & 	21.02	   & 	28.28	   & 	25.76	   & \textbf{	37.56	 }  & 	27.88	   \\
Lingala	 & \textbf{	31.47	 }  & 	26.00	   & 	17.83	   & 	19.71	   & 	18.91	   & 	26.81	   & 	16.77	   \\
Luganda	 & 	30.86	   & \textbf{	35.22	 }  & 	25.46	   & 	28.49	   & 	27.42	   & 	33.85	   & 	22.01	   \\
Nepali	 & 	50.90	   & \textbf{	54.58	 }  & 	43.52	   & 	51.03	   & 	50.73	   & 	50.93	   & 	46.37	   \\
Nigerian Fulfulde	 & 	---	   & \textbf{	26.38	 }  & 	12.03	   & 	17.02	   & 	11.85	   & 	21.31	   & 	---	   \\
Oromo	 & 	---	   & 	31.12	   & 	25.69	   & 	29.40	   & 	27.43	   & \textbf{	31.83	 }  & 	18.99	   \\
Pashto	 & 	32.44	   & \textbf{	43.48	 }  & 	35.37	   & 	40.61	   & 	40.24	   & 	41.33	   & 	32.76	   \\
Somali	 & 	33.73	   & \textbf{	39.69	 }  & 	32.90	   & 	38.20	   & 	36.72	   & 	38.28	   & 	26.00	   \\
Tigrinya (Eritrean)	 & 	22.62	   & 	24.52	   & 	15.88	   & 	25.91	   & 	24.86	   & \textbf{	27.44	 }  & 	17.52	   \\
Tigrinya (Ethiopian)	 & 	27.58	   & 	29.61	   & 	14.00	   & 	31.21	   & 	31.04	   & \textbf{	31.39	 }  & 	21.81	   \\
Zulu	 & 	43.68	   & \textbf{	49.81	 }  & 	41.35	   & 	47.83	   & 	46.45	   & 	47.63	   & 	34.94	   \\

 \hline
 
%Congolese Swahili & \cellcolor{gray}{---} & \cellcolor{gray}{---} & \cellcolor{gray}{---} & \cellcolor{gray}{---} & \cellcolor{gray}{---} & \cellcolor{gray}{---} & \cellcolor{gray}{---} & \cellcolor{gray}{---} \\ \hline 

\end{tabular}
\caption{2026 XE BLEU scores. The best system for each language is in bold. %(See Section~\ref{sec:TICO-data} for TwB definitions of these terms, and Section~\ref{sec:quality-usability-readiness} for our definition of usability.)
}
\label{tab:bleu_XE_2026}
\end{table*}
\clearpage

\begin{table*}
%\begin{tabular}{|*{9}{c|}}
%\small
\begin{tabular}{|c|c c c c c c c |}
\hline
\multicolumn{1}{|c|}{\textbf{Language}} & 
\multicolumn{7}{|c|}{\textbf{X to English}}  \\ \hline
& MS & Google & GPT-4o & GPT-5.4 & GPT-5.4-mini & Gemini & Claude  \\ \hline

%% Pivot Languages first
\multicolumn{1}{|c|}{\textbf{------Pivot------}} & \multicolumn{7}{|c|}{} \\ \hline
Arabic	 & 	68.06	   & 	69.17	   & 	66.88	   & 	69.12	   & 	68.78	   & \textbf{	69.19	 }  & 	66.51	   \\
Chinese	 & 	62.16	   & 	61.87	   & 	60.40	   & \textbf{	62.53	 }  & 	62.35	   & 	62.20	   & 	60.76	   \\
French	 & \textbf{	64.76	 }  & 	64.06	   & 	63.18	   & 	63.97	   & 	63.68	   & 	64.04	   & 	62.67	   \\
Hindi	 & 	73.92	   & \textbf{	76.79	 }  & 	69.39	   & 	74.52	   & 	74.29	   & 	74.13	   & 	72.26	   \\
Indonesian	 & 	72.53	   & \textbf{	74.33	 }  & 	71.53	   & 	73.77	   & 	73.27	   & 	72.70	   & 	71.61	   \\
Portuguese (BR)	 & 	75.86	   & 	75.21	   & 	73.35	   & 	75.77	   & 	75.61	   & \textbf{	75.91	 }  & 	74.37	   \\
Russian	 & 	63.73	   & \textbf{	64.70	 }  & 	62.71	   & 	64.03	   & 	63.54	   & 	64.07	   & 	62.51	   \\
Spanish (LA)	 & \textbf{	75.22	 }  & 	74.70	   & 	74.09	   & 	74.87	   & 	74.47	   & 	75.08	   & 	73.57	   \\
Swahili	 & 	63.97	   & \textbf{	68.14	 }  & 	63.51	   & 	68.10	   & 	66.35	   & 	67.91	   & 	62.19	   \\
 \hline

%% Important languages
%\\ \hline
\multicolumn{1}{|c|}{\textbf{------Important------}} & \multicolumn{7}{|c|}{} \\ \hline    
Bengali	 & 	68.76	   & \textbf{	71.94	 }  & 	66.46	   & 	70.77	   & 	70.21	   & 	70.19	   & 	68.40	   \\
Farsi	 & 	64.64	   & \textbf{	67.55	 }  & 	64.25	   & 	66.21	   & 	65.77	   & 	66.14	   & 	63.45	   \\
Malay	 & 	74.17	   & \textbf{	76.92	 }  & 	71.59	   & 	76.21	   & 	75.56	   & 	74.68	   & 	73.24	   \\
Marathi	 & 	63.75	   & \textbf{	67.19	 }  & 	62.86	   & 	66.43	   & 	65.76	   & 	65.49	   & 	62.58	   \\
Myanmar	 & 	57.02	   & \textbf{	61.79	 }  & 	53.90	   & 	61.43	   & 	60.78	   & 	60.87	   & 	55.82	   \\
Tagalog	 & 	77.01	   & \textbf{	79.52	 }  & 	72.39	   & 	77.33	   & 	77.42	   & 	76.67	   & 	73.85	   \\
Tamil	 & 	62.74	   & \textbf{	66.14	 }  & 	57.57	   & 	64.78	   & 	64.25	   & 	64.72	   & 	61.31	   \\
Urdu	 & 	62.53	   & \textbf{	66.55	 }  & 	61.18	   & 	65.30	   & 	64.48	   & 	64.84	   & 	62.01	   \\
 \hline 

%% Priority languages
%\\ \hline
\multicolumn{1}{|c|}{\textbf{------Priority------}} & \multicolumn{7}{|c|}{} \\ \hline
Amharic	 & 	57.54	   & 	60.98	   & 	51.41	   & 	60.71	   & 	59.82	   & \textbf{	61.11	 }  & 	56.14	   \\
Dari	 & 	63.11	   & \textbf{	65.88	 }  & 	62.54	   & 	65.09	   & 	64.67	   & 	65.36	   & 	62.05	   \\
Dinka	 & 	---	   & \textbf{	33.63	 }  & 	---	   & 	18.71	   & 	17.48	   & 	30.69	   & 	---	   \\
Hausa	 & 	55.40	   & \textbf{	60.16	 }  & 	53.78	   & 	57.75	   & 	56.23	   & 	58.91	   & 	48.57	   \\
Kanuri	 & 	---	   & \textbf{	46.33	 }  & 	21.82	   & 	31.26	   & 	23.53	   & 	32.21	   & 	---	   \\
Khmer	 & 	60.67	   & \textbf{	67.50	 }  & 	60.85	   & 	66.87	   & 	66.43	   & 	66.46	   & 	61.21	   \\
Kinyarwanda	 & 	48.40	   & \textbf{	52.93	 }  & 	48.08	   & 	50.88	   & 	49.87	   & 	52.52	   & 	43.64	   \\
Kurdish (Kurmanji)	 & 	58.73	   & \textbf{	63.61	 }  & 	52.97	   & 	63.19	   & 	58.86	   & 	63.52	   & 	56.32	   \\
Kurdish (Sorani)	 & 	56.53	   & 	57.09	   & 	41.84	   & 	49.45	   & 	46.52	   & \textbf{	60.66	 }  & 	53.37	   \\
Lingala	 & \textbf{	52.21	 }  & 	47.85	   & 	36.95	   & 	39.07	   & 	37.69	   & 	48.86	   & 	40.07	   \\
Luganda	 & 	51.28	   & 	55.21	   & 	45.27	   & 	47.99	   & 	46.41	   & \textbf{	55.49	 }  & 	46.26	   \\
Nepali	 & 	71.96	   & \textbf{	74.38	 }  & 	67.49	   & 	72.92	   & 	72.50	   & 	72.42	   & 	69.83	   \\
Nigerian Fulfulde	 & 	---	   & \textbf{	47.30	 }  & 	30.12	   & 	35.94	   & 	27.56	   & 	42.43	   & 	---	   \\
Oromo	 & 	---	   & 	54.24	   & 	49.91	   & 	52.21	   & 	49.94	   & \textbf{	55.45	 }  & 	43.42	   \\
Pashto	 & 	58.82	   & \textbf{	65.92	 }  & 	59.97	   & 	64.95	   & 	64.14	   & 	64.71	   & 	58.96	   \\
Somali	 & 	54.71	   & \textbf{	59.39	 }  & 	53.90	   & 	59.09	   & 	57.74	   & 	58.96	   & 	49.26	   \\
Tigrinya (Eritrean)	 & 	46.59	   & 	49.32	   & 	37.82	   & 	52.21	   & 	50.89	   & \textbf{	52.64	 }  & 	43.59	   \\
Tigrinya (Ethiopian)	 & 	51.17	   & 	53.70	   & 	36.56	   & \textbf{	56.83	 }  & 	55.97	   & 	56.17	   & 	47.83	   \\
Zulu	 & 	64.51	   & \textbf{	68.42	 }  & 	62.32	   & 	67.13	   & 	65.90	   & 	67.68	   & 	57.93	   \\
 \hline
 
%Congolese Swahili & \cellcolor{gray}{---} & \cellcolor{gray}{---} & \cellcolor{gray}{---} & \cellcolor{gray}{---} & \cellcolor{gray}{---} & \cellcolor{gray}{---} & \cellcolor{gray}{---} & \cellcolor{gray}{---} \\ \hline 

\end{tabular}
\caption{2026 XE CHRF++ scores. The best system for each language is in bold. %(See Section~\ref{sec:TICO-data} for TwB definitions of these terms, and Section~\ref{sec:quality-usability-readiness} for our definition of usability.)
}
\label{tab:CHRF_XE_2026}
\end{table*}
\clearpage

\begin{table*}
%\begin{tabular}{|*{9}{c|}}
%\small
\begin{tabular}{|c|c c c c c c c |}
\hline
\multicolumn{1}{|c|}{\textbf{Language}} & 
\multicolumn{7}{|c|}{\textbf{English to X}}  \\ \hline
& MS & Google & GPT-4o & GPT-5.4 & GPT-5.4-mini & Gemini & Claude  \\ \hline

%% Pivot Languages first
\multicolumn{1}{|c|}{\textbf{------Pivot------}} & \multicolumn{7}{|c|}{} \\ \hline
Arabic	 & 	0.8744	   & 	0.8800	   & 	0.8759	   & \textbf{	0.8830	 }  & 	0.8788	   & 	0.8797	   & 	0.8718	   \\
Chinese	 & 	0.8801	   & \textbf{	0.8882	 }  & 	0.8834	   & 	0.8710	   & 	0.8843	   & 	0.8716	   & 	0.8773	   \\
French	 & 	0.8772	   & 	0.8440	   & 	0.8692	   & \textbf{	0.8801	 }  & 	0.8791	   & 	0.8703	   & 	0.8701	   \\
Hindi	 & \textbf{	0.9038	 }  & 	0.8876	   & 	0.8809	   & 	0.8863	   & 	0.8982	   & 	0.8915	   & 	0.8956	   \\
Indonesian	 & \textbf{	0.9301	 }  & 	0.9249	   & 	0.9283	   & 	0.9292	   & 	0.9297	   & 	0.9273	   & 	0.9258	   \\
Portuguese (BR)	 & 	0.9310	   & 	0.9251	   & 	0.9213	   & 	0.9321	   & 	0.9314	   & \textbf{	0.9332	 }  & 	0.9320	   \\
Russian	 & 	0.8723	   & \textbf{	0.8781	 }  & 	0.8737	   & 	0.8765	   & 	0.8764	   & 	0.8764	   & 	0.8735	   \\
Spanish (LA)	 & 	0.9287	   & 	0.9284	   & 	0.9311	   & \textbf{	0.9348	 }  & 	0.9318	   & 	0.9329	   & 	0.9302	   \\
Swahili	 & 	0.8722	   & \textbf{	0.8737	 }  & 	0.8728	   & 	0.8690	   & 	0.8713	   & 	0.8694	   & 	0.8556	   \\
 \hline

%% Important languages
%\\ \hline
\multicolumn{1}{|c|}{\textbf{------Important------}} & \multicolumn{7}{|c|}{} \\ \hline    
Bengali	 & \textbf{	0.8668	 }  & 	0.8564	   & 	0.8602	   & 	0.8542	   & 	0.8627	   & 	0.8481	   & 	0.8576	   \\
Farsi	 & 	0.8747	   & \textbf{	0.8851	 }  & 	0.8773	   & 	0.8670	   & 	0.8756	   & 	0.8700	   & 	0.8389	   \\
Malay	 & \textbf{	0.9269	 }  & 	0.9236	   & 	0.8855	   & 	0.9234	   & 	0.9264	   & 	0.9176	   & 	0.9154	   \\
Marathi	 & \textbf{	0.8400	 }  & 	0.8362	   & 	0.8296	   & 	0.8330	   & 	0.8347	   & 	0.8327	   & 	0.8229	   \\
Myanmar	 & 	0.8187	   & \textbf{	0.8375	 }  & 	0.7726	   & 	0.8317	   & 	0.8355	   & 	0.8275	   & 	0.7865	   \\
Tagalog	 & \textbf{	0.8910	 }  & 	0.8902	   & 	0.8882	   & 	0.8705	   & 	0.8894	   & 	0.8797	   & 	0.8809	   \\
Tamil	 & \textbf{	0.8673	 }  & 	0.8480	   & 	0.8159	   & 	0.8403	   & 	0.8369	   & 	0.8433	   & 	0.8293	   \\
Urdu	 & 	0.8582	   & \textbf{	0.8613	 }  & 	0.8464	   & 	0.8484	   & 	0.8558	   & 	0.8473	   & 	0.8513	   \\
 \hline 

%% Priority languages
%\\ \hline
\multicolumn{1}{|c|}{\textbf{------Priority------}} & \multicolumn{7}{|c|}{} \\ \hline
Amharic	 & 	0.9508	   & 	0.9544	   & 	0.9545	   & \textbf{	0.9588	 }  & 	0.9576	   & 	0.9569	   & 	0.9330	   \\
Dari	 & \textbf{	0.8537	 }  & 	---	   & 	0.8161	   & 	0.8388	   & 	0.8441	   & 	0.8433	   & 	0.8381	   \\
Dinka	 & 	---	   & \textbf{	0.7552	 }  & 	---	   & 	0.7328	   & 	0.7265	   & 	---	   & 	---	   \\
Hausa	 & \textbf{	0.8379	 }  & 	0.8206	   & 	0.8251	   & 	0.8173	   & 	0.8257	   & 	0.8149	   & 	0.8022	   \\
Kanuri	 & 	---	   & \textbf{	0.7318	 }  & 	---	   & 	0.6999	   & 	0.6888	   & 	---	   & 	---	   \\
Khmer	 & 	0.9488	   & 	0.9413	   & 	0.9373	   & 	0.9270	   & \textbf{	0.9507	 }  & 	0.9264	   & 	0.9118	   \\
Kinyarwanda	 & 	0.7959	   & 	0.7995	   & 	0.7799	   & \textbf{	0.8140	 }  & 	0.8109	   & 	0.8128	   & 	0.7551	   \\
Kurdish (Kurmanji)	 & \textbf{	0.8161	 }  & 	0.8118	   & 	0.7862	   & 	---	   & 	0.8038	   & 	---	   & 	0.7909	   \\
Kurdish (Sorani)	 & \textbf{	0.8648	 }  & 	0.8615	   & 	0.8463	   & 	0.8540	   & 	0.8509	   & 	0.8570	   & 	0.8311	   \\
Lingala	 & \textbf{	0.8667	 }  & 	0.7822	   & 	0.7922	   & 	0.7940	   & 	0.7932	   & 	0.7926	   & 	0.7668	   \\
Luganda	 & 	0.7829	   & \textbf{	0.8095	 }  & 	0.7797	   & 	0.7921	   & 	0.7904	   & 	0.7886	   & 	0.7386	   \\
Nepali	 & 	0.8676	   & \textbf{	0.8754	 }  & 	0.8409	   & 	0.8527	   & 	0.8668	   & 	0.8566	   & 	0.8614	   \\
Nigerian Fulfulde	 & 	---	   & 	0.7300	   & 	---	   & \textbf{	0.7396	 }  & 	0.7295	   & 	0.7131	   & 	---	   \\
Oromo	 & 	---	   & \textbf{	0.8754	 }  & 	0.7681	   & 	0.7940	   & 	0.7874	   & 	0.7875	   & 	0.7518	   \\
Pashto	 & 	0.8200	   & \textbf{	0.8421	 }  & 	0.8232	   & 	0.8248	   & 	0.8285	   & 	0.8175	   & 	0.7847	   \\
Somali	 & 	0.8211	   & \textbf{	0.8216	 }  & 	0.7516	   & 	0.8081	   & 	0.8146	   & 	0.8092	   & 	0.7823	   \\
Tigrinya (Eritrean)	 & 	0.9483	   & 	---	   & 	---	   & 	0.9514	   & \textbf{	0.9523	 }  & 	0.9466	   & 	---	   \\
Tigrinya (Ethiopian)	 & 	0.9542	   & 	0.9462	   & 	---	   & 	0.9533	   & \textbf{	0.9563	 }  & 	0.9510	   & 	---	   \\
Zulu	 & \textbf{	0.8478	 }  & 	0.8385	   & 	0.7969	   & 	0.8330	   & 	0.8394	   & 	0.8269	   & 	0.7982	   \\
 \hline
 
%Congolese Swahili & \cellcolor{gray}{---} & \cellcolor{gray}{---} & \cellcolor{gray}{---} & \cellcolor{gray}{---} & \cellcolor{gray}{---} & \cellcolor{gray}{---} & \cellcolor{gray}{---} & \cellcolor{gray}{---} \\ \hline 

\end{tabular}
\caption{2026 EX BERTScores. The best system for each language is in bold. %(See Section~\ref{sec:TICO-data} for TwB definitions of these terms, and Section~\ref{sec:quality-usability-readiness} for our definition of usability.)
}
\label{tab:BERTScore_EX_2026}
\end{table*}
\clearpage

\begin{table*}
%\begin{tabular}{|*{9}{c|}}
%\small
\begin{tabular}{|c|c c c c c c c |}
\hline
\multicolumn{1}{|c|}{\textbf{Language}} & 
\multicolumn{7}{|c|}{\textbf{X to English}}  \\ \hline
& MS & Google & GPT-4o & GPT-5.4 & GPT-5.4-mini & Gemini & Claude  \\ \hline

%% Pivot Languages first
\multicolumn{1}{|c|}{\textbf{------Pivot------}} & \multicolumn{7}{|c|}{} \\ \hline
Arabic	 & 	0.7486	   & 	0.7416	   & 	0.7406	   & 	0.7514  & 	0.7506	   & \textbf{0.7641}	   &	0.7285   \\
Chinese	 & 0.7148  & 	0.6959	   & 	-0.5152	   & 	0.7098	   & 	0.7118	   & \textbf{0.7160}	   & 0.6920	   \\
French	 & \textbf{	0.6865	 }  & 	0.6627	   & 	0.6702	   & 	0.6762	   & 	0.6744	   & 	0.6810   & 	 0.6700  \\
Hindi	 & 	0.8113	   & \textbf{	0.8225	 }  & 	-0.7376	   & 	0.8118	   & 	0.8162	   & 	0.8206   & 	0.7948   \\
Indonesian	 & 	0.8087	   & 	0.8108	   & 	0.8005	   & 0.8155  & 	0.8099	   & 	\textbf{0.8167 }  & 	0.8030   \\
Portuguese (BR)	 & 	0.8307	   & 	0.8058	   & 	-0.0443	   & 0.8332	   & 	0.8317	   & \textbf{0.8343}	   & 	0.8181   \\
Russian	 & 0.7173  & 	0.7133	   & 	0.7018	   & 	0.7096	   & 	0.7058	   & \textbf{0.7233}	   & 	0.7041   \\
Spanish (LA)	 & 0.8199  & 	0.7965	   & 	0.8134	   & 	0.8191	   & 	0.8166	   & \textbf{0.8204}	   & 0.8057	   \\
Swahili	 & 	0.7236	   & 0.7574  & 	-0.4285	   & 	0.7447	   & 	0.7310	   & 	\textbf{0.7756}   & 	0.7111   \\
 \hline

%% Important languages
%\\ \hline
\multicolumn{1}{|c|}{\textbf{------Important------}} & \multicolumn{7}{|c|}{} \\ \hline    
Bengali	 & 	0.7711	   & 0.7744  & 	0.7398	   & 	0.7665	   & 	0.7661	   &	\textbf{0.7919}   & 	0.7605   \\
Farsi	 & 	0.7426	   & 0.7535  & 	0.7395	   & 	0.7467	   & 	0.7470	   & 	\textbf{0.7611}   & 	0.7237   \\
Malay	 & 	0.8220	   & 	0.8267	   & 	-0.4093	   & 0.8331	  & 	0.8285	   & \textbf{0.8344}	   & 	0.8141   \\
Marathi	 & 	0.7319	   & 0.7456	  & 	0.7208	   & 	0.7436	   & 	0.7400	   & 	\textbf{0.7499}   & 	0.7059   \\
Myanmar	 & 	0.6585	   & 	0.6929	   & 	-0.7524	   & 0.6982  & 	0.6908	   & 	\textbf{0.7001}   & 	0.6259   \\
Tagalog	 & 	0.8380	   & 	0.8394	   & 	-0.3387	   & 	0.8471	   & \textbf{	0.8472	 }  & 0.8468	   & 	0.8079   \\
Tamil	 & 	0.7033	   & 0.7126  & 	-0.9708	   & 	0.6792	   & 	0.6738	   & \textbf{0.7322}	   & 	0.6792	   \\
Urdu	 & 	0.7104	   & 0.7375  & 	-0.5403	   & 	0.7314	   & 	0.7273	   & 	\textbf{0.7421}   & 	0.7019   \\
 \hline 

%% Priority languages
%\\ \hline
\multicolumn{1}{|c|}{\textbf{------Priority------}} & \multicolumn{7}{|c|}{} \\ \hline
Amharic	 & 	0.6584	   & 	0.6944	   & 	0.6055	   &  0.6933 & 	0.6915	   & \textbf{0.7103} & 	0.6346   \\
Dari	 & 	0.7166	   & 	0.7214	   & 	-0.5382	   &	0.7306  & 	0.7261	   & 	\textbf{0.7390}   & 	0.7016   \\
Dinka	 & 	---	   & \textbf{	0.3791	 }  & 	---	   & 	-1.2368	   & 	-1.4992	   & 	0.3156   & 	 ---  \\
Hausa	 & 	0.6388	   & 0.6850  & 	0.3746	   & 	0.4153	   & 	0.3907	   & \textbf{0.6986}	   & 0.5458	   \\
Kanuri	 & 	---	   & \textbf{	0.4763	 }  & 	-0.7413	   & 	0.2510	   & 	-0.7590	   & 0.2883	   & 	---   \\
Khmer	 & 	0.6914	   & 	0.7431	   & 	-0.8413	   & 0.7518  & 	0.7496	   &	\textbf{0.7563}   &	0.6946   \\
Kinyarwanda	 & 	0.5777	   & 0.6236  & 	-0.6382	   & 	0.4456	   & 	0.4462	   & \textbf{0.6426}	   & 	0.5005   \\
Kurdish (Kurmanji)	 & 	0.6673	   & 	0.7046	   & 	0.4258	   & 0.7096  & 	0.4806	   & 	\textbf{0.7238}   & 	0.6259   \\
Kurdish (Sorani)	 & 	0.6267	   & 0.6448  & 	-0.5270	   & 	-0.3058	   & 	-0.4956	   & 	\textbf{0.6883}   & 0.5896	   \\
Lingala	 & \textbf{	0.6376	 }  & 	0.5847	   & 	-0.9227	   & 	-0.8905	   & 	-0.8785	   & 0.6056	   & 	0.4390   \\
Luganda	 & 	0.6038	   & 0.6555	  & 	-0.3596	   & 	-0.3446	   & 	-0.3695	   & 	\textbf{0.6655}   & 	 0.5135  \\
Nepali	 & 	0.7943	   & 0.8004  & 	-0.7168	   & 	0.7972	   & 	0.7978	   & 	\textbf{0.8091}   & 0.7706	   \\
Nigerian Fulfulde	 & 	---	   & \textbf{	0.5271	 }  & 	-0.5379	   & 	0.1528	   & 	-1.0359	   & 	0.4766   & 	---   \\
Oromo	 & 	---	   & 0.6211  & 	-0.5438	   & 	0.2383	   & 	0.1954	   & 	\textbf{0.6474}   & 	0.4658   \\
Pashto	 & 	0.6427	   & 0.7312  & 	-0.5797	   & 	0.7239	   & 	0.7172	   & 	\textbf{0.7379}   & 	0.6623   \\
Somali	 & 	0.6139	   & 0.6601  & 	0.6208	   & 	0.6291	   & 	0.6148	   & 	\textbf{0.6667}   & 	0.5355   \\
Tigrinya (Eritrean)	 & 	0.5596	   & 	0.5740	   & 	0.4233	   & 	0.6063	   & 	0.5983	   & \textbf{0.6208} & 0.4677	   \\
Tigrinya (Ethiopian)	 & 	0.6141	   & 	0.6262	   & 	-0.7240	   & 	0.6680	   & 	0.6628	   & \textbf{0.6706}  & 	0.5311   \\
Zulu	 & 	0.7397	   & 0.7619  & 	-0.4587	   & 	0.6760	   & 	0.6737	   & \textbf{0.7823}	   & 	0.6637   \\
 \hline
 
%Congolese Swahili & \cellcolor{gray}{---} & \cellcolor{gray}{---} & \cellcolor{gray}{---} & \cellcolor{gray}{---} & \cellcolor{gray}{---} & \cellcolor{gray}{---} & \cellcolor{gray}{---} & \cellcolor{gray}{---} \\ \hline 

\end{tabular}
\caption{2026 XE BERTScores (rescaled). The best system for each language is in bold. %(See Section~\ref{sec:TICO-data} for TwB definitions of these terms, and Section~\ref{sec:quality-usability-readiness} for our definition of usability.)
}
\label{tab:BERTScore_XE_2026}
\end{table*}
\clearpage

\begin{table*}
%\begin{tabular}{|*{9}{c|}}
%\small
\begin{tabular}{|c|c c c c c c c |}
\hline
\multicolumn{1}{|c|}{\textbf{Language}} & 
\multicolumn{7}{|c|}{\textbf{English to X}}  \\ \hline
& MS & Google & GPT-4o & GPT-5.4 & GPT-5.4-mini & Gemini & Claude  \\ \hline

%% Pivot Languages first
\multicolumn{1}{|c|}{\textbf{------Pivot------}} & \multicolumn{7}{|c|}{} \\ \hline
Arabic	 & 	0.8485	   & 	0.8750	   & 	0.8683	   & \textbf{	0.8764	 }  & 	0.8688	   & 	0.8761	   & 	0.8649	   \\
Chinese	 & 	0.8749	   & \textbf{	0.8953	 }  & 	0.8840	   & 	0.8753	   & 	0.8862	   & 	0.8779	   & 	0.8804	   \\
French	 & 	0.8154	   & 	0.8242	   & 	0.8199	   & \textbf{	0.8245	 }  & 	0.8229	   & 	0.8239	   & 	0.8199	   \\
Hindi	 & 	0.8076	   & 	0.8203	   & 	0.8064	   & 	0.8061	   & \textbf{	0.8217	 }  & 	0.8177	   & 	0.8148	   \\
Indonesian	 & 	0.9189	   & \textbf{	0.9254	 }  & 	0.9234	   & 	0.9251	   & 	0.9237	   & 	0.9253	   & 	0.9212	   \\
Portuguese (BR)	 & 	0.8994	   & 	0.9037	   & 	0.8965	   & 	0.9041	   & 	0.9020	   & \textbf{	0.9053	 }  & 	0.9022	   \\
Russian	 & 	0.8775	   & \textbf{	0.9027	 }  & 	0.8889	   & 	0.8939	   & 	0.8898	   & 	0.8996	   & 	0.8898	   \\
Spanish (LA)	 & 	0.8839	   & \textbf{	0.8932	 }  & 	0.8886	   & 	0.8923	   & 	0.8879	   & 	0.8903	   & 	0.8880	   \\
Swahili	 & 	0.8236	   & 	0.8392	   & 	0.8406	   & 	0.8431	   & 	0.8400	   & \textbf{	0.8448	 }  & 	0.8139	   \\
 \hline

%% Important languages
%\\ \hline
\multicolumn{1}{|c|}{\textbf{------Important------}} & \multicolumn{7}{|c|}{} \\ \hline    
Bengali	 & 	0.8513	   & \textbf{	0.8749	 }  & 	0.8624	   & 	0.8629	   & 	0.8686	   & 	0.8634	   & 	0.8565	   \\
Farsi	 & 	0.8782	   & \textbf{	0.8901	 }  & 	0.8822	   & 	0.8716	   & 	0.8825	   & 	0.8806	   & 	0.8476	   \\
Malay	 & 	0.9071	   & 	0.9065	   & 	0.8838	   & 	0.9074	   & \textbf{	0.9086	 }  & 	0.9068	   & 	0.9016	   \\
Marathi	 & 	0.7216	   & \textbf{	0.7460	 }  & 	0.7240	   & 	0.7366	   & 	0.7332	   & 	0.7409	   & 	0.7133	   \\
Myanmar	 & 	0.8719	   & 	0.8849	   & 	0.8407	   & 	0.8946	   & 	0.8906	   & \textbf{	0.8961	 }  & 	0.8016	   \\
Tagalog	 & 	0.8672	   & 	0.8682	   & \textbf{	0.8685	 }  & 	0.8449	   & 	0.8655	   & 	0.8638	   & 	0.8543	   \\
Tamil	 & 	0.8751	   & \textbf{	0.8911	 }  & 	0.8503	   & 	0.8787	   & 	0.8706	   & 	0.8851	   & 	0.8409	   \\
Urdu	 & 	0.8230	   & 	0.8172	   & 	0.8181	   & 	0.8210	   & \textbf{	0.8282	 }  & 	0.8247	   & 	0.8194	   \\
 \hline 

%% Priority languages
%\\ \hline
\multicolumn{1}{|c|}{\textbf{------Priority------}} & \multicolumn{7}{|c|}{} \\ \hline
Amharic	 & 	0.8533	   & 	0.8656	   & 	0.7891	   & 	0.8612	   & 	0.8526	   & \textbf{	0.8696	 }  & 	0.8057	   \\
Hausa	 & 	0.7993	   & 	0.8087	   & 	0.8053	   & 	0.8060	   & 	0.8052	   & \textbf{	0.8108	 }  & 	0.7592	   \\
Khmer	 & 	0.8560	   & 	0.8734	   & 	0.8364	   & 	0.8683	   & 	0.8662	   & \textbf{	0.8744	 }  & 	0.7787	   \\
Kinyarwanda	 & 	0.6271	   & 	0.6473	   & 	0.6327	   & \textbf{	0.6549	 }  & 	0.6496	   & 	0.6536	   & 	0.6060	   \\
Kurdish (Kurmanji)	 & 	0.8033	   & \textbf{	0.8116	 }  & 	0.7359	   & 	---	   & 	0.7917	   & 	---	   & 	0.7630	   \\
Nepali	 & 	0.8159	   & 0.8358	   & 	0.8112	   & 	0.8203	   & 	\textbf{0.8368}	   & 	0.8328	   & 	0.8227	   \\
Oromo	 & 	---	   & \textbf{	0.8358	 }  & 	0.7683	   & 	0.8040	   & 	0.7917	   & 	0.7995	   & 	0.7507	   \\
Pashto	 & 	0.7877	   & \textbf{	0.8128	 }  & 	0.7999	   & 	0.8115	   & 	0.8120	   & 	0.8106	   & 	0.7561	   \\
Somali	 & 	0.8008	   & \textbf{	0.8070	 }  & 	0.6765	   & 	0.7950	   & 	0.7996	   & 	0.8029	   & 	0.7389	   \\
 \hline
 
%Congolese Swahili & \cellcolor{gray}{---} & \cellcolor{gray}{---} & \cellcolor{gray}{---} & \cellcolor{gray}{---} & \cellcolor{gray}{---} & \cellcolor{gray}{---} & \cellcolor{gray}{---} & \cellcolor{gray}{---} \\ \hline 

\end{tabular}
\caption{2026 EX COMET. The best system for each language is in bold. Only supported languages are included. %(See Section~\ref{sec:TICO-data} for TwB definitions of these terms, and Section~\ref{sec:quality-usability-readiness} for our definition of usability.)
}
\label{tab:COMET_EX_2026}
\end{table*}
\clearpage

\begin{table*}
%\begin{tabular}{|*{9}{c|}}
%\small
\begin{tabular}{|c|c c c c c c c |}
\hline
\multicolumn{1}{|c|}{\textbf{Language}} & 
\multicolumn{7}{|c|}{\textbf{X to English}}  \\ \hline
& MS & Google & GPT-4o & GPT-5.4 & GPT-5.4-mini & Gemini & Claude  \\ \hline

%% Pivot Languages first
\multicolumn{1}{|c|}{\textbf{------Pivot------}} & \multicolumn{7}{|c|}{} \\ \hline
Arabic	 & 	0.8725	   & 	0.8779	   & 	0.8765	   & 0.8788	  & 	0.8784	   & \textbf{0.8820}	   & 0.8716	   \\
Chinese	 & 	0.8751	   & 	0.8765	   & 	0.6173	   & 	0.8778	   & 	0.8781	   & \textbf{0.8790}	   & 0.8728  \\
French	 & 	0.8360	   & 	0.8353	   & 	0.8352	   & \textbf{	0.8370	 }  & 	0.8362	   & 	0.8367   & 	0.8334   \\
Hindi	 & 	0.9070	   & \textbf{	0.9150	 }  & 	0.4515	   & 	0.9121	   & 	0.9125	   &	0.9135   &	0.9075   \\
Indonesian	 & 	0.9032	   & 	\textbf{0.9086}	   & 	0.9047	   & 	0.9079	   & 	0.9067	   &  0.9072 &	0.9036   \\
Portuguese (BR)	 & 	0.9018	   & 	0.9009	   & 	0.7162	   & 	\textbf{0.9042}	   & 	0.9039	   & 0.9041  & 	0.8990   \\
Russian	 & 	0.8637	   & 	\textbf{0.8684}	   & 	0.8640	   & 	0.8664	   & 	0.8652	   & \textbf{0.8684}  &	 0.8628  \\
Spanish (LA)	 & 	0.8981	   & 	0.8966	   & 	0.8975	   & 0.8989	   & 	0.8984	   & \textbf{0.8995}	   & 0.8948	   \\
Swahili	 & 	0.8459	   & 	0.8679	   & 	0.5409	   &	0.8685	   & 	0.8633	   & 	\textbf{0.8722}   & 	 0.8472  \\
 \hline

%% Important languages
%\\ \hline
\multicolumn{1}{|c|}{\textbf{------Important------}} & \multicolumn{7}{|c|}{} \\ \hline    
Bengali	 & 	0.8995	   & 	0.9048	   & 	0.9005	   & 0.9062  & 0.9054	   & 	\textbf{0.9087 }  & 	0.9016   \\
Farsi	 & 	0.8846	   & 0.8924  & 	0.8889	   & 	0.8906	   & 	0.8901	   & \textbf{0.8930}  & 	0.8823   \\
Malay	 & 	0.8984	   & 	0.9061	   & 	0.6778	   & 	\textbf{0.9067}	   & 	0.9051	   &  0.9060 & 	0.8992   \\
Marathi	 & 	0.8806	   & \textbf{	0.8903	 }  & 	0.8838	   & 	0.8894	   & 	0.8879	   & 	0.8899   & 	0.8778   \\
Myanmar	 & 	0.8696	   & 	0.8876	   & 	0.3951	   & 	0.8894	   & 	0.8868	   & \textbf{0.8897}  &	 0.8689  \\
Tagalog	 & 	0.8899	   & 	0.8970	   & 	0.6188	   & \textbf{	0.8997	 }  & 	0.8982	   & 	0.8988   & 	0.8850   \\
Tamil	 & 	0.8710	   & 	0.8816	   & 	0.4526	   & 	0.8798	   & 	0.8784	   & \textbf{0.8869} & 	0.8731   \\
Urdu	 & 	0.8694	   & 0.8854  & 	0.4548	   & 	0.8838	   & 	0.8816	   & 	\textbf{0.8856}   & 	0.8743   \\
 \hline 

%% Priority languages
%\\ \hline
\multicolumn{1}{|c|}{\textbf{------Priority------}} & \multicolumn{7}{|c|}{} \\ \hline
Amharic	 & 	0.8569	   & 	0.8782	   & 	0.8486	   & 0.8784  & 	0.8748	   & \textbf{0.8813	}   & 	0.8576   \\
Hausa	 & 	0.7886	   & 0.8243  & 	0.7936	   & 	0.8064	   & 	0.7956	   &	\textbf{0.8302}   &	0.7569   \\
Khmer	 & 	0.8647	   & 	0.8910	   & 	0.4306	   & 0.8938  & 	0.8923	   & 	\textbf{0.8944}   & 	0.8750   \\
Kinyarwanda	 & 	0.7233	   & 0.7621  & 	0.2931	   & 	0.7519	   & 	0.7460	   & \textbf{0.7693}	   & 	0.6950   \\
Kurdish (Kurmanji)	 & 	0.8139	   & 	0.8403	   & 	0.7938	   & 0.8439  & 	0.8131	   & 	\textbf{0.8468}   & 	0.8063   \\
Nepali	 & 	0.9113	   & 	0.9173	   & 	0.4189	   & 0.9175  & 	0.9170	   & 	\textbf{0.9188}   & 	0.9103   \\
Oromo	 & 	---	   & 	0.7876	   & 	0.3212	   & 	0.7731	   & 	0.7595	   & \textbf{0.8080}  & 	0.7250   \\
Pashto	 & 	0.8499	   & 0.8752  & 	0.3743	   & 	0.8749	   & 	0.8716	   & 	\textbf{0.8765}   & 0.8532	   \\
Somali	 & 	0.7857	   & \textbf{	0.8250	 }  & 	0.8101	   & 	0.8182	   & 	0.8105	   & 	0.6990   & 	0.7607   \\
 \hline
 
%Congolese Swahili & \cellcolor{gray}{---} & \cellcolor{gray}{---} & \cellcolor{gray}{---} & \cellcolor{gray}{---} & \cellcolor{gray}{---} & \cellcolor{gray}{---} & \cellcolor{gray}{---} & \cellcolor{gray}{---} \\ \hline 

\end{tabular}
\caption{2026 XE COMET. The best system for each language is in bold. Only supported languages are included.%(See Section~\ref{sec:TICO-data} for TwB definitions of these terms, and Section~\ref{sec:quality-usability-readiness} for our definition of usability.)
}
\label{tab:COMET_XE_2026}
\end{table*}
\clearpage

%\input{tables/2026_EX_bleuchrf}
%\clearpage

%\input{tables/bleu_and_usability}

%\clearpage
%\input{tables/chrf_2025}

%\clearpage

%\input{tables/BERTScore_2025}

%\clearpage
%\input{tables/COMET_2025}

\clearpage
\begin{table*}
\centering
%\begin{tabular}{|*{9}{c|}}
\begin{tabular}{|c|c c c c|c c c c|}
\hline
\multicolumn{1}{|c|}{} & 
\multicolumn{4}{|c|}{\textbf{English to X}} & \multicolumn{4}{|c|}{\textbf{X to English}} \\ \hline
\multicolumn{1}{|c|}{\textbf{Language}} & \multicolumn{2}{|c|}{\textbf{Microsoft}} & \multicolumn{2}{|c|}{\textbf{Google}} & \multicolumn{2}{|c|}{\textbf{Microsoft}} & \multicolumn{2}{|c|}{\textbf{Google}} \\ \hline
& 2023 & 2025 & 2023 & 2025 & 2023 & 2025 & 2023 & 2025 \\ \hline

%% Pivot Languages first
\multicolumn{1}{|c|}{\textbf{------Pivot------}} & \multicolumn{8}{|c|}{} \\ \hline
Arabic	&	30.26	&	25.16	&	29.15	&	32.88	&	44.88	&	44.87	&	46.90	&	49.28	\\	
Portuguese (BR)	&	52.42	&	47.83	&	52.27	&	54.26	&	54.81	&	55.33	&	55.19	&	57.63	\\	
Chinese	&	52.89	&	50.24	&	54.29	&	54.27	&	36.46	&	36.48	&	37.71	&	39.56	\\	
French	&	45.61	&	40.35	&	34.14	&	45.42	&	45.25	&	45.32	&	44.25	&	46.03	\\	
Hindi	&	43.24	&	40.27	&	46.00	&	43.48	&	53.17	&	53.28	&	56.24	&	58.80	\\	
Indonesian	&	54.39	&	46.85	&	55.07	&	54.82	&	51.73	&	50.29	&	51.15	&	54.61	\\	
Spanish (LA)	&	56.06	&	51.33	&	56.86	&	57.12	&	54.75	&	54.59	&	54.15	&	55.88	\\	
Russian	&	36.88	&	32.95	&	36.71	&	37.35	&	36.06	&	36.84	&	37.14	&	40.14	\\	
Swahili	&	33.10	&	30.74	&	33.88	&	33.87	&	41.04	&	41.92	&	45.56	&	49.21	\\	\hline

%% Important languages
%\\ \hline
\multicolumn{1}{|c|}{\textbf{------Important------}} & \multicolumn{8}{|c|}{} \\ \hline    
Bengali	&	19.29	&	17.18	&	22.95	&	21.14	&	44.38	&	46.08	&	49.46	&	51.33	\\	
Farsi	&	27.38	&	25.25	&	29.20	&	29.88	&	38.97	&	39.35	&	39.65	&	44.45	\\	
Malay	&	51.06	&	45.78	&	53.33	&	53.31	&	50.88	&	54.08	&	57.63	&	59.70	\\	
Marathi	&	15.52	&	13.53	&	16.75	&	16.09	&	37.21	&	37.34	&	42.30	&	44.43	\\	
Myanmar	&	53.55	&	56.15	&	56.95	&	56.62	&	23.82	&	31.44	&	32.39	&	37.33	\\	
Tagalog	&	---	&	49.40	&	48.46	&	48.82	&	---	&	59.92	&	---	&	65.55	\\	
Tamil	&	24.17	&	24.84	&	21.53	&	22.76	&	34.21	&	40.73	&	39.38	&	43.11	\\	
Urdu	&	23.50	&	20.41	&	28.14	&	28.12	&	35.93	&	35.88	&	40.61	&	43.82	\\	\hline

%% Priority languages
\multicolumn{1}{|c|}{\textbf{------Priority------}} & \multicolumn{8}{|c|}{} \\ \hline  
Amharic	&	8.67	&	7.04	&	11.88	&	11.79	&	26.49	&	33.02	&	36.51	&	38.60	\\	
Dari	&	17.32	&	15.21	&	---	&	---	&	31.21	&	31.22	&	---	&	---	\\	
Dinka	&	---	&	---	&	---	&	4.54	&	---	&	---	&	---	&	14.47	\\	
Eritrean Tigrinya	&	3.39	&	3.47	&	6.13	&	6.13	&	18.07	&	18.00	&	24.83	&	---	\\	
Ethiopian Tigrinya	&	3.83	&	3.98	&	6.64	& 6.69	&	21.60	&	21.52	&	30.33	&	29.94	\\	
Hausa	&	15.98	&	25.23	&	25.87	&	28.17	&	---	&	33.92	&	36.17	&	40.65	\\	
Kanuri	&	---	&	---	&	---	&	5.51	&	---	&	---	&	---	&	26.08	\\	
Khmer	&	61.20	&	65.93	&	66.50	&	66.08	&	27.41	&	37.04	&	36.31	&	45.76	\\	
Kinyarwanda	&	---	&	18.75	&	---	&	17.88	&	---	&	25.39	&	---	&	29.49	\\	
Kurdish Kurmanji	&	---	&	11.82	&	---	&	10.05	&	---	&	28.17	&	---	&	28.80	\\	
Kurdish Sorani	&	---	&	10.38	&	10.66	&	10.85	&	---	&	31.87	&	34.18	&	34.13	\\	
Lingala	&	32.07	&	28.35	&	15.08	&	15.17	&	---	&	29.14	&	26.04	&	26.22	\\	
Luganda	&	---	&	17.26	&	19.71	&	20.64	&	---	&	30.52	&	---	&	35.45	\\	
Nepali	&	16.71	&	14.20	&	25.44	&	23.56	&	48.38	&	48.38	&	53.76	&	55.16	\\	
Nigerian Fulfulde	&	---	&	---	&	---	&	3.70	&	---	&	---	&	---	&	26.97	\\	
Oromo	&	---	&	---	&	9.06	&	10.23	&	---	&	---	&	31.36	&	31.52	\\	
Pashto	&	21.12	&	12.79	&	26.09	&	24.63	&	27.57	&	32.16	&	40.75	&	44.07	\\	
Somali	&	26.01	& 26.02	& 17.90	& 17.56 & 33.03	& 33.12	& 37.61	&	19.99 \\	
Zulu	&	19.76	&	20.28	&	19.60	&	19.71	&	40.26	&	43.40	&	47.58	&	50.63	\\	\hline

\end{tabular}
\caption{BLEU Score Differences between 2023 and 2025 for Microsoft and Google}
\label{tab:2023_2025_BLEU_diffs}
\end{table*}

\clearpage
\begin{table*}
\centering
%\begin{tabular}{|*{9}{c|}}
\begin{tabular}{|c|c c c c|c c c c|}
\hline
\multicolumn{1}{|c|}{} & 
\multicolumn{4}{|c|}{\textbf{English to X}} & \multicolumn{4}{|c|}{\textbf{X to English}} \\ \hline
\multicolumn{1}{|c|}{\textbf{Language}} & \multicolumn{2}{|c|}{\textbf{Microsoft}} & \multicolumn{2}{|c|}{\textbf{Google}} & \multicolumn{2}{|c|}{\textbf{Microsoft}} & \multicolumn{2}{|c|}{\textbf{Google}} \\ \hline
& 2023 & 2025 & 2023 & 2025 & 2023 & 2025 & 2023 & 2025 \\ \hline

%% Pivot Languages first
\multicolumn{1}{|c|}{\textbf{------Pivot------}} & \multicolumn{8}{|c|}{} \\ \hline
Arabic	& 57.73	& 56.41	& 56.70	& 59.18 & 67.74	& 67.73	& 69.61 & 70.87 \\	
Portuguese (BR)	& 72.79	& 71.12	& 72.48	& 74.04	& 75.43	& 75.88	& 75.45	& 76.79 \\	
Chinese	& 39.76	& 45.07 & 48.72	& 45.33	& 62.08	& 62.69	& 39.76	& 64.60	\\	
French	& 64.72	& 64.32	& 62.09	& 65.16	& 64.50 & 64.74	& 64.02	& 65.25	\\	
Hindi	& 62.72	& 64.56	& 65.04	& 64.83	& 73.50	& 73.67	& 75.63 & 77.15 \\	
Indonesian	& 74.32	& 72.03	& 74.28	& 75.10	& 72.89	& 72.53	& 72.73	& 74.71 \\	
Spanish (LA)	& 74.39	& 73.16	& 74.47	& 75.30	& 75.13	& 75.33		& 74.97	& 75.48	\\	
Russian	& 59.89	& 58.91	& 59.56	& 60.61	& 63.50	& 63.75	& 63.73	& 65.27	\\	
Swahili	& 58.41	& 57.99	& 59.18	& 59.34 & 62.44	& 63.95	& 66.22	& 68.58	\\	\hline

%% Important languages
%\\ \hline
\multicolumn{1}{|c|}{\textbf{------Important------}} & \multicolumn{8}{|c|}{} \\ \hline    
Bengali	& 49.04	& 48.5	& 52.4	& 52.0	& 68.01	& 69.17	& 70.84	& 72.53	\\	
Farsi	& 53.52	& 53.58	& 54.69	& 56.49	& 63.61	& 64.57	& 64.84 & 67.88	\\	
Malay	& 72.78	& 71.73	& 73.76	& 74.0	& 71.99	& 74.15	& 76.12	& 77.53	\\	
Marathi	& 46.34	& 44.68	& 46.08	& 46.40	& 62.3	& 62.34	& 65.83	& 67.53	\\	
Myanmar	& 37.89	& 39.86	& 43.36	& 43.21	& 50.36	& 57.57	& 59.35 & 62.10	\\
Tagalog	& ---	& 70.41	& 69.36	& 69.93	& ---	& 77.06	& ---	& 80.40	\\	
Tamil	& 56.68 & 57.22 & 54.18	& 56.13	& 61.83	& 64.78 & 63.9 & 66.76	\\	
Urdu	& 50.03	& 48.96	& 52.99	& 53.15	& 60.91	& 60.9 &	64.76 & 66.89	\\	\hline

%% Priority languages
\multicolumn{1}{|c|}{\textbf{------Priority------}} & \multicolumn{8}{|c|}{} \\ \hline  
Amharic	& 30.12	& 29.61	& 33.92	& 33.83	& 51.10	& 57.00 & 59.63	& 61.38	\\	
Dari	& 42.20 & 41.55	&	---	&	---	& 57.44	& 57.45	& --- &	---	\\	
Dinka	&	---	&	---	&	---	& 25.37	&	---	&	---	&	---	&	33.82 \\	
Eritrean Tigrinya	& 19.30 & 19.31	& 25.00 & 25.12	& 42.98	& 42.93	&	49.69 & ---	\\	
Ethiopian Tigrinya	& 19.49	& 19.52 & 25.81	& 25.99	& 46.29	& 46.36	&	54.67 &	54.19 \\	
Hausa	& 40.87	& 52.66	& 52.52 & 52.71	& --- &	55.44 &	57.37 &	60.58 \\	
Kanuri	&	---	&	---	&	---	&	&	---	&	---	&	---	& 30.16	\\	
Khmer	& 45.82	& 52.38	&	53.10 & 52.97	& 53.7	& 58.69	& 60.26	& 67.88	\\	
Kinyarwanda	&	---	&	45.85 &	---	& 45.18	&	---	& 48.35	&	---	&	53.13 \\	
Kurdish Kurmanji &	---	& 39.87	&	---	& 33.49	&	---	& 51.81 & ---	& 47.14	\\	
Kurdish Sorani	&	---	& 41.83	& 43.61	& 43.77	& --- & 56.70 & 57.29 & 57.26 \\	
Lingala	& 59.00 & 58.18	& 43.83	& 43.86	&	---	& 50.49	&	47.99 & 48.02 \\	
Luganda	&	---	& 43.07	& 48.24	& 48.53	&	---	&	51.23 &	---	& 55.67	\\	
Nepali	& 49.30	&	48.72 & 57.01	& 56.55	& 70.46	& 70.4	& 73.75	& 74.95 \\	
Nigerian Fulfulde	&	---	&	---	&	---	& 22.38	&	---	&	---	&	---	& 48.12	\\	
Oromo	&	---	&	---	& 43.75  & 44.00	& --- &	---	& 54.42	& 54.48 \\	
Pashto & 46.63 & 37.84 & 49.11	& 48.76	& 54.45	& 58.83	& 63.51	& 66.21	\\	
Somali & 53.01 & 53.01  & 45.16 & 45.97	& 54.30	& 54.37 & 57.53	& 37.13 \\	
Zulu	& 53.43	& 53.46	& 53.21	& 53.13	& 60.78	& 64.51	&	67.02 & 69.21	\\	\hline 
  
\end{tabular}
\caption{CHRF++ Score Differences between 2023 and 2025 for Microsoft and Google}
\label{tab:2023_2025_chrf_diffs}
\end{table*}

\clearpage
\begin{table*}
%\begin{tabular}{|*{9}{c|}}
\begin{tabular}{|c|c c c c|c c c c|}
\hline
\multicolumn{1}{|c|}{} & 
\multicolumn{4}{|c|}{\textbf{English to X}} & \multicolumn{4}{|c|}{\textbf{X to English}} \\ \hline
\multicolumn{1}{|c|}{\textbf{Language}} & \multicolumn{2}{|c|}{\textbf{Microsoft}} & \multicolumn{2}{|c|}{\textbf{Google}} & \multicolumn{2}{|c|}{\textbf{Microsoft}} & \multicolumn{2}{|c|}{\textbf{Google}} \\ \hline
& 2023 & 2025 & 2023 & 2025 & 2023 & 2025 & 2023 & 2025 \\ \hline

%% Pivot Languages first
\multicolumn{1}{|c|}{\textbf{------Pivot------}} & \multicolumn{8}{|c|}{} \\ \hline
Arabic	& 0.8841	& 0.8375	& 0.8791	& 0.8833	& 0.7661	& 0.7465	& 0.7701	& 0.7987 \\
Portuguese (BR) &	0.9363	& 0.8750	& 0.9339	& 0.9521	& 0.8434	& 0.8302	& 0.8248	& 0.8437 \\
Chinese	& 0.8930	& 0.8619	& 0.8971 &	0.9450	& 0.7373	& 0.7234	& 0.7271	& 0.7400 \\
French	& 0.8837	& 0.8300	& 0.8465	& 0.8838	& 0.7078	& 0.6863	& 0.6930	& 0.7164 \\
Hindi	& 0.8945	& 0.8570	& 0.9023	& 0.9649	& 0.8156	& 0.8103	& 0.8166	& 0.8423 \\
Indonesian	& 0.9325	& 0.8705	& 0.9308	& 0.9352	& 0.8300	& 0.8123	& 0.8114	& 0.8451 \\
Spanish (LA)	& 0.9379	& 0.8773	& 0.9349	& 0.9394	& 0.8362	& 0.8278	& 0.8191	& 0.8427 \\
Russian	& 0.8869 &	0.8339	& 0.8873	& 0.8907	& 0.7373	& 0.7214	& 0.7291 &	0.7584 \\
Swahili	& 0.8706	& 0.8263	& 0.8720	& 0.9281	& 0.7282	& 0.7235 &	0.7564	& 0.7814 \\ \hline

%% Important languages
%\\ \hline
\multicolumn{1}{|c|}{\textbf{------Important------}} & \multicolumn{8}{|c|}{} \\ \hline    
Bengali	& 0.8475	& 0.8192	& 0.8625	& 0.8755	& 0.7748	& 0.7702	& 0.7878	& 0.8169 \\
Farsi	& 0.8778	& 0.8377	& 0.8820	& 0.8904	& 0.7528	& 0.7400	& 0.7428	& 0.7900 \\
Malay	& 0.9283	& 0.8691	& 0.9286	& 0.9494	& 0.8151	& 0.8214	& 0.8342	& 0.8465 \\
Marathi	& 0.8417	& 0.8044	& 0.8413	& 0.8461	& 0.7302	& 0.7123	& 0.7487	& 0.7822 \\
Myanmar	& 0.7982	& 0.7835	& 0.8288	& 0.9441	& 0.5944	& 0.6627	& 0.6788	& 0.7104 \\
Tagalog	& ---	& 0.9400	& 0.9017	& 0.9406	& ---	& 0.8423	& ---	& 0.7427 \\
Tamil	& 0.8674	& 0.8352	& 0.8588	& 0.9647	& 0.6981	& 0.7203	& 0.7083	& 0.7465 \\
Urdu	& 0.8491	& 0.8095	& 0.8623	& 0.9514	& 0.7104	& 0.6927	& 0.7349	& 0.7536 \\ \hline

%% Priority languages
%\\ \hline
\multicolumn{1}{|c|}{\textbf{------Priority------}} & \multicolumn{8}{|c|}{} \\ \hline
Amharic	& 0.9503	& 0.8473	& 0.9412	& 0.9422	& 0.5900	& 0.6559	& 0.6915	& 0.7347 \\
Dari	& 0.8365	& 0.7956	& ---	& ---	& 0.6642	& 0.6578	& ---	& --- \\
Dinka	& ---	& ---	& ---	& 0.8674	& ---	& ---	& ---	& 0.3917 \\
Eritrean Tigrinya	& 0.9366	& 0.8418	& 0.9431	& 0.9375	& 0.5233	& 0.5212	& 0.6006	& 0.6394 \\
Ethiopian Tigrinya	& 0.9408	& 0.8449	& 0.9464	& 0.9374	& 0.5687	& 0.5227	& 0.6668	& 0.6394 \\
Hausa	& 0.7840	& 0.7853	& 0.8287	& 0.8359	& ---	& 0.6398	& 0.6719	& 0.7234 \\
Kanuri	& ---	& ---	& ---	& 0.8330	& ---	& ---	& ---	& 0.7800 \\
Khmer	& 0.8755	& 0.8017	& 0.9480	& 0.9600	& 0.6098	& 0.7234	& 0.6976	& 0.8200 \\
Kinyarwanda	& ---	& 0.8800	& ---	& 0.8000	& ---	& 0.5752	& ---	& 0.6358 \\
Kurdish Kurmanji	& ---	& 0.9300	& ---	& 0.8128	& ---	& 0.6337	& ---	& 0.7466 \\
Kurdish Sorani	& ---	& 0.8700	& 0.8643	& 0.8674	& ---	& 0.5620	& 0.6651	& 0.6773 \\
Lingala	& 0.8481	& 0.7961	& 0.7924	& 0.7936	& ---	& 0.6126	& 0.6075	& 0.6247 \\
Luganda	& ---	& 0.8700	& 0.8112	& 0.8147	& ---	& 0.6091	& ---	& 0.6972 \\
Nepali	& 0.8534	& 0.8180	& 0.8804	& 0.9595	& 0.7879	& 0.7823	& 0.8087	& 0.8293 \\
Nigerian Fulfulde	& ---	& ---	& ---	& 0.8400	& ---	& ---	& ---	& 0.5623 \\
Oromo	& ---	& ---	& 0.7963	& 0.8885	& ---	& ---	& 0.6477	& 0.6318 \\
Pashto	& 0.8472	& 0.7908	& 0.8517	& 0.9456	& 0.5942	& 0.6430	& 0.7229	& 0.7420 \\
Somali	& 0.7808	& 0.7487	& 0.7543	& 0.8635	& 0.3793	& 0.3405	& 0.3945	& 0.3474 \\
Zulu	& 0.8418	& 0.7876	& 0.8403	& 0.9136	& 0.7179	& 0.7396	& 0.7738	& 0.7536 \\ \hline

\end{tabular}
\caption{BERTScore Differences between 2023 and 2025 for Microsoft and Google}
\label{tab:2023_2025_BERTScore_diffs}
\end{table*}

\clearpage
\begin{table*}
%\begin{tabular}{|*{9}{c|}}
\begin{tabular}{|c|c c c c|c c c c|}
\hline
\multicolumn{1}{|c|}{} & 
\multicolumn{4}{|c|}{\textbf{English to X}} & \multicolumn{4}{|c|}{\textbf{X to English}} \\ \hline
\multicolumn{1}{|c|}{\textbf{Language}} & \multicolumn{2}{|c|}{\textbf{Microsoft}} & \multicolumn{2}{|c|}{\textbf{Google}} & \multicolumn{2}{|c|}{\textbf{Microsoft}} & \multicolumn{2}{|c|}{\textbf{Google}} \\ \hline
& 2023 & 2025 & 2023 & 2025 & 2023 & 2025 & 2023 & 2025 \\ \hline

%% Pivot Languages first
\multicolumn{1}{|c|}{\textbf{------Pivot------}} & \multicolumn{8}{|c|}{} \\ \hline
Arabic	& 0.8450	& 0.8253	& 0.8486	& 0.8813	& 0.8710	& 0.8400	& 0.8777	& 0.8891 \\
Portuguese (BR)	& 0.8978	& 0.8698	& 0.8976	& 0.8962	& 0.9007	& 0.8790	& 0.9002	& 0.8983 \\
Chinese	& 0.8824	& 0.8652	& 0.8924	& 0.8910	& 0.8738	& 0.8090	& 0.8755	& 0.8752 \\
French	& 0.8139	& 0.7967	& 0.8099	& 0.8910	& 0.8351	& 0.8400	& 0.8333	& 0.9028 \\
Hindi	& 0.7908	& 0.7835	& 0.8035	& 0.8272	& 0.9038	& 0.8700	& 0.9094	& 0.9136 \\
Indonesian	& 0.9156	& 0.8894	& 0.9168	& 0.9239	& 0.9042	& 0.9000	& 0.9032	& 0.9097 \\
Spanish	(LA) & 0.8858	& 0.8583	& 0.8835	& 0.8918	& 0.8979	& 0.8600	& 0.8962	& 0.9025 \\
Russian	& 0.8843	& 0.8633	& 0.8819	& 0.9000	& 0.8621	& 0.8100	& 0.8633	& 0.8724 \\
Swahili	& 0.8119	& 0.8007	& 0.8200	& 0.8412	& 0.8380	& 0.8200	& 0.8588	& 0.8649 \\ \hline

%% Important languages
%\\ \hline
\multicolumn{1}{|c|}{\textbf{------Important------}} & \multicolumn{8}{|c|}{} \\ \hline    
Bengali	& 0.8465	& 0.8328	& 0.8556	& 0.8751	& 0.8952	& 0.8700	& 0.9046	& 0.9124 \\
Farsi	& 0.8578	& 0.8490	& 0.8704	& 0.8933	& 0.8801	& 0.8300	& 0.8836	& 0.8931 \\
Malay	& 0.8964	& 0.8736	& 0.9010	& 0.8977	& 0.8901	& 0.8560	& 0.9032	& 0.9016 \\
Marathi	& 0.7255	& 0.7113	& 0.7164	& 0.7402	& 0.8741	& 0.8800	& 0.8858	& 0.8930 \\
Myanmar	& ---	& 0.8629	& ---	& 0.8673	& ---	& 0.8840	& ---	& 0.8801 \\
Tagalog	& ---	& 0.8140	& ---	& 0.8503	& ---	& 0.8200	& ---	& 0.9005 \\
Tamil	& ---	& 0.9348	& ---	& 0.92420	& ---	& 0.8600	& ---	& 0.8540 \\
Urdu	& 0.8062	& 0.7892	& 0.8168	& 0.8125	& 0.8618	& 0.8400	& 0.8770	& 0.8821 \\ \hline

%% Priority languages
%\\ \hline
\multicolumn{1}{|c|}{\textbf{------Priority------}} & \multicolumn{8}{|c|}{} \\ \hline
Amharic	&	0.8390	& 0.8306	& 0.8400	& 0.8641	& 0.8161	& 0.8500	& 0.8691	& 0.8722	\\	
Dari	&	---	&	---	&	---	&	---	&	---	&	---	&	---	&	---	\\	
Dinka	&	---	&	---	&	---	&	---	&	---	&	---	&	---	&	---	\\	
Eritrean Tigrinya	&	---	&	---	&	---	&	---	&	---	&	---	&	---	&	---	\\	
Ethiopian Tigrinya	&	---	&	---	&	---	&	---	&	---	&	---	&	---	&	---	\\	
Hausa	&	---	& 0.8548	&	--- & 0.7957	& ---	& 0.8100	& ---	& 0.8217	\\	
Kanuri	&	---	&	---	&	---	&	---	&	---	&	---	&	---	&	---	\\	
Khmer	&	--- &	0.7848	&	--- & 0.7928	&	--- & 0.8900	&	--- & 0.8854	\\	
Kinyarwanda	&	---	&	---	&	---	&	---	&	---	&	---	&	---	&	---	\\	
Kurdish Kurmanji	&	--- &	0.4583	& ---	& 0.8049	& --- &	0.8090	& ---	& 0.8370	\\	
Kurdish Sorani	&	---	&	---	&	---	&	---	&	---	&	---	&	---	&	---	\\	
Lingala	&	---	&	---	&	---	&	---	&	---	&	---	&	---	&	---	\\	
Luganda	&	---	&	---	&	---	&	---	&	---	&	---	&	---	&	---	\\	
Nepali	&	0.8013	& 0.7854	& 0.8227	& 0.8413	& 0.9050	& 0.8600	& 0.9162	& 0.9192	\\	
Nigerian Fulfulde	&	---	&	---	&	---	&	---	&	---	&	---	&	---	&	---	\\	
Oromo	&	---	&	---	&	---	&	0.7912	&	---	&	---	&	---	&	0.9030	\\	
Pashto	&	0.7828	& 0.7617	& 0.7921	& 0.8112	& 0.8192	& 0.8430	& 0.8631	& 0.8720	\\	
Somali	&	0.7189	& 0.7077	& 0.6995	& 0.7018	& 0.6617	& 0.8200	& 0.6831	& 0.4749	\\	
Zulu	&	---	&	---	&	---	&	---	&	---	&	---	&	---	&	---	\\	\hline

%Congolese Swahili & \cellcolor{gray}{---} & \cellcolor{gray}{---} & \cellcolor{gray}{---} & \cellcolor{gray}{---} & \cellcolor{gray}{---} & \cellcolor{gray}{---} & \cellcolor{gray}{---} & \cellcolor{gray}{---} \\ \hline 

\end{tabular}
\caption{COMET Score Differences between 2023 and 2025 for Microsoft and Google}
\label{tab:2023_2025_COMET_diffs}
\end{table*}

%\clearpage
%\input{tables/domains_for_contamination}

\end{document}